\documentclass[10pt,journal,compsoc]{IEEEtran}

\usepackage{times}
\usepackage{helvet}
\usepackage{courier}
\usepackage{caption}
\usepackage{subcaption}
\usepackage{graphicx}
\usepackage{multirow}
\usepackage{amsfonts}
\usepackage{amsmath}
\usepackage{relsize}
\usepackage{tabularx}
\usepackage{enumitem}
\usepackage{xcolor}
\usepackage[para,online,flushleft]{threeparttable}
\usepackage[ruled,vlined,noend]{algorithm2e}
\usepackage{xcolor}
\usepackage[hidelinks]{hyperref} 
\usepackage{ragged2e}
\usepackage{balance}

\newcommand{\MetaLoss}{\mathcal{M}}
\newcommand{\Loss}{\mathcal{L}}
\newcommand{\Task}{\mathcal{T}}
\newcommand{\Transpose}{\mathsf{T}}
\newcommand{\Dataset}{\mathcal{D}}
\newcommand{\Fitness}{\mathcal{F}}

\DeclareMathOperator*{\argminA}{arg\,min}

\ifCLASSOPTIONcompsoc
  \usepackage[nocompress]{cite}
\else
  \usepackage{cite}
\fi

\ifCLASSINFOpdf
\else
\fi

\begin{document}

\title{Learning Symbolic Model-Agnostic Loss Functions via Meta-Learning}
%
%
%
%

\author{
Christian~Raymond,~\IEEEmembership{Graduate Student Member,~IEEE,}
Qi~Chen,~\IEEEmembership{Member,~IEEE,}
Bing~Xue,~\IEEEmembership{Senior Member,~IEEE,}
and~Mengjie~Zhang,~\IEEEmembership{Fellow,~IEEE}


}

%
%

\markboth{IEEE Transaction on Pattern Analysis and Machine Intelligence,~Vol.~45, No.~11, November~2023}%
{Shell \MakeLowercase{\textit{et al.}}: Bare Demo of IEEEtran.cls for Computer Society Journals}

\makeatletter
\long\def\@IEEEtitleabstractindextextbox#1{\parbox{0.922\textwidth}{#1}}
\makeatother

\IEEEtitleabstractindextext{%
\begin{abstract}
In this paper, we develop upon the emerging topic of loss function learning, which aims to learn loss functions that significantly improve the performance of the models trained under them. Specifically, we propose a new meta-learning framework for learning model-agnostic loss functions via a hybrid neuro-symbolic search approach. The framework first uses evolution-based methods to search the space of primitive mathematical operations to find a set of symbolic loss functions. Second, the set of learned loss functions are subsequently parameterized and optimized via an end-to-end gradient-based training procedure. The versatility of the proposed framework is empirically validated on a diverse set of supervised learning tasks. Results show that the meta-learned loss functions discovered by the newly proposed method outperform both the cross-entropy loss and state-of-the-art loss function learning methods on a diverse range of neural network architectures and datasets.
\end{abstract}

}
\maketitle

\IEEEdisplaynontitleabstractindextext
\IEEEpeerreviewmaketitle

\IEEEraisesectionheading{\section{Introduction}
\label{sec:introduction}}

\IEEEPARstart{T}{he} field of learning-to-learn or \textit{meta-learning} has been an area of increasing interest to the machine learning community in recent years \cite{vanschoren2018meta,peng2020comprehensive}. In contrast to conventional learning approaches, which learn from scratch using a static learning algorithm, meta-learning aims to provide an alternative paradigm whereby intelligent systems leverage their past experiences on related tasks to improve future learning performances \cite{hospedales2020meta}. This paradigm has provided an opportunity to utilize the shared structure between problems to tackle several traditionally very challenging deep learning problems in domains where both data and computational resources are limited \cite{altae2017low,ignatov2019ai}. 

Many meta-learning approaches have been proposed for optimizing various components of deep neural networks. For example, early research on the topic explored using meta learning for generating learning rules \cite{schmidhuber1987evolutionary,schmidhuber1992learning,bengio1994use,andrychowicz2016learning}. More recent research has extended itself to learning everything from activation functions \cite{ramachandran2017searching}, shared parameter/weight initializations \cite{finn2017model,nichol2018first, rajeswaran2019meta, song2019maml}, and neural network architectures \cite{kim2018auto, stanley2019designing, elsken2020meta, ding2022learning} to whole learning algorithms from scratch \cite{real2020automl,co2021evolving} and more \cite{raymond2023fast, raymond2024metanpbml}.

However, one component that has been overlooked until very recently is the loss function \cite{wang2022comprehensive}. Typically in deep learning, neural networks are trained through the backpropagation of gradients originating from a handcrafted and manually selected loss function \cite{rumelhart1986learning}. One significant drawback of this approach is that traditionally loss functions have been designed with task-generality in mind, \textit{i.e.} large classes of tasks in mind, but the system itself is only concerned with a single instantiation or small subset of that class. However, as shown by the \textit{No Free Lunch Theorems} \cite{wolpert1997no} no algorithm is able to do better than a random strategy in expectation --- this suggests that specialization to a subclass of tasks is in fact the only way that performance can be improved in general.

Given this importance, the prototypical approach of selecting a loss function heuristically from a modest set of handcrafted loss functions should be reconsidered in favor of a more principled data-informed approach. The new and emerging subfield of loss function learning \cite{raymond2024meta} offers an alternative to this, which instead aims to leverage task-specific information and past experiences to infer and discover highly performant loss functions directly from the data. Initial approaches to loss function learning have shown promise in improving various aspects of deep neural networks training. However, they have several key issues and limitations which must be addressed for meta-learned loss functions to become a more desirable alternative than handcrafted loss functions.

In particular, many loss function learning approaches use a parametric loss representation such as a neural network \cite{bechtle2021meta} or Taylor polynomial \cite{gonzalez2021optimizing,gao2021searching,gao2022loss}, which is limited as it imposes unnecessary assumptions and constraints on the structure of the learned loss function. However, the current non-parametric alternative to this is to use a two-stage discovery and optimization process, which infers both the loss function structure and parameters simultaneously using genetic programming and covariance matrix adaptation \cite{gonzalez2020improved}, and quickly become intractable for large-scale optimization problems. Subsequent work \cite{liu2020loss,li2022autoloss} has attempted to address this issue; however, they crucially omit the optimization stage, which is known to produce sub-optimal performance. 

This paper aims to resolve these issues through a newly proposed framework called Evolved Model-Agnostic Loss (EvoMAL), which meta-learns non-parametric symbolic loss functions via a hybrid neuro-symbolic search approach. The newly proposed framework aims to resolve the limitations of past approaches to loss function learning by combining genetic programming \cite{koza1992genetic} with an efficient gradient-based local-search procedure \cite{maclaurin2015gradient,grefenstette2019generalized}. This unifies two previously divergent lines of research on loss function learning, which prior to this method, exclusively used either a gradient-based or an evolution-based approach.

This work innovates on the prior loss function learning approaches by introducing the first computationally tractable approach to optimizing symbolic loss functions. Consequently, improving the scalability and performance of symbolic loss function learning algorithm. Furthermore, unlike prior approaches, the proposed framework is both task and model-agnostic, as it can be applied to learning algorithm trained with a gradient descent style procedure and is compatible with different model architectures. This branch of general-purpose loss function learning algorithms provides a new powerful avenue for improving a neural network's performance, which has until recently not been explored.

The performance of EvoMAL is assessed on a diverse range of datasets and neural network architectures in the direct learning and transfer learning settings, where the empirical performance is compared with the ubiquitous cross-entropy loss and other state-of-the-art loss function learning methods. Finally, an analysis of the meta-learned loss functions produced by EvoMAL is presented, where several reoccurring trends are identified in both the shape and structure. Further analysis is also given to show why meta-learned loss functions are so performant through 1) examining the loss landscapes of the meta-learned loss functions and 2) investigating the relationship between the base learning rate and the meta-learned loss functions.

\subsection{Contributions:}

The key contributions of this work are as follows:

\begin{itemize}[leftmargin=*]

    \item We propose a new task and model-agnostic search space and a corresponding search algorithm for meta-learning interpretable symbolic loss functions.

    \item We demonstrate a simple transition procedure for converting expression tree-based symbolic loss functions into gradient trainable loss networks.

    \item We utilize the new loss function representation to integrate the first computationally tractable approach to optimizing symbolic loss functions into the framework.

    \item We evaluate the proposed framework by performing the first-ever comparison of existing loss function learning techniques in both direct learning and transfer learning settings.

    \item We analyze the meta-learned loss functions to highlight key trends and explore why meta-learned loss functions are so performant.
    
\end{itemize}

\section{Background and Related Work}
\label{sec:background}

The goal of loss function learning in the meta-learning context is to learn a loss function $\MetaLoss_{\phi}$ with parameters $\phi$, at meta-training time over a distribution of tasks $p(\Task)$. A \textit{task} is defined as a set of input-output pairs $\Task = \{(x_{1}, y_{1}), \dots, (x_{N}, y_{N})\}$, and multiple tasks compose a \textit{meta-dataset} $\Dataset = \{\Task_{1}, \dots, \Task_{M}\}$. Then, at meta-testing time the learned loss function $\MetaLoss_{\phi}$ is used in place of a traditional loss function to train a base learner, e.g. a classifier or regressor, denoted by $f_{\theta}(x)$ with parameters $\theta$ on a new unseen task from $p(\Task)$. In this paper, we constrain the selection of base learners to models trainable via a gradient descent style procedures such that we can optimize weights $\theta$ as follows:
\begin{equation}
\theta_{new} = \theta - \alpha \nabla_{\theta} \MetaLoss_{\phi}(y, f_{\theta}(x))
\label{eq:setup}
\end{equation}
\noindent
Several approaches have recently been proposed to accomplish this task, and an  observable trend is that most of these methods fall into one of the following two key categories.
\subsection{Gradient-Based Approaches}

Gradient-based approaches predominantly aim to learn a loss function $\MetaLoss$ through the use of a meta-level neural network to improve on various aspects of the training. For example, in \cite{grabocka2019learning,huang2019addressing}, differentiable surrogates of non-differentiable performance metrics are learned to reduce the misalignment problem between the performance metric and the loss function. Alternatively, in \cite{houthooft2018evolved,wu2018learning,antoniou2019learning,kirsch2019improving,bechtle2021meta,collet2022loss,leng2022polyloss,raymond2023online}, loss functions are learned to improve sample efficiency and asymptotic performance in supervised and reinforcement learning, while in \cite{balaji2018metareg,barron2019general,li2019feature,gao2022loss}, they improved on the robustness of a model.

While the aforementioned approaches have achieved some success, they have notable limitations. The most salient limitation is that they \textit{a priori} assume a parametric form for the loss functions. For example, in \cite{bechtle2021meta} and \cite{psaros2022meta}, it is assumed that the loss functions take on the parametric form of a two hidden layer feed-forward neural network with 50 nodes in each layer and ReLU activations. However, such an assumption imposes a bias on the search, often leading to an over parameterized and sub-optimal loss function. Another limitation is that these approaches often learn black-box (sub-symbolic) loss functions, which is not ideal, especially in the meta-learning context where \textit{post hoc} analysis of the learned component is crucial, before transferring the learned loss function to new unseen problems at meta-testing time.

\subsection{Evolution-Based Approaches}

A promising alternative paradigm is to use evolution-based methods to learn $\MetaLoss_{\phi}$, favoring their inherent ability to avoid local optima via maintaining a population of solutions, their ease of parallelization of computation across multiple processors, and their ability to optimize for non-differentiable functions directly. Examples of such work include \cite{gonzalez2021optimizing} and \cite{gao2021searching}, which both represent $\MetaLoss_{\phi}$ as parameterized Taylor polynomials optimized with covariance matrix adaptation evolutionary strategies (CMA-ES). These approaches successfully derive interpretable loss functions, but similar to previously, they also assume the parametric form via the degree of the polynomial.

To resolve the issue of having to assume the parametric form of $\MetaLoss_{\phi}$, another avenue of research first presented in \cite{gonzalez2020improved} investigated the use of genetic programming (GP) to learn the structure of $\MetaLoss_{\phi}$ in a symbolic form before applying CMA-ES to optimize the parameterized loss. The proposed method was effective at learning performant loss functions and clearly demonstrated the importance of local-search. However, the method had intractable computational costs as using a population-based method (GP) with another population-based method (CMA-ES) resulted in a significant expansion in the number of evaluations at meta-training time, hence it needing to be run on a supercomputer in addition to using a truncated number of training steps.

Subsequent work in \cite{liu2020loss} and \cite{li2022autoloss} reduced the computational cost of GP-based loss function learning approaches by proposing time saving mechanisms such as: rejection protocols, gradient-equivalence-checking, convergence property verification and model optimization simulation. These methods successfully reduced the wall-time of GP-based approaches; however, both papers omit the use of local-search strategies, which is known to cause sub-optimal performance when using GP \cite{topchy2001faster,smart2004applying,zhang2005learning}. Furthermore, neither method is task and model-agnostic, limiting their utility to a narrow set of domains and applications.

\section{Evolved Model-Agnostic Loss}
\label{sec:method}

In this work, a novel hybrid neuro-symbolic search approach named \textit{Evolved Model-Agnostic Loss (EvoMAL)} is proposed, which consolidates and extends past research on the topic of loss function learning. The proposed method learns performant and interpretable symbolic loss functions by inferring both the structure and the weights/coefficients directly from the data. The evolution-based technique GP \cite{koza1992genetic} is used to solve the discrete problem of deriving the symbolic structure of the learned loss functions, while unrolled differentiation \cite{wengert1964simple, domke2012generic, deledalle2014stein, maclaurin2015gradient}, a gradient-based technique previously used in Meta-Learning via Learned Loss (ML$^3$) \cite{bechtle2021meta}, and sometimes referred to as Generalized Inner Loop Meta-Learning \cite{grefenstette2019generalized}, is used to solve the continuous problem of optimizing their weights/coefficients.

\subsection{Offline Loss Function Learning Setup}

EvoMAL is a new approach to offline loss function learning, a meta-learning paradigm concerned with learning new and performant loss functions that can be used as a drop-in replacement for a prototypical handcrafted loss function such as the squared loss for regression or cross-entropy loss for classification \cite{gonzalez2020improving,gao2023meta}. Offline loss function learning follows a conventional offline meta-learning setup, which partitions the learning into two sequential phases: meta-training and meta-testing.

\subsubsection{Meta-Training Phase}

The meta-training phase in EvoMAL is formulated as a bilevel optimization problem\footnote{For simplicity, EvoMAL is presented as a bilevel optimization problem; however, since $\MetaLoss$ and $\phi$ are learned using separate processes, EvoMAL can be viewed as solving for a tri-level optimization problem with the outer optimization from Equation (\ref{eq:loss-function-learning-goal-new}) being split into two distinct phases, 1) learning the symbolic loss functions structure, and 2) optimizing the loss functions weights/coefficients.}, where the goal of the outer optimization is to meta-learn a performant loss function $\MetaLoss_{\phi}$ minimizing the average task loss $\Loss_{\Task}$ (the meta-objective) across $m$ related tasks, where $\Loss_{\Task}$ is selected based on the desired task. The inner optimization uses $\MetaLoss_{\phi}$ as the base loss function to train the base model parameters $\theta$. Formally, the meta-training phase of EvoMAL is defined as follows:
\begin{align}
\label{eq:loss-function-learning-goal-new}
\begin{split}
    \MetaLoss_{\phi}^* &= \argminA_{\MetaLoss_\phi} \frac{1}{m}\sum_{i=1}^{m}\Loss_{\Task}(y_{i}, f_{\theta^{*}_{i}}(x_{i}))\\ 
    s.t.\;\;\;\;\; \theta^{*}_{i}(\phi)  &= \argminA_{\theta_{i}} \big[\MetaLoss_{\phi}(y, f_{\theta_{i}}(x))\big]
\end{split}
\end{align}

\subsubsection{Meta-Testing Phase}

In the meta-testing phase, the best-performing loss function learned at meta-training time $\MetaLoss_{\phi}^*$ is used directly to train and optimize the base model parameters $\theta$.
\begin{equation}\label{eq:meta-testing-phase}
\theta^{*} = \argminA_{\theta} \big[\MetaLoss_{\phi}^{*}(y, f_{\theta}(x))\big]
\end{equation}
In contrast to online loss function learning \cite{raymond2023online}, offline loss function learning does not require any alterations to the existing training pipelines to accommodate the meta-learned loss function, this makes the loss functions easily transferable and straightforward to use in existing code bases.

\subsection{Learning Symbolic Loss Functions}
\label{sec:search-space}

To learn the symbolic structure of the loss functions in EvoMAL, we propose using GP, a powerful population-based technique that employs an evolutionary search to directly search the set of primitive mathematical operations \cite{koza1992genetic}. In GP, solutions are composed of terminal and function nodes in a variable-length hierarchical expression tree-based structure. This symbolic structure is a natural and convenient way to represent loss functions, due to its high interpretability and trivial portability to new problems. Transferring a learned loss function from one problem to another requires very little effort, typically only a line or two of additional code. The task and model-agnostic loss functions produced by EvoMAL can be used directly as a drop-in replacement for handcrafted loss functions without requiring any new sophisticated meta-learning pipelines to train the loss on a per-task basis.

\subsubsection{Search Space Design}

\begin{table}[]
\captionsetup{justification=centering}
\caption{Set of searchable primitive mathematical operations.}
\centering
\resizebox{0.8\columnwidth}{!}{
\begin{tabular}{lcc}
\hline
\textbf{Operator}  & \textbf{Expression}      & \textbf{Arity} \\ \hline \noalign{\vskip 1mm} 
Addition           & $x_1 + x_2$              & 2              \\
Subtraction        & $x_1 - x_2$              & 2              \\
Multiplication     & $x_1 * x_2$              & 2              \\
Division ($AQ$)    & $x_1 / \sqrt{1 + \smash{x_2^2}}$ & 2   \\
Minimum            & $\min(x_1, x_2)$         & 2              \\
Maximum            & $\max(x_1, x_2)$         & 2              \\ \noalign{\vskip 1mm}  \hline \noalign{\vskip 1mm} 
Sign               & sign$(x)$                & 1              \\
Square             & $x^2$                    & 1              \\
Absolute           & $|x|$                    & 1              \\
Logarithm          & $\log(|x| + \epsilon)$   & 1              \\
Square Root        & $\sqrt{\smash[b]{|x| + \epsilon}}$ & 1    \\
Hyperbolic Tangent & $\tanh(x)$               & 1              \\ \noalign{\vskip 1mm}  \hline 
\end{tabular}
}
\label{table:function-set}
\end{table}

In order to utilize GP, a search space containing promising loss functions must first be designed. When designing the desired search space, four key considerations are made --- first, the search space should superset existing loss functions such as the squared error in regression and the cross entropy loss in classification. Second, the search space should be dense with promising new loss functions while also containing sufficiently simple loss functions such that cross task generalization can occur successfully at meta-testing time. Third, ensuring that the search space satisfies the key property of GP closure, \textit{i.e.} loss functions will not cause \textit{NaN}, \textit{Inf}, undefined, or complex output. Finally, ensuring that the search space is both task and model-agnostic. With these considerations in mind, we present the function set in Table \ref{table:function-set}. Regarding the terminal set, the loss function arguments $f_{\theta}(x)$ and $y$ are used, as well as (ephemeral random) constants $+1$ and $-1$. Unlike previously proposed search spaces for loss function learning, we have made several necessary amendments to ensure proper GP closure, and sufficient task and model-generality. The salient differences are as follows: 

\begin{itemize}[leftmargin=*]

  \item Previous work in \cite{gonzalez2020improved} uses unprotected operations: natural log ($\log(x)$), square root ($\sqrt x$), and division $(x_1/x_2)$. Using these unprotected operations can result in imaginary or undefined output violating the GP closure property. To satisfy the closure property, we replace both the natural log and square root with protected alternatives, as well as replace the division operator with the analytical quotient ($AQ$) operator, a smooth and differentiable approximation to the division operator \cite{ni2012use}.
  
  \item The proposed search space for loss functions is both task and model-agnostic in contrast to \cite{liu2020loss} and \cite{li2022autoloss}, which use multiple aggregation-based and element-wise operations in the function set. These operations are suitable for object detection (the respective paper's target domain) but are not compatible when applied to other tasks such as tabulated and natural language processing problems.
  
\end{itemize}

\subsubsection{Search Algorithm Design}

\begin{figure}
\centering
    \includegraphics[width=0.87\columnwidth]{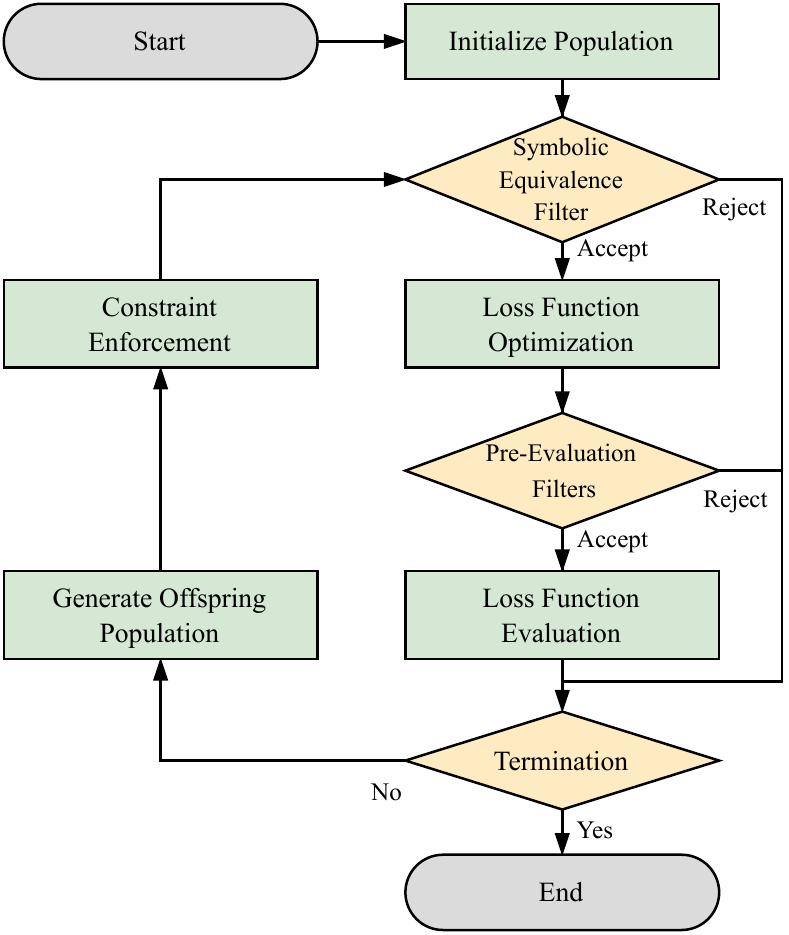}
    \caption{An overview of the EvoMAL framework, showing the pipeline for learning loss functions at meta-training time.}
\label{fig:flowchart}
\end{figure}

The symbolic search algorithm used in EvoMAL follows a prototypical implementation of GP; as shown in Fig. \ref{fig:flowchart}. First, initialization is performed via randomly generating a population of 25 expression trees, where the inner nodes are selected from the function set and the leaf nodes from the terminal set. Subsequently, the main loop begins by performing the loss function optimization and evaluation stages to determine each loss function's respective fitness, discussed in Sections \ref{sec:optimization} and \ref{sec:evaluation}, respectively. Following this, a new offspring population of equivalent size is constructed via crossover, mutation, and elitism. For crossover, two loss functions are selected via tournament selection and combined using a one-point crossover with a crossover rate of $70\%$. For mutation, a loss function is selected, and a uniform mutation is applied with a mutation rate of $25\%$. Finally, to ensure performance does not degrade, elitism is used to retain top-performing loss functions with an elitism rate of $5\%$. The main loop is iteratively repeated 50 times, and the loss function with the best fitness is selected as the final learned loss function. Note, to reduce the computational overhead of the meta-learning process, a number of time-saving measures, which we further refer to as filters, have been incorporated into the EvoMAL algorithm. This is discussed in detail in Section \ref{sec:filters}.

\subsubsection{Constraint Enforcement}
\label{sec:constraints}

When using GP, the evolved expressions often violate the constraint that a loss function must have as arguments $f_{\theta}(x)$ and $y$. Our preliminary investigation found that often over 50\% of the loss functions in the first few generations violated this constraint. Thus far, existing methods for handling this issue have been inadequate; for example, in \cite{gonzalez2020improved}, violating loss functions were assigned the worst-case fitness, such that selection pressure would phase out those loss functions from the population. Unfortunately, this approach degrades search efficiency, as a subset of the population is persistently searching infeasible regions of the search space. To resolve this, we propose a simple but effective corrections strategy to violating loss functions, which randomly selects a terminal node and replaces it with a random binary node, with arguments $f_{\theta}(x)$ and $y$ in no predetermined order.

An additional optional constraint enforceable in the EvoMAL algorithm is that the learned loss function can always return a non-negative output $\MetaLoss:\mathbb{R}^2 \rightarrow \mathbb{R}_{0}^{+}$. This is achieved via the loss function's output being passed through an output activation function $\varphi$, such as the smooth $Softplus(x)=ln(1+e^{x})$ activation. Note that this can be omitted by using an $Identity(x)=x$ activation if we choose not to enforce this constraint. 

\subsection{Loss Function Optimization}
\label{sec:optimization}

\begin{figure}%
    \centering
    \subfloat[\centering Example learned loss function $\MetaLoss$.]
    {{\includegraphics[width=6cm]{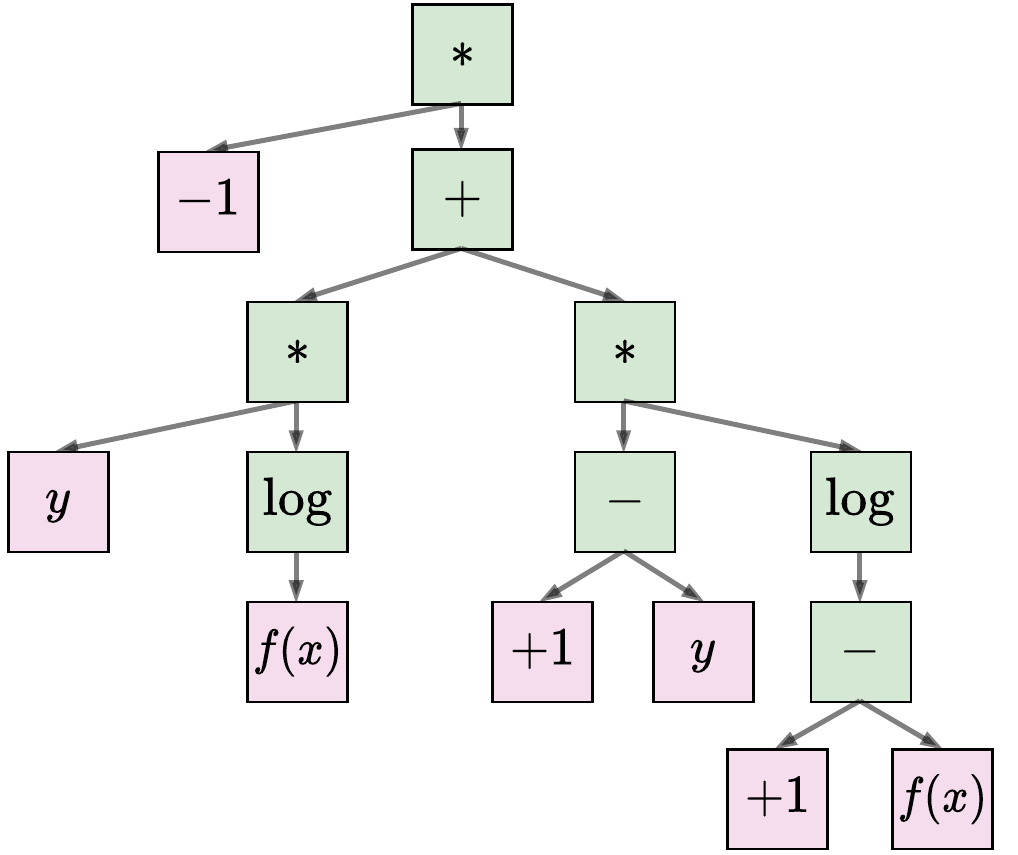} }}%
    \qquad
    \subfloat[\centering Example meta-loss network $\MetaLoss_{\phi}^{\Transpose}$.]
    {{\includegraphics[width=7.5cm]{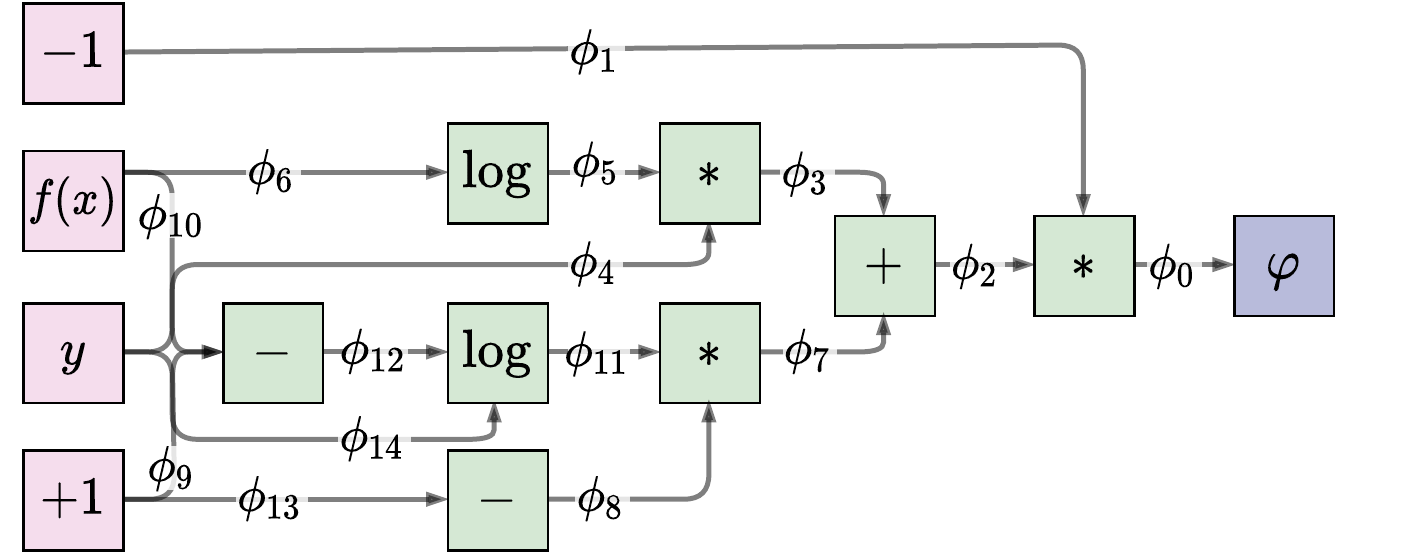} }}%
    \caption{Overview of the transitional procedure used to covert $\MetaLoss$ into a trainable meta-loss network $\MetaLoss_{\phi}^{\Transpose}$.}%
    \label{fig:phase-transformer}%
    \vspace{-3mm}
\end{figure}

Numerous empirical results have shown that local-search is imperative when using GP to get state-of-the-art results \cite{chen2015generalisation,gonzalez2020improved}. Therefore, unrolled differentiation \cite{franceschi2017forward, franceschi2018bilevel, shaban2019truncated, scieur2022curse}, an efficient gradient-based local search approach is integrated into the proposed method. To utilize unrolled differentiation in the EvoMAL framework, we must first transform the expression tree-based representation of $\MetaLoss$ into a compatible representation. In preparation for this, a transitional procedure takes each loss function $\MetaLoss$, represented as a GP expression and converts it into a trainable network, as shown in Fig. \ref{fig:phase-transformer}. First, a graph transpose operation $\MetaLoss^{\Transpose}$ is applied to reverse the edges such that they now go from the terminal (leaf) nodes to the root node. Following this, the edges of $\MetaLoss^{\Transpose}$ are parameterized by $\phi$, giving $\MetaLoss_{\phi}^\Transpose$, which we further refer to as a meta-loss network to delineate it clearly from its prior state. Finally, to initialize $\MetaLoss_{\phi}^\Transpose$, the weights are sampled from $\phi\sim\mathcal{N}(1, 1\mathrm{e}{-3})$, such that $\MetaLoss_{\phi}^\Transpose$ is initialized from its (near) original unit form, where the small amount of variance is to break any network symmetry. 

\subsubsection{Unrolled Differentiation}
\begin{algorithm}[t]
\SetAlgoLined
\DontPrintSemicolon
\SetKwInput{Input}{In}
\BlankLine
\Input{
    $\MetaLoss \leftarrow$ Loss function learned by GP\newline
    $\mathcal{S}_{meta} \leftarrow$ Number of meta gradient steps\newline
    $\mathcal{S}_{base} \leftarrow$ Number of base gradient steps
}

\hrulefill
\BlankLine
$\MetaLoss_{\phi}^{\Transpose} \leftarrow$ Transpose and parameterize edges of $\MetaLoss$\;
$\phi_0 \leftarrow$ Initialize loss network weights $\MetaLoss_{\phi}^{\Transpose}$\;
\For{$i \in \{0, ... , \mathcal{S}_{meta}\}$}{
    \For{$j \in \{0, ... , |\Dataset_{Train}|\}$}{
        $\theta_0 \leftarrow$ Initialize parameters of base learner\;
        \For{$t \in \{0, ... , \mathcal{S}_{base}\}$}{
            $X_j$, $y_j$ $\leftarrow$ Sample task $\Task_{j} \sim p(\Task)$\;
            $\MetaLoss_{learned} \leftarrow \MetaLoss_{\phi_i}^{\Transpose}(y_j, f_{\theta_t}(X_j))$\;
            $\theta_{t+1} \leftarrow \theta_{t} - \alpha \nabla_{\theta_t} \MetaLoss_{learned}$\;
        }
        $\Loss_{task_{j}} \leftarrow \Loss_{\Task}(y_j, f_{\theta_{t+1}}(X_j))$\;
    }
    $\phi_{i+1} \leftarrow \phi_{i} - \eta \nabla_{\phi_i} \sum_{j} \Loss_{task_{j}}$\;
}

\BlankLine

\caption{Loss Function Optimization}
\label{alg:meta-training}
\end{algorithm}
For simplicity, we constrain the description of loss function optimization to the vanilla backpropagation case where the meta-training set $\Dataset_{Train}$ contains one task, \textit{i.e.} $|\Dataset_{Train}| = 1$; however, the full process where $|\Dataset_{Train}| > 1$ is given in Algorithm \ref{alg:meta-training}. 

To learn the weights $\phi$ of the meta-loss network $\MetaLoss_{\phi}^\Transpose$ at meta-training time with respect to base learner $f_{\theta}(x)$, we first use the initial values of $\phi$ to produce a base loss value $\MetaLoss_{learned}$ based on the forward pass of $f_{\theta}(x)$.
\begin{equation}
\MetaLoss_{learned} = \MetaLoss_{\phi}^{\Transpose}(y, f_{\theta}(x))
\label{eq:loss-base}
\end{equation}
\noindent
Using $\MetaLoss_{learned}$, the weights $\theta$ are optimized by taking a predetermined number of inner base gradient steps $\mathcal{S}_{base}$, where at each step a new batch is sampled and a new base loss value is computed. Similar to the findings in \cite{bechtle2021meta}, we find $\mathcal{S}_{base}=1$ is usually sufficient to obtain good results. Each step is computed by taking the gradient of the loss value with respect to $\theta$, where $\alpha$ is the base learning rate. 
\begin{equation}
\begin{split}
\theta_{new}
& = \theta - \alpha \nabla_{\theta} \MetaLoss_{\phi}^{\Transpose}(y, f_{\theta}(x)) \\
& = \theta - \alpha \nabla_{\theta} \mathbb{E}_{\mathsmaller{X}, y} \big[ \MetaLoss_{\phi}^{\Transpose}(y, f_{\theta}(x)) \big]
\end{split}
\label{eq:backward-base}
\end{equation}
\noindent
where the gradient computation can be decomposed via the chain rule into the gradient of $\MetaLoss_{\phi}^{\Transpose}$ with respect to the product of the base learner predictions $f_{\theta}(x)$ and the gradient of $f$ with parameters $\theta$.
\begin{equation}
\theta_{new} = \theta - \alpha \nabla_{f} \MetaLoss_{\phi}^{\Transpose}(y, f_{\theta}(x)) \nabla_{\theta}f_{\theta}(x)
\label{eq:backward-base-decompose}
\end{equation}
Following this, $\theta$ has been updated to $\theta_{new}$ based on the current meta-loss network weights; $\phi$ now needs to be updated to $\phi_{new}$ based on how much learning progress has been made. Using the new base learner weights $\theta_{new}$ as a function of $\phi$, we utilize the concept of a \textit{task loss} $\Loss_{\Task}$ to produce a meta loss value $\Loss_{task}$ to optimize $\phi$ through $\theta_{new}$.
\begin{equation}
\Loss_{task} = \Loss_{\Task}(y, f_{\theta_{new}}(x))
\label{eq:loss-meta}
\end{equation}
where $\Loss_{\Task}$ is selected based on the respective application --- for example, the mean squared error loss for the task of regression or the categorical cross-entropy loss for multi-class classification. Optimization of the meta-loss network loss weights $\phi$ now occurs by taking the gradient of $\Loss_{\Task}$ with respect to $\phi$, where $\eta$ is the meta learning rate.
\begin{equation}
\begin{split}
\phi_{new}
& = \phi - \eta \nabla_{\phi}\Loss_{\Task}(y, f_{\theta_{new}}(x)) \\
& = \phi - \eta \nabla_{\phi} \mathbb{E}_{\mathsmaller{X}, y} \big[ \Loss_{\Task}(y, f_{\theta_{new}}(x)) \big]
\end{split}
\label{eq:backward-meta}
\end{equation}
where the gradient computation can be decomposed by applying the chain rule as shown in Equation \eqref{eq:backward-meta-decompose} where the gradient with respect to the meta-loss network weights $\phi$ requires the new model parameters $\theta_{new}$. 
\begin{equation}
\phi_{new} = \phi - \eta \nabla_{f}\Loss_{\Task} \nabla_{\theta_{new}} f_{\theta_{new}} \nabla_{\phi}\theta_{new} 
\label{eq:backward-meta-decompose}
\end{equation}
This process is repeated for a predetermined number of meta gradient steps $S_{meta}=250$, which was selected via cross-validation. Following each meta gradient step, the base learner weights $\theta$ are reset. Note that Equations (\ref{eq:backward-base})--(\ref{eq:backward-base-decompose}) and (\ref{eq:backward-meta})--(\ref{eq:backward-meta-decompose}) can alternatively be performed via automatic differentiation.

\subsection{Loss Function Evaluation}
\label{sec:evaluation}

\begin{figure*}[htb!]
    
    \centering
    \includegraphics[width=0.89\textwidth]{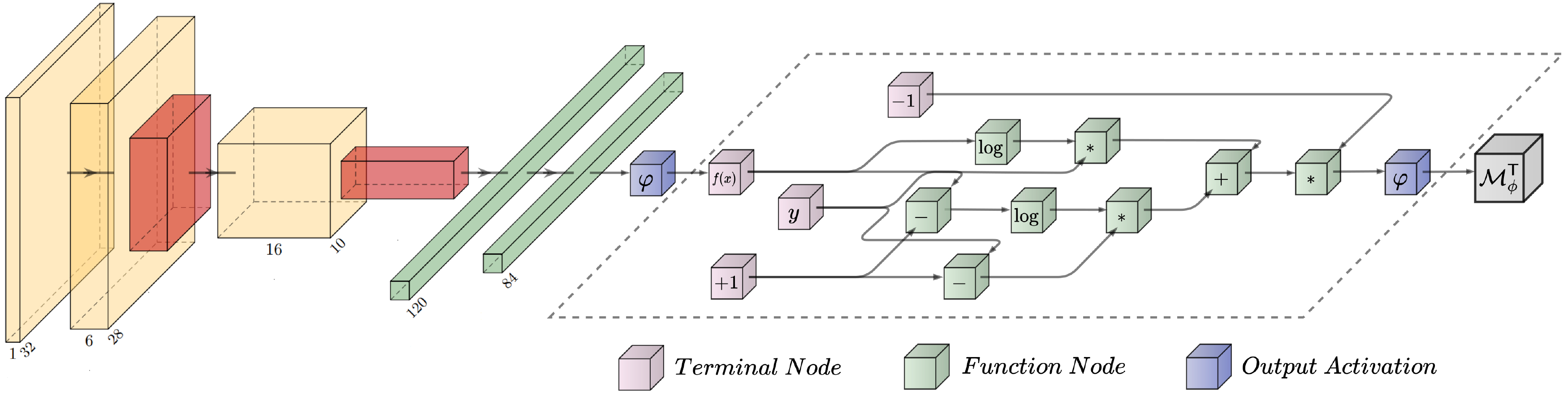}
    \captionsetup{justification=centering}
    \caption{Overview of the EvoMAL algorithm at \textit{meta-testing} time, where the base network $f_{\theta}(x)$, shown (left) as the popular LeNet-5 architecture, is trained using the meta-loss network $\MetaLoss_{\phi}^{\Transpose}$ (right) found at \textit{meta-training} time as the loss function.}
    \vspace{-3mm}

\label{fig:network}
\end{figure*}

To derive the fitness $\Fitness$ of $\MetaLoss_{\phi}^{\Transpose}$, a conventional training procedure is used as summarized in Algorithm \ref{alg:meta-testing}, where $\MetaLoss_{\phi}^{\Transpose}$ is used in place of a traditional loss function to train $f_{\theta}(x)$ over a predetermined number of base gradient steps $S_{testing}$. This training process is identical to training at meta-testing time as shown in Fig. \ref{fig:network}. The final inference performance of $\MetaLoss_{\phi}^{\Transpose}$ is assigned to $\Fitness$, where any differentiable or non-differentiable performance metric $\Loss_{\mathcal{P}}$ can be used. For our experiments, we calculate $\Fitness$ using the error rate for classification and mean squared error for regression.

\begin{algorithm}[t]
\SetAlgoLined
\DontPrintSemicolon
\SetKwInput{Input}{In}
\BlankLine
\Input{
    $\MetaLoss_{\phi}^{\Transpose} \leftarrow$ Loss function learned by EvoMAL\newline
    $S_{testing} \leftarrow$ Number of base testing gradient steps
}

\hrulefill
\BlankLine
\For{$j \in \{0, ... , |\Dataset|\}$}{
    $\theta_{0} \leftarrow$ Initialize parameters of base learner\;
    $X_{j}$, $y_{j}$ $\leftarrow$ Sample task $\Task_{j} \sim p(\Task)$\;
    \For{$t \in \{0, ... , S_{testing}\}$}{
        $\Loss_{learned} \leftarrow \MetaLoss_{\phi}^{\Transpose}(y_{j}, f_{\theta_{t}}(X_{j}))$\;
        $\theta_{t+1} \leftarrow \theta_{t} - \alpha \nabla_{\theta_{t}} \Loss_{learned}$\;
    }
}
$\Fitness \leftarrow \frac{1}{|\Dataset|}\sum_{j}\Loss_{\mathcal{P}}(y_{j}, f_{\theta_{t+1}}(X_{j}))$

\BlankLine

\caption{Loss Function Evaluation}
\label{alg:meta-testing}
\end{algorithm}

\subsection{Time Saving Measures (Filters)}
\label{sec:filters}

Optimizing and evaluating a large number of candidate loss functions can become prohibitively expensive. Fortunately, in the case of loss function learning, a number of techniques can be employed to reduce significantly the computational overhead of the otherwise very costly meta-learning process. 

\subsubsection{Symbolic Equivalence Filter}

For the GP-based symbolic search, a loss archival strategy based on a key-value pair structure with $\Theta(1)$ lookup is used to ensure that symbolically equivalent loss functions are not reevaluated. Two expression trees are said to be symbolically equivalent when they contain identical operations (nodes) in an identical configuration \cite{wong2006algebraic, kinzett2009numerical, javed2022simplification}. Loss functions that are identified by the symbolic equivalence filter skip both the loss function optimization and evaluation stage and are placed directly in the offspring population with their fitness cached.

\subsubsection{Pre-Evaluation Filter - Poor Training Dynamics}

Following loss function optimization, the candidate solution's fitnesses are evaluated. This costly fitness evaluation can be obviated in many cases since many of the loss functions found, especially in the early generations, are non-convergent and produce poor training dynamics. We use the loss rejection protocol employed in \cite{li2022autoloss} as a filter to identify candidate loss functions that should skip evaluation and be assigned the worst-case fitness automatically.

The loss rejection protocol takes a batch of $B$ randomly sampled instances from $\Dataset_{Train}$ and using an untrained network $f_{\theta_{0}}(x)$ produces a set of predictions and their corresponding true target values $\{(\hat{y_b}, y_b)\}_{b=1}^{B}$. As minimizing the proper loss function $\MetaLoss_{\phi}^{\Transpose}$ should correspond closely with optimizing the performance metric $\Loss_{\mathcal{P}}$, a correlation $g$ between $\MetaLoss_{\phi}^{\Transpose}$ and $\Loss_{\mathcal{P}}$ can be calculated. 
\begin{equation}
g(\MetaLoss_{\phi}^{\Transpose}) = \sum_{b=1}^{B} \Big[\Loss_{\mathcal{P}}(\hat{y_{b}}, y_{b}) - \Loss_{\mathcal{P}}(\hat{y_{b}^{*}}(\MetaLoss_{\phi}^{\Transpose}), y_{b})\Big],
\end{equation}
where $\hat{y_{b}^{*}}$ is the network predictions optimized with the candidate loss function $\MetaLoss_{\phi}^{\Transpose}$. Importantly, optimization is performed directly to $\hat{y_{b}^{*}}$, as opposed to the network parameters $\theta$, thus omitting any base network computation (\textit{i.e.} no training of the base network).
\begin{equation}
\hat{y_{b}^{*}}(\MetaLoss_{\phi}^{\Transpose}) = \arg \min_{\hat{y_{b}^{*}}} \MetaLoss_{\phi}^{\Transpose}(\hat{y_{b}^{*}}, y_{b})
\end{equation}
A large positive correlation indicates that minimizing $\MetaLoss_{\phi}^{\Transpose}$ corresponds to minimizing the given performance metric $\Loss_{\mathcal{P}}$ (assuming both $\MetaLoss_{\phi}^{\Transpose}$ and $\mathcal{L}_{\mathcal{P}}$ are for minimization). In contrast to this, if $g\leq0$, then $\MetaLoss_{\phi}^{\Transpose}$ is regarded as being unpromising and should be assigned the worst-case fitness and not evaluated. The underlying assumption here is that if a loss function cannot directly optimize the labels, it is unlikely to be able to optimize the labels through the model weights $\theta$ successfully.

\subsubsection{Pre-Evaluation Filter - Gradient Equivalence}

Many of the candidate loss functions found in the later generations, as convergence is approached, have near-identical gradient behavior (\textit{i.e.} functionally equivalence). To address this, the gradient equivalence checking strategy from \cite{li2022autoloss} is employed as another filter to identify loss functions that have near-identical behavior to those seen previously. Using the prediction from previously, the gradient norms are computed.
\begin{equation}
\{\parallel\nabla_{\hat{y}_{b}}\MetaLoss_{\phi}^{\Transpose}\parallel_{2}\}_{b=1}^{B}
\end{equation}
If, for all of the $B$ samples, two-loss functions have the same gradient norms within two significant digits, they are considered functionally equivalent, and their fitness is cached.

\subsubsection{Partial Training Sessions}

For the remaining loss functions whose fitness evaluation cannot be fully obviated, we compute the fitness $\Fitness$ using a truncated number of gradient steps $\mathcal{S}_{testing}=500$. As noted in \cite{grefenstette1985genetic, jin2011surrogate}, performance at the beginning of training is correlated with the performance at the end of training; consequently, we can obtain an estimate of what $\Fitness$ would be by performing a partial training session of the base model. Preliminary experiments with EvoMAL showed minimal short-horizon bias \cite{wu2018understanding}, and the ablation study found in \cite{gonzalez2021optimizing} indicated that $500$ gradient steps during loss function evaluation is a good trade-off between final base-inference performance and meta-training time. In addition to significantly reducing the run-time of EvoMAL, reducing the value of $\mathcal{S}_{testing}$ has the effect of implicitly optimizing for the base-training convergence and sample-efficiency, as mentioned in \cite{hospedales2020meta}.

\section{Experimental Setup}
\label{sec:experiment-setup}

In this section, the performance of EvoMAL is evaluated. A wide range of experiments are conducted across seven datasets and numerous popular network architectures, with the performance contrasted against a representative set of benchmark methods implemented in DEAP \cite{fortin2012deap}, PyTorch \cite{paszke2019pytorch} and Higher \cite{grefenstette2019generalized}. The code for our experiments can be found at the following: \href{https://github.com/Decadz/Evolved-Model-Agnostic-Loss}{https://github.com/Decadz/Evolved-Model-Agnostic-Loss}

\subsection{Benchmark Methods}

\begin{table*}[t!]
\centering
\captionsetup{justification=centering}
\caption{Classification results reporting the mean $\pm$ standard deviation final inference error rate across 5 independent executions of each algorithm on each task + model pair. Loss functions are directly meta-learned and applied to the same respective task.}
\resizebox{0.9\textwidth}{!}{
\begin{threeparttable}

\begin{tabular}{lccccc}
\hline
\noalign{\vskip 1mm}
Task and Model          & Baseline          & ML$^3$            & TaylorGLO         & GP-LFL            & EvoMAL (Ours)                     \\ \hline \noalign{\vskip 1mm}
\textbf{MNIST}          &                   &                   &                   &                   &                                   \\
Logistic \tnote{1}      & 0.0787$\pm$0.0009 & 0.0768$\pm$0.0061 & 0.0725$\pm$0.0013 & 0.0781$\pm$0.0052 & \textbf{0.0721$\pm$0.0017}        \\ 
MLP \tnote{2}           & 0.0247$\pm$0.0005 & 0.0201$\pm$0.0081 & \textbf{0.0151$\pm$0.0013} & 0.0156$\pm$0.0012 & 0.0152$\pm$0.0012        \\ 
LeNet-5 \tnote{3}       & 0.0203$\pm$0.0025 & 0.0135$\pm$0.0039 & 0.0137$\pm$0.0038 & 0.0115$\pm$0.0015 & \textbf{0.0100$\pm$0.0010}        \\ \hline \noalign{\vskip 1mm} 
\textbf{CIFAR-10}       &                   &                   &                   &                   &                                   \\ 
AlexNet \tnote{4}       & 0.1544$\pm$0.0012 & 0.1450$\pm$0.0028 & 0.1499$\pm$0.0075 & 0.1506$\pm$0.0047 & \textbf{0.1437$\pm$0.0033}        \\
VGG-16 \tnote{5}        & 0.0771$\pm$0.0023 & 0.0700$\pm$0.0006 & 0.0700$\pm$0.0022 & \textbf{0.0686$\pm$0.0014} & 0.0687$\pm$0.0016        \\
AllCNN-C \tnote{6}      & 0.0761$\pm$0.0015 & 0.0712$\pm$0.0043 & 0.0735$\pm$0.0030 & 0.0701$\pm$0.0022 & \textbf{0.0697$\pm$0.0010}        \\
ResNet-18 \tnote{7}     & 0.0658$\pm$0.0019 & 0.0584$\pm$0.0022 & 0.0546$\pm$0.0033 & 0.0818$\pm$0.0391 & \textbf{0.0528$\pm$0.0015}        \\
PreResNet \tnote{8}     & 0.0661$\pm$0.0015 & 0.0660$\pm$0.0016 & 0.0660$\pm$0.0027 & 0.0658$\pm$0.0023 & \textbf{0.0655$\pm$0.0018}        \\
WideResNet \tnote{9}    & 0.0548$\pm$0.0016 & 0.0549$\pm$0.0040 & 0.0493$\pm$0.0023 & 0.0489$\pm$0.0014 & \textbf{0.0484$\pm$0.0018}        \\
SqueezeNet \tnote{10}   & 0.0838$\pm$0.0013 & 0.0800$\pm$0.0012 & 0.0800$\pm$0.0025 & 0.0810$\pm$0.0016 & \textbf{0.0796$\pm$0.0017}        \\ \noalign{\vskip 1mm} \hline \noalign{\vskip 1mm} 
\textbf{CIFAR-100}      &                   &                   &                   &                   &                                   \\
WideResNet \tnote{9}    & 0.2293$\pm$0.0017 & 0.2299$\pm$0.0027 & 0.2347$\pm$0.0077 & 0.3382$\pm$0.1406 & \textbf{0.2276$\pm$0.0033}        \\
PyramidNet \tnote{11}   & \textbf{0.2527$\pm$0.0028} & 0.2792$\pm$0.0226 & 0.3064$\pm$0.0549 & 0.2747$\pm$0.0087 & 0.2664$\pm$0.0063        \\ \noalign{\vskip 1mm} \hline \noalign{\vskip 1mm} 
\textbf{SVHN}           &                   &                   &                   &                   &                                   \\
WideResNet \tnote{9}    & 0.0340$\pm$0.0005 & 0.0335$\pm$0.0003 & 0.0343$\pm$0.0016 & 0.0340$\pm$0.0015 & \textbf{0.0329$\pm$0.0013}        \\ \noalign{\vskip 1mm} \hline
\end{tabular}

\begin{tablenotes}\centering
\tiny Network architecture references: \item[1] McCullagh et al. (2019) \item[2] Baydin et al. (2018) \item[3] LeCun et al. (1998) \item[4] Krizhevsky et al. (2012) \item[5] Simonyan and Zisserman (2015) \item[6] Springenberg et al. (2015) \item[7] He et al. (2015)  \item[8] He et al. (2016) \item[9] Zagoruyko and Komodakis (2016) \item[10] N. Iandola et al. (2016)  \item[11] Han et al. (2017)
\end{tablenotes}

\end{threeparttable}
}
\vspace{-3mm}\label{table:meta-testing-results-classification}
\end{table*}

The selection of benchmark methods is intended to showcase the performance of the newly proposed algorithm against the current state-of-the-art. Additionally, the selected methods enable direct comparison between EvoMAL and its derivative methods, which aids in validating the effectiveness of hybridizing the approaches into one unified framework.
\begin{itemize}[leftmargin=*]

  \item \textbf{Baseline} -- Directly using $\Loss_{\Task}$ as the loss function, \textit{i.e.} using the squared error loss (regression) or cross-entropy loss (classification) and a prototypical training loop (\textit{i.e.} no meta-learning). 
  
  \item \textbf{ML$^3$ Supervised} -- Gradient-based method proposed in \cite{bechtle2021meta}, which uses a parametric loss function defined by a two hidden layer feed-forward network trained with unrolled differentiation, \textit{i.e.} the method shown in Section \ref{sec:optimization}.
  
  \item \textbf{TaylorGLO} -- Evolution-based method proposed by \cite{gonzalez2021optimizing}, which uses a third-order Taylor-polynomial representation for the meta-learned loss functions, optimized via covariance matrix adaptation evolution strategy.
  
  \item \textbf{GP-LFL} -- A proxy method used to aggregate previous GP-based approaches \textit{without} any local-search mechanisms, using an identical setup to EvoMAL, and excluding Section \ref{sec:optimization}.
  
\end{itemize}
\noindent
Where possible, hyper-parameter selection has been standardized across the benchmark methods to allow for a fair comparison. For example, in TaylorGLO, GP-LFL, and EvoMAL we use an identical population size $=25$ and number of generations = $50$. For unique hyper-parameters, the suggested values from the respective publications are utilized.

\vspace{-3mm}\subsection{Benchmark Problems}

Regarding the problem domains, seven datasets have been selected. Three tabulated regression tasks are initially used: Diabetes, Boston Housing, and California Housing, all taken from the UCI's dataset repository \cite{asuncion2007uci}. Following this, analogous to the prior literature \cite{gonzalez2020improved, gonzalez2021optimizing, bechtle2021meta}, both MNIST \cite{lecun1998gradient} and CIFAR-10 \cite{krizhevsky2009learning} are employed to evaluate the benchmark methods. Finally, experiments are conducted on the more challenging but related domains of SVHN \cite{netzer2011reading} and CIFAR-100 \cite{krizhevsky2009learning}, respectively.

For the three tabulated regression tasks, the datasets are partitioned 60:20:20 for training, validation, and testing. Furthermore, to improve the training dynamics, both the features and labels are normalized. For the remaining datasets, the original training-testing partitioning is used, with 10\% of the training instances allocated for validation. In addition, data augmentation techniques consisting of normalization, random horizontal flips, and cropping are applied to the training data during meta and base training.

Regarding the base models, a diverse set of neural network architectures are utilized to evaluate the selected benchmark methods. For Diabetes, Boston Housing, and California Housing, a simple Multi-Layer Perceptron (MLP) taken from \cite{baydin2018hypergradient} with 1000 hidden nodes and ReLU activations are employed. For MNIST, Logistic Regression \cite{mccullagh2019generalized}, MLP, and the well-known LeNet-5 architecture \cite{lecun1998gradient} are used. While on CIFAR-10 AlexNet \cite{krizhevsky2012imagenet}, VGG-16 \cite{simonyan2014very}, AllCNN-C \cite{springenberg2014striving}, ResNet-18 \cite{he2016deep}, Preactivation ResNet-101 \cite{he2016identity}, WideResNet 28-10 \cite{zagoruyko2016wide} and SqueezeNet \cite{iandola2016squeezenet} are used. For the remaining datasets, WideResNet 28-10 is again used, as well as PyramidNet \cite{han2017deep} on CIFAR-100. All models are trained using stochastic gradient descent (SGD) with momentum. The model hyper-parameters are selected using their respective values from the literature in an identical setup to \cite{gonzalez2021optimizing, bechtle2021meta}.

Finally, due to the stochastic nature of the benchmark methods, we perform five independent executions of each method on each dataset + model pair. Furthermore, we control for the base initializations such that each method gets identical initial conditions across the same random seed; thus, any difference in variance between the methods can be attributed to the respective algorithms.

\begin{table*}[t!]
\centering
\captionsetup{justification=centering}
\caption{Regression results reporting the mean $\pm$ standard deviation final inference testing mean squared error across 5 executions of each algorithm on each task + model pair. Loss functions are directly meta-learned and applied to the same respective task.}
\resizebox{0.9\textwidth}{!}{
\begin{threeparttable}

\begin{tabular}{lccccc}
\hline
\noalign{\vskip 1mm}
Task and Model          & Baseline          & ML$^3$            & TaylorGLO         & GP-LFL            & EvoMAL (Ours)                     \\ \hline \noalign{\vskip 1mm}
\textbf{Diabetes}       &                   &                   &                   &                   &                                   \\
MLP      \tnote{1}      & 0.3829$\pm$0.0065 & 0.4278$\pm$0.1080 & 0.3620$\pm$0.0096 & 0.3746$\pm$0.0577 & \textbf{0.3573$\pm$0.0198}        \\ \hline \noalign{\vskip 1mm} 
\textbf{Boston}         &                   &                   &                   &                   &                                   \\
MLP \tnote{1}           & \textbf{0.1304$\pm$0.0057} & 0.2474$\pm$0.0526 & 0.1349$\pm$0.0074 & 0.1341$\pm$0.0082 & 0.1314$\pm$0.0131        \\ \hline \noalign{\vskip 1mm} 
\textbf{California}     &                   &                   &                   &                   &                                   \\
MLP \tnote{1}           & 0.2326$\pm$0.0019 & 0.2438$\pm$0.0189 & 0.1794$\pm$0.0156 & 0.2071$\pm$0.0609 & \textbf{0.1723$\pm$0.0048}        \\ \hline \noalign{\vskip 1mm}
\end{tabular}

\begin{tablenotes}\centering
\tiny Network architecture references: \item[1] Baydin et al. (2018)
\end{tablenotes}

\end{threeparttable}
}
\label{table:meta-testing-results-regression}
\end{table*}

\begin{table*}[t!]
\centering
\captionsetup{justification=centering}
\caption{Loss function transfer results reporting the mean $\pm$ standard deviation final inference testing error rate across 5 independent executions of each algorithm on each task + model pair. The best performance and those within its standard deviation are bolded. Loss functions are meta-learned on CIFAR-10 with the respective model, and then transferred to CIFAR-100 using that same model.}
\resizebox{0.9\textwidth}{!}{
\begin{threeparttable}

\begin{tabular}{lccccc}
\hline
\noalign{\vskip 1mm}
Task and Model          & Baseline          & ML$^3$            & TaylorGLO         & GP-LFL            & EvoMAL (Ours)                 \\ \hline \noalign{\vskip 1mm}
\textbf{CIFAR-100}      &                   &                   &                   &                   &                               \\
AlexNet \tnote{1}       & \textbf{0.5262$\pm$0.0094} & 0.7735$\pm$0.0295 & 0.5543$\pm$0.0138 & 0.5329$\pm$0.0037 & 0.5324$\pm$0.0031             \\
VGG-16 \tnote{2}        & \textbf{0.3025$\pm$0.0022} & 0.3171$\pm$0.0019 & 0.3155$\pm$0.0021 & 0.3171$\pm$0.0041 & 0.3115$\pm$0.0038             \\
AllCNN-C \tnote{3}      & 0.2830$\pm$0.0021 & 0.2817$\pm$0.0032 & 0.4191$\pm$0.0058 & 0.2849$\pm$0.0012 & \textbf{0.2807$\pm$0.0028}             \\
ResNet-18 \tnote{4}     & 0.2474$\pm$0.0018 & 0.6000$\pm$0.0173 & 0.2436$\pm$0.0032 & 0.2373$\pm$0.0013 & \textbf{0.2326$\pm$0.0014}             \\
PreResNet \tnote{5}     & 0.2908$\pm$0.0065 & \textbf{0.2838$\pm$0.0019} & 0.2993$\pm$0.0030 & 0.2839$\pm$0.0025 & 0.2899$\pm$0.0024             \\
WideResNet \tnote{6}    & 0.2293$\pm$0.0017 & 0.2448$\pm$0.0063 & 0.2285$\pm$0.0031 & 0.2276$\pm$0.0028 & \textbf{0.2238$\pm$0.0017}             \\
SqueezeNet \tnote{7}    & 0.3178$\pm$0.0015 & 0.3402$\pm$0.0057 & 0.3178$\pm$0.0012 & 0.3343$\pm$0.0054 & \textbf{0.3166$\pm$0.0026}             \\ \noalign{\vskip 1mm} \hline \noalign{\vskip 1mm} 
\end{tabular}

\begin{tablenotes}\centering
\tiny Network architecture references: \item[1] Krizhevsky et al. (2012) \item[2] Simonyan and Zisserman (2015) \item[3] Springenberg et al. (2015) \item[4] He et al. (2015) \item[5] He et al. (2016) \item[6] Zagoruyko and Komodakis (2016) \item[7] N. Iandola et al. (2016)
\end{tablenotes}
\vspace{-3mm}
\end{threeparttable}
}
\label{table:meta-testing-transfer}
\end{table*}

\vspace{-2mm}\section{Results and Analysis}
\label{sec:results}

The results and analysis are approached from three distinct angles. First, the experimental results reporting the final inference meta-testing performance when using meta-learned loss functions for base-training are presented in Section \ref{sec:meta-testing}. Following this, the performance of the loss function learning algorithms themselves at meta-training time is compared in Section \ref{sec:meta-training}. The focus is turned towards an analysis of the meta-learned loss functions themselves to highlight some of the loss functions developed by EvoMAL on both classification and regression tasks in Section \ref{sec:loss-functions}. Finally, an analysis is given in Sections \ref{sec:loss-landscapes} and \ref{sec:implicit-tuning}, which explore two of the central hypotheses for why meta-learned loss functions are so performant compared to their handcrafted counterparts.

\vspace{-3mm}\subsection{Meta-Testing Performance}
\label{sec:meta-testing}

A summary of the final inference testing results reporting the average error rate (classification) or mean squared error (regression) across the seven tested datasets is shown in Tables \ref{table:meta-testing-results-classification} and \ref{table:meta-testing-results-regression} respectively, where the same dataset and model pair are used for both meta-training and meta-testing. The results show that meta-learned loss functions consistently produce superior performance compared to the baseline handcrafted squared error loss and cross-entropy loss. Large performance gains are made on the California Housing, MNIST, and CIFAR-10 datasets, while more modest gains are observed on Diabetes, WideResNet CIFAR-100 and SVHN, and worse performance on Boston Housing MLP and CIFAR-100 PyramidNet, a similar finding to that found in \cite{gonzalez2021optimizing}. 

Contrasting the performance of EvoMAL to the benchmark loss function learning methods, it is shown that EvoMAL consistently meta-learns more performant loss functions, with better performance on all task + model pairs except for on CIFAR-10 VGG-16, where performance is comparable to the next best method. Furthermore, compared to its derivative methods ML$^3$ and GP-LFL, EvoMAL successfully meta-learns loss functions on more complex tasks, \textit{i.e.} CIFAR-100 and SVHN, whereas the other techniques often struggle to improve upon the baseline. These results empirically confirm the benefits of unifying existing approaches to loss function learning into one unified framework. Furthermore, the results clearly show the necessity for integrating local-search techniques in GP-based loss function learning.

Compared to prior research on loss function learning, our method exhibits a relatively smaller improvement when comparing the use of meta-learned loss functions to using the cross-entropy loss. For example, prior research has reported increasing the accuracy of a classification model by up to 5\% in some cases when using meta-learned loss functions compared to the cross-entropy loss. However, with heavily tuned baselines, optimizing for both $\alpha$ and $\mathcal{S}_{testing}$, such performance gains were very hard to obtain. This suggests that a proportion of the performance gains reported previously by loss function learning methods likely come from an implicit tuning effect on the training dynamics as opposed to a direct effect from using a different loss function. Implicit tuning is not a drawback of loss function learning as a paradigm; however, it is essential to disentangle the effects. In Section \ref{sec:loss-functions}, further experiments are given to isolate the effects.

Finally, although the performance gains are at times modest, it is impressive that performance gains can be made by a straightforward change to the selection of the loss function. Many previous developments in neural networks have added many millions of parameters or incorporated complex training routines to achieve similar improvements in performance \cite{liu2017survey,gu2018recent}. Furthermore, our results show that performance gains can be made upon well-tuned models which already employ a multitude of regularization techniques. This suggests that meta-learned loss functions can learn a distinct form of regularization that existing techniques such as data augmentation, dropout, weight decay, batch norm, skip layers, hyper-parameter optimization, etc., fail to capture.

\subsubsection{Loss Function Transfer}

\begin{figure*}[t!]

    \centering
    \captionsetup[subfigure]{justification=centering}
    \begin{subfigure}{0.17\textwidth}
        \centering
        \includegraphics[width=1\textwidth, height=3.25cm]{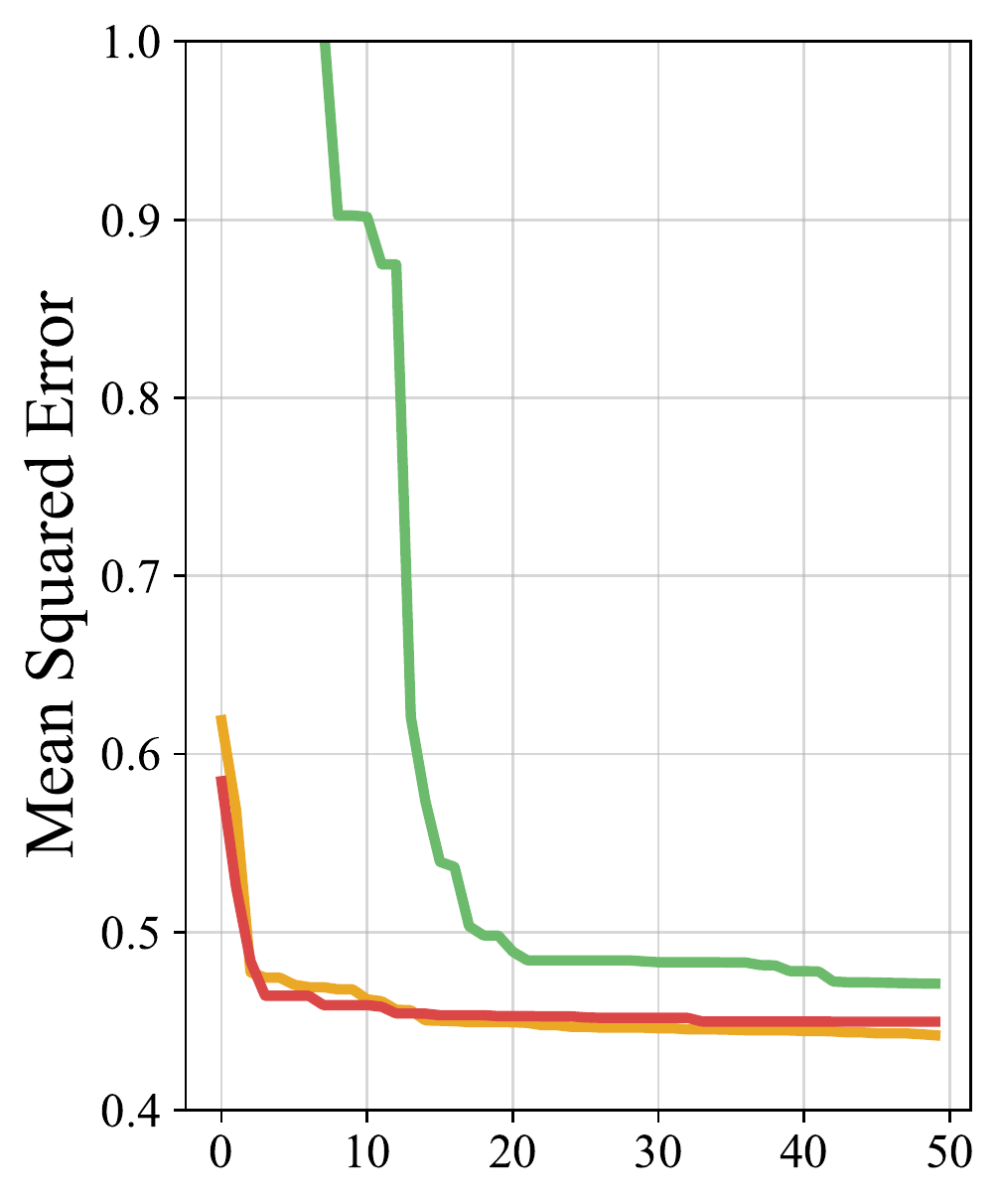}
        \caption{Diabetes\\MLP}
    \end{subfigure}%
    \hfill
    \begin{subfigure}{0.17\textwidth}
        \centering
        \includegraphics[width=1\textwidth, height=3.25cm]{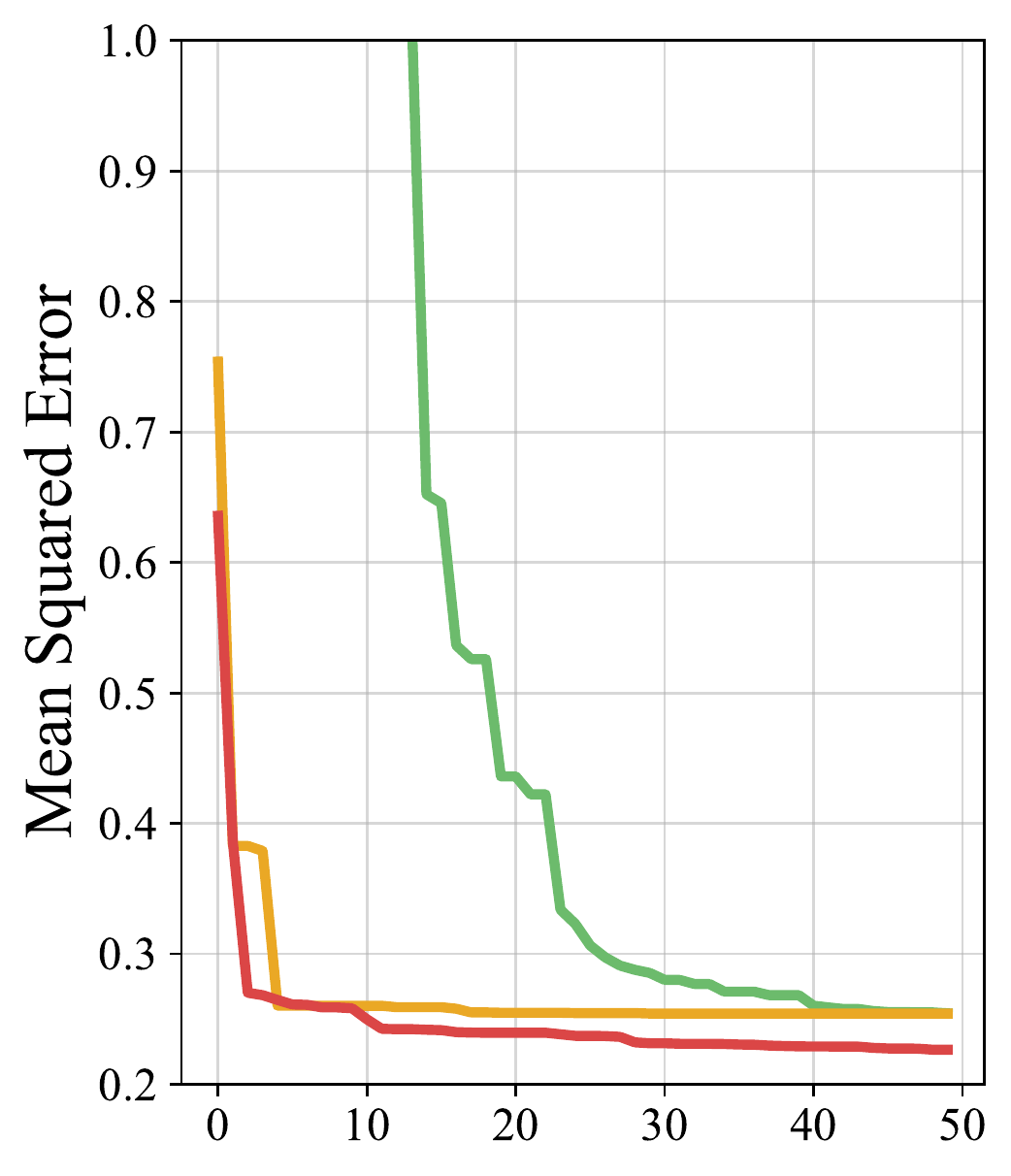}
        \caption{Boston\\MLP}
    \end{subfigure}%
    \hfill
    \begin{subfigure}{0.17\textwidth}
        \centering
        \includegraphics[width=1\textwidth, height=3.25cm]{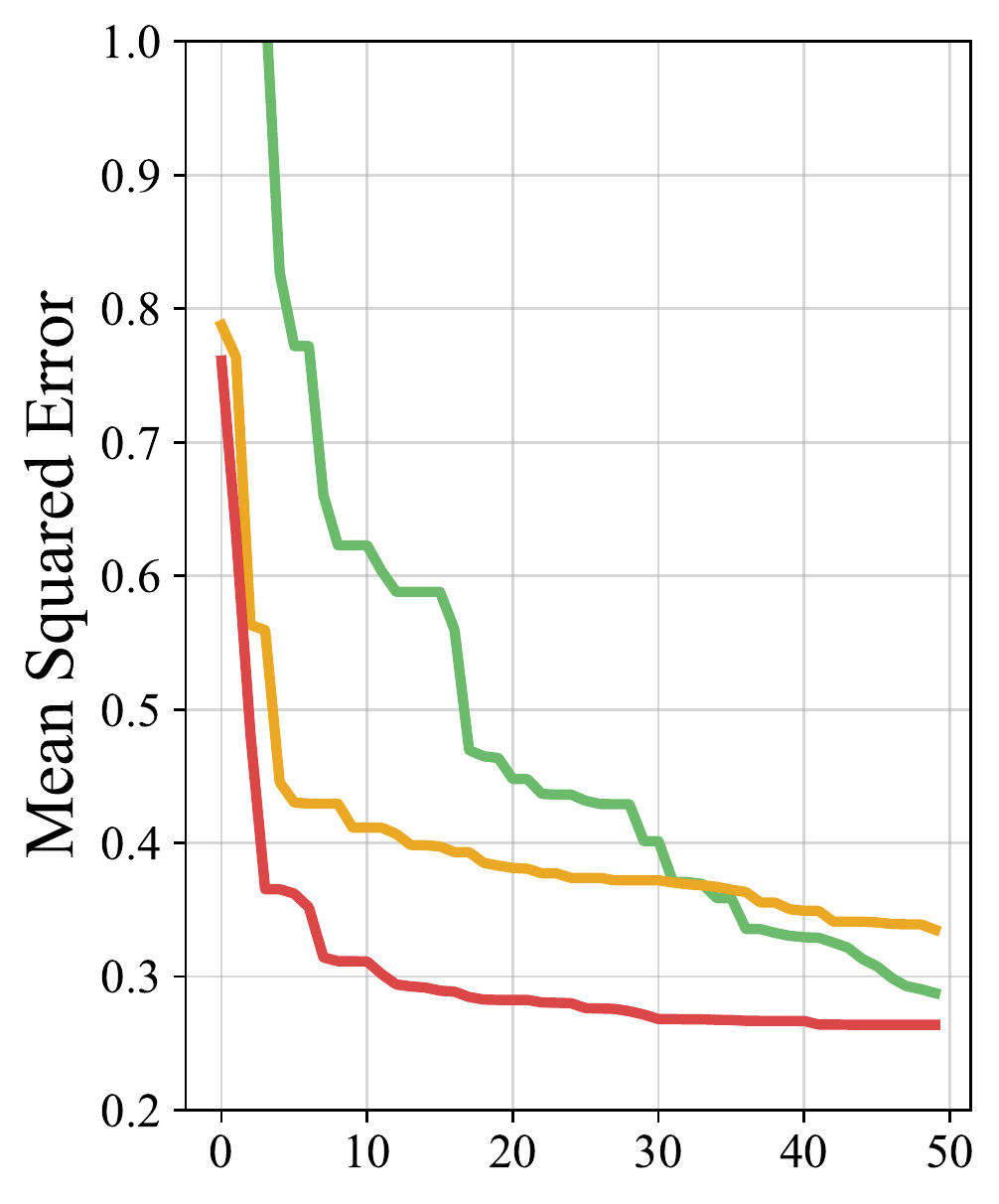}
        \caption{California\\MLP}
    \end{subfigure}%
    \hfill
    \begin{subfigure}{0.17\textwidth}
        \centering
        \includegraphics[width=1\textwidth, height=3.25cm]{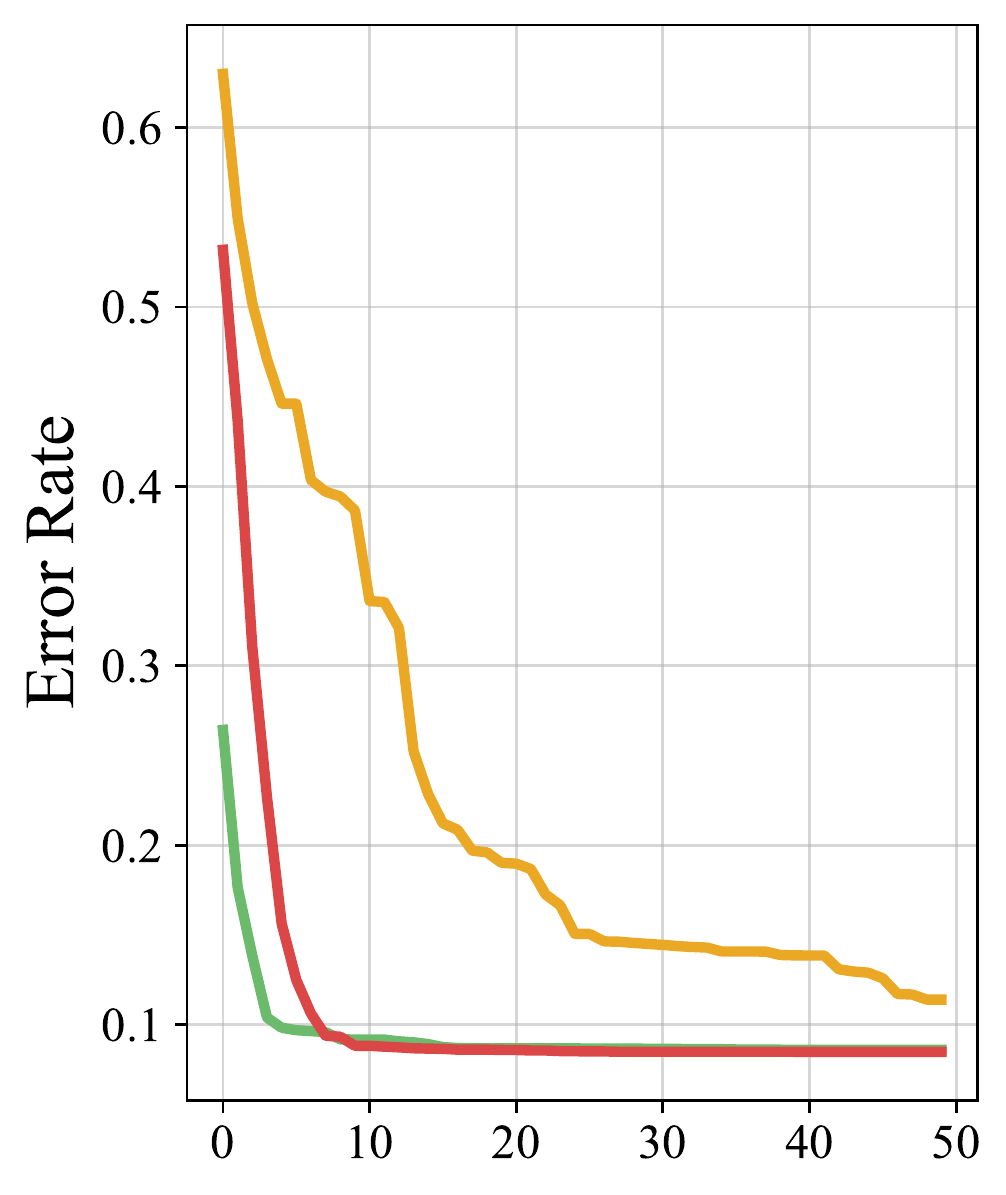}
        \caption{MNIST\\Logistic}
    \end{subfigure}%
    \hfill
    \begin{subfigure}{0.17\textwidth}
        \centering
        \includegraphics[width=1\textwidth, height=3.25cm]{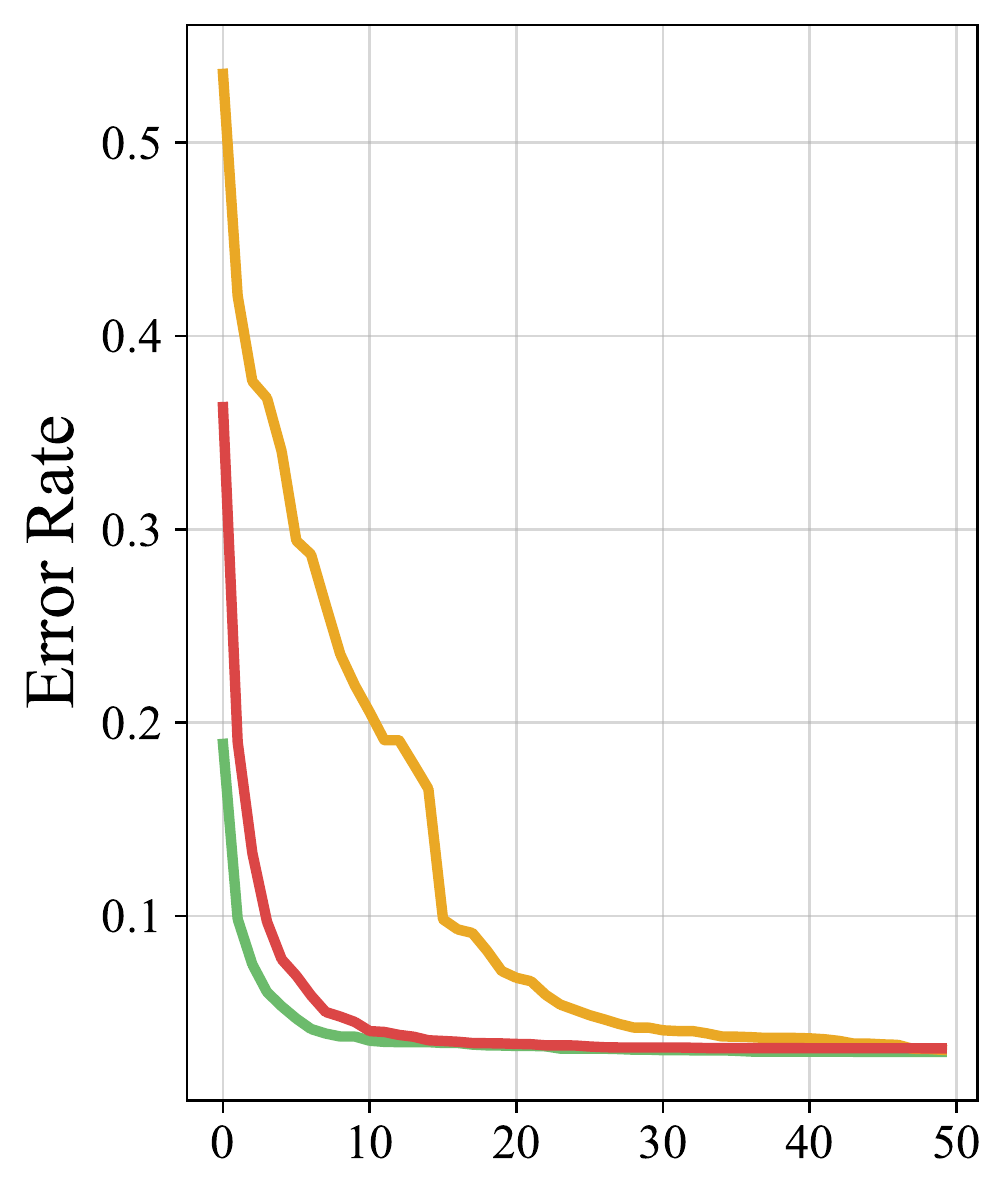}
        \caption{MNIST\\MLP}
    \end{subfigure}
    
    \vspace{3pt}

    \centering
    \captionsetup[subfigure]{justification=centering}
    \begin{subfigure}{0.17\textwidth}
        \centering
        \includegraphics[width=1\textwidth, height=3.25cm]{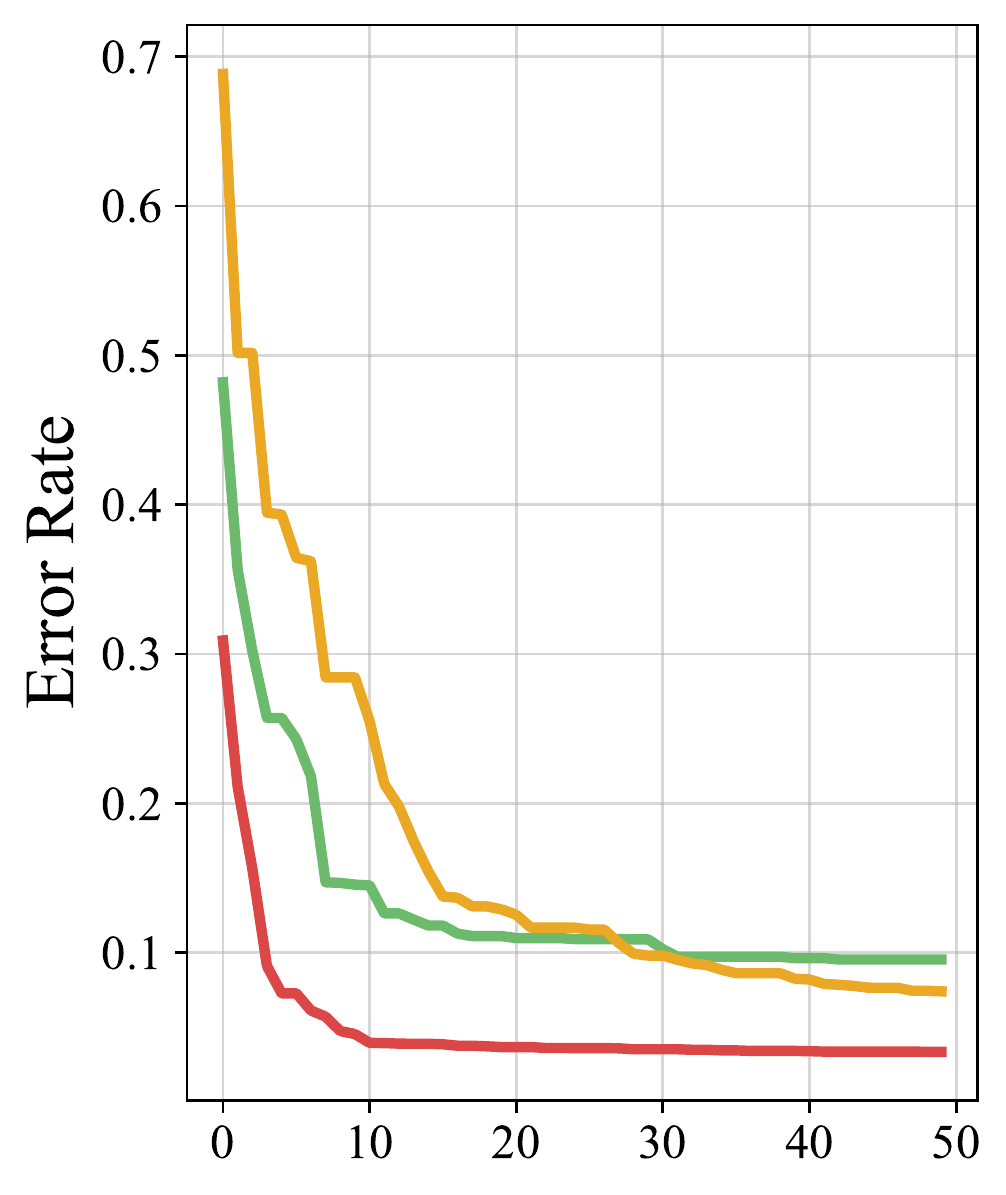}
        \caption{MNIST\\LeNet-5}
    \end{subfigure}%
    \hfill
    \begin{subfigure}{0.17\textwidth}
        \centering
        \includegraphics[width=1\textwidth, height=3.25cm]{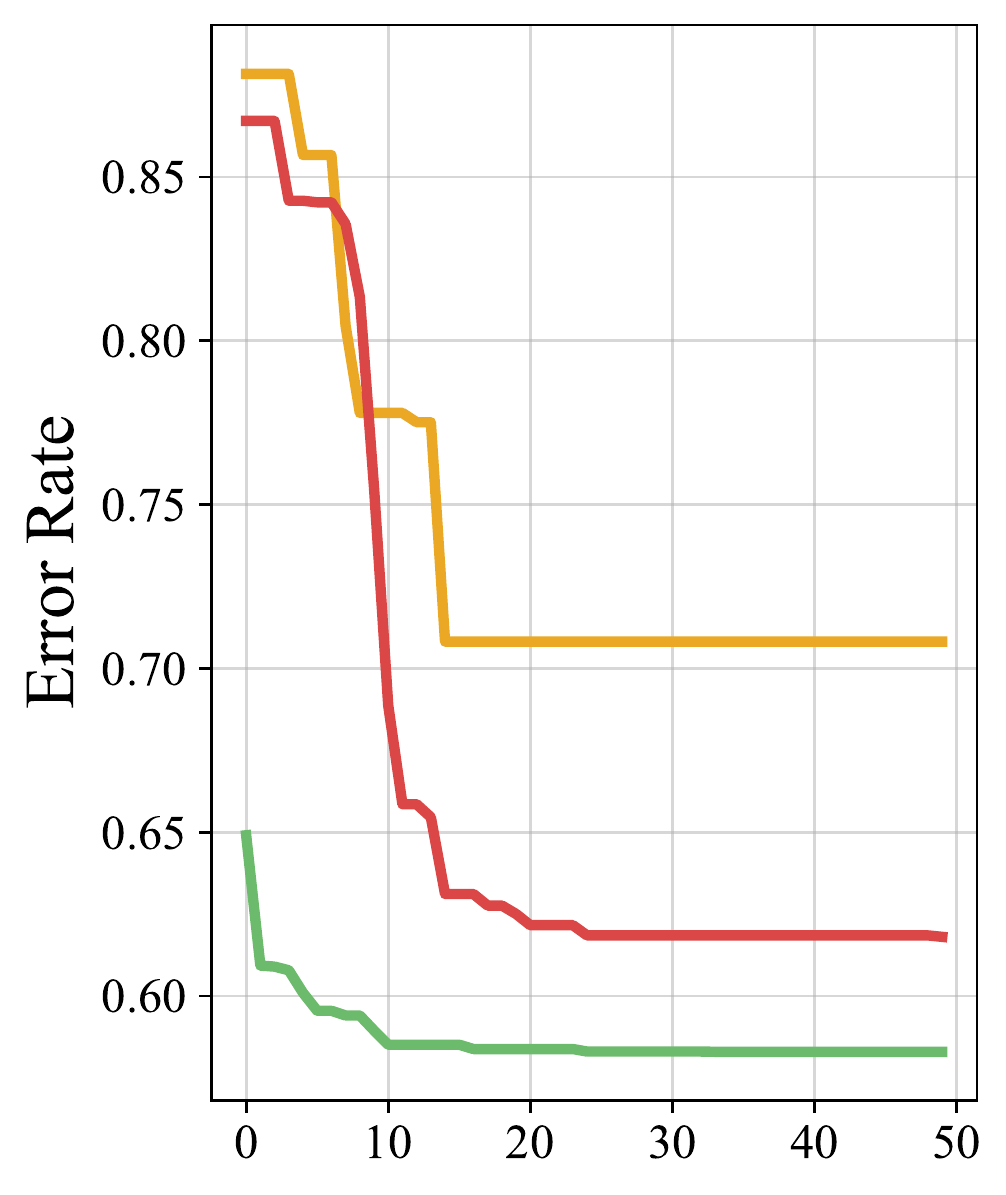}
        \caption{CIFAR-10\\AlexNet}
    \end{subfigure}%
    \hfill
    \begin{subfigure}{0.17\textwidth}
        \centering
        \includegraphics[width=1\textwidth, height=3.25cm]{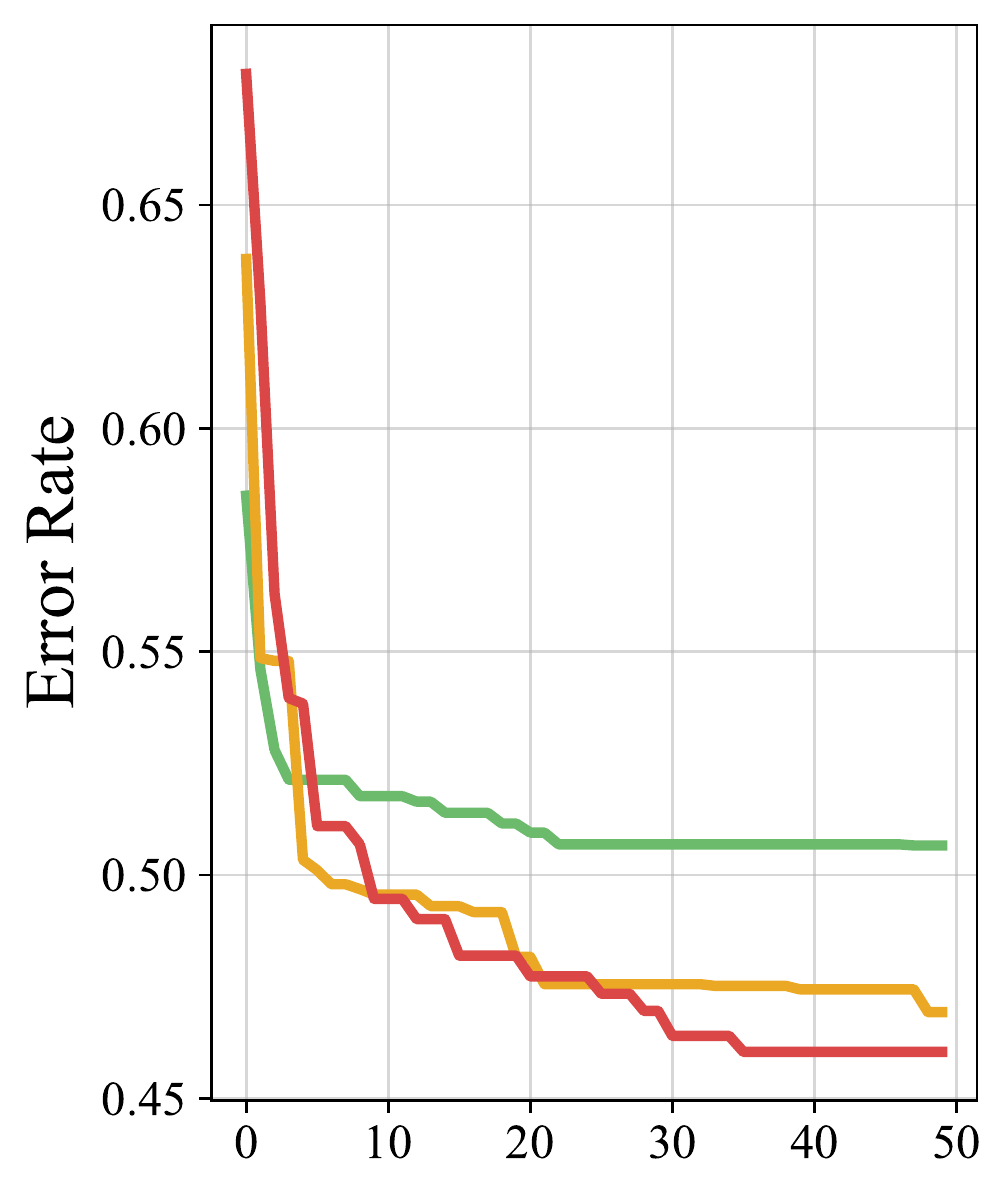}
        \caption{CIFAR-10\\VGG-16}
    \end{subfigure}%
    \hfill
    \begin{subfigure}{0.17\textwidth}
        \centering
        \includegraphics[width=1\textwidth, height=3.25cm]{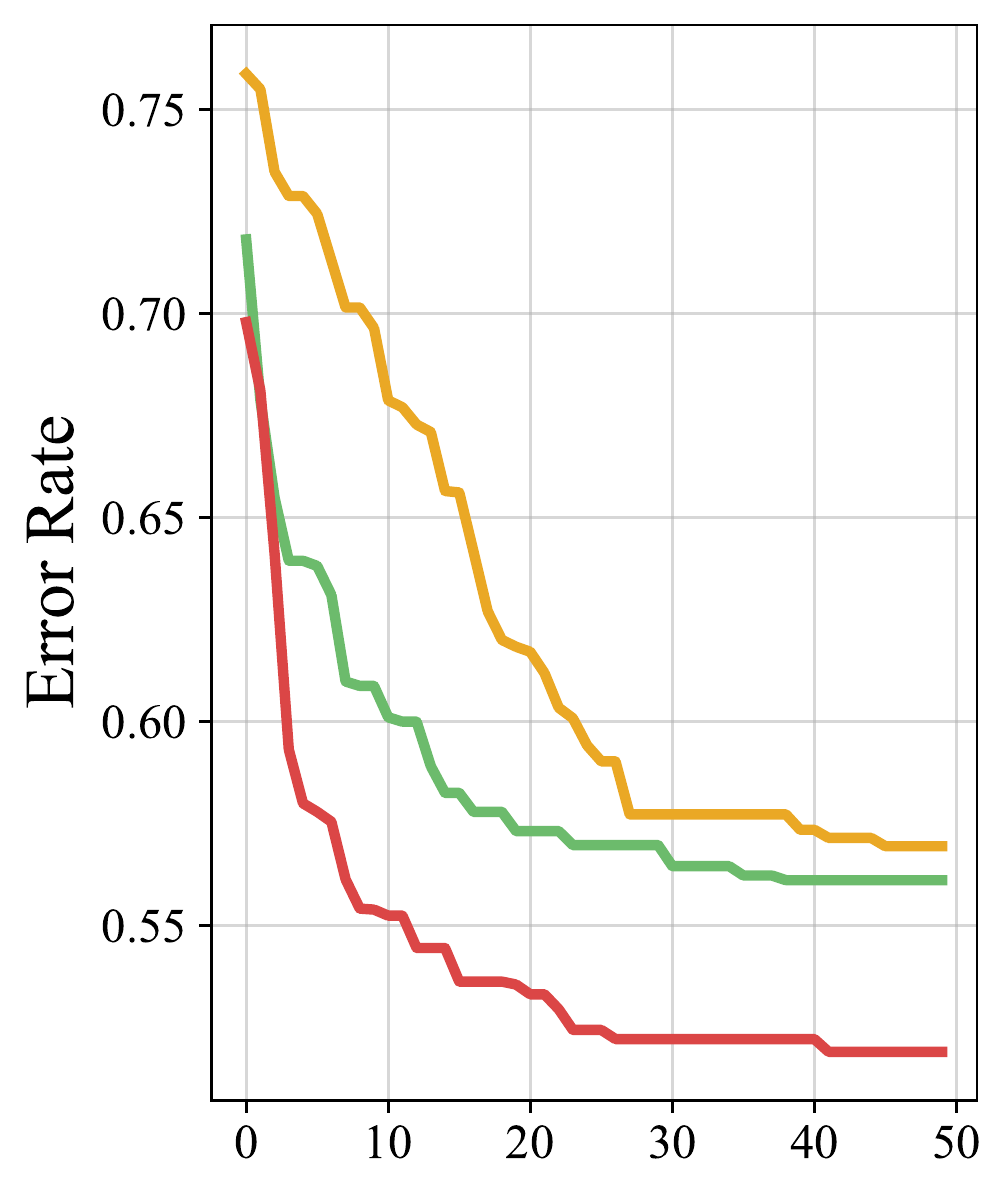}
        \caption{CIFAR-10\\AllCNN-C}
    \end{subfigure}%
    \hfill
    \begin{subfigure}{0.17\textwidth}
        \centering
        \includegraphics[width=1\textwidth, height=3.25cm]{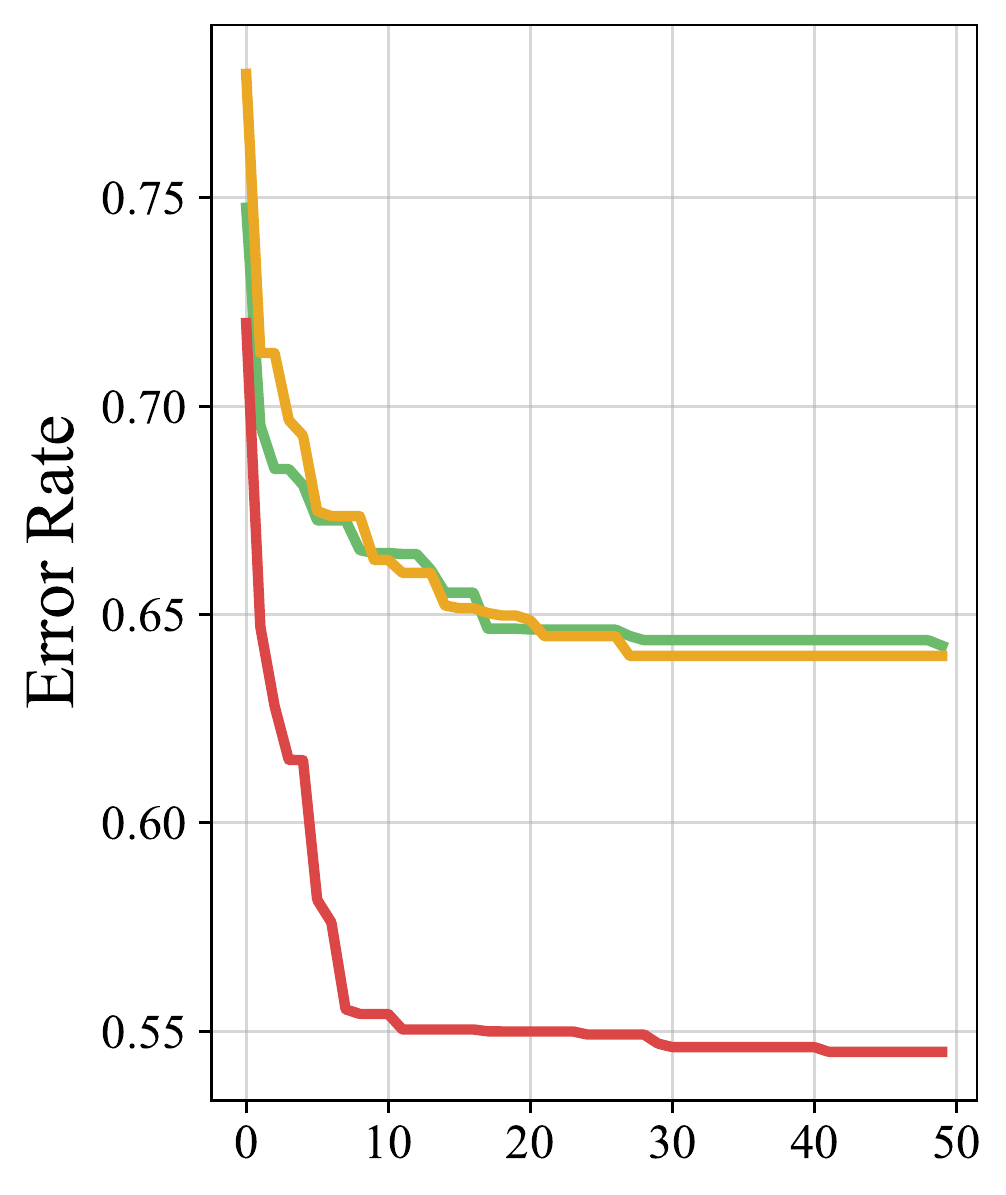}
        \caption{CIFAR-10\\ResNet-18}
    \end{subfigure}
    
    \vspace{3pt}
    
    \centering
    \begin{subfigure}{0.17\textwidth}
        \centering
        \includegraphics[width=1\textwidth, height=3.25cm]{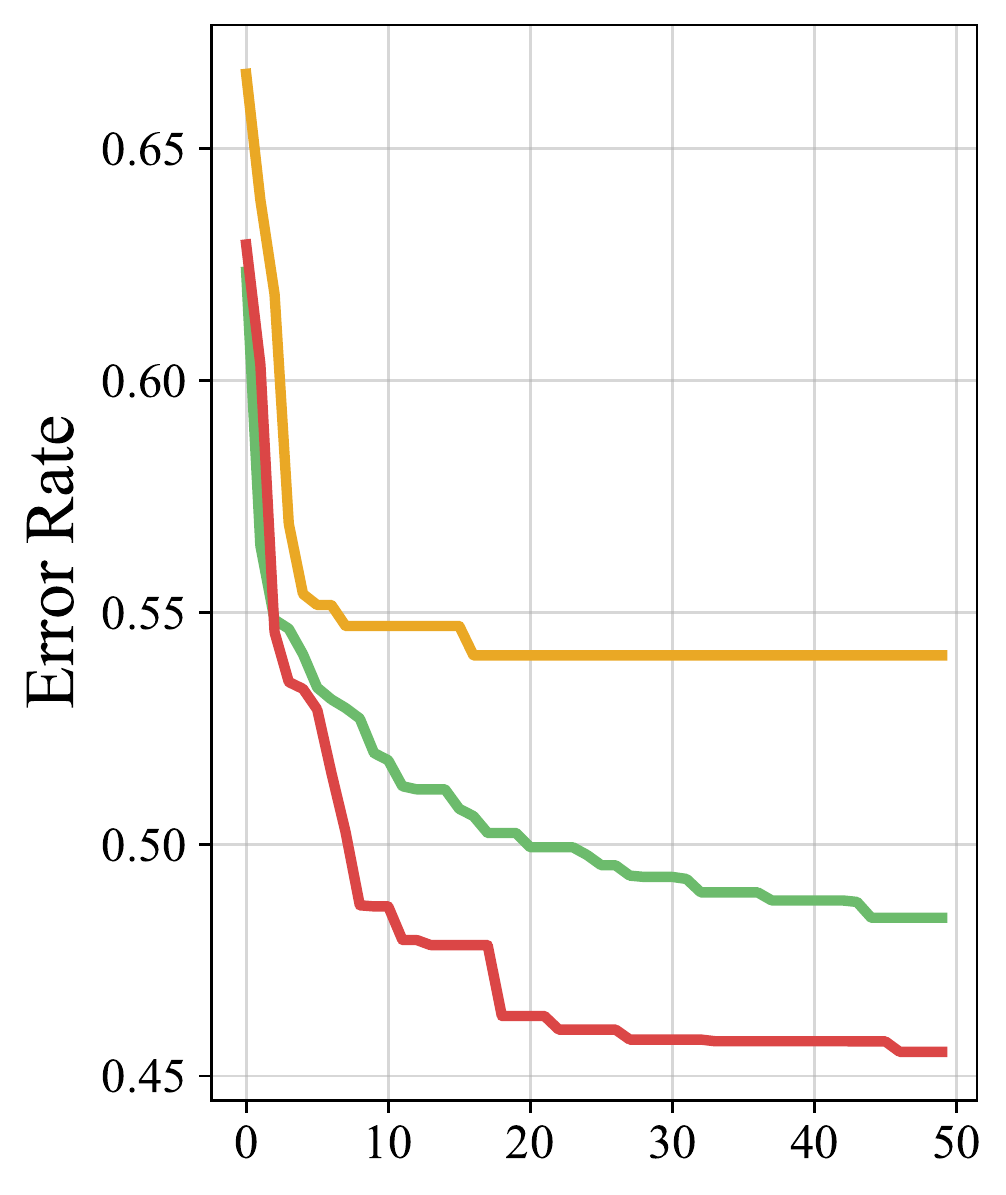}
        \caption{CIFAR-10\\WideResNet 28-10}
    \end{subfigure}%
    \hfill
    \begin{subfigure}{0.17\textwidth}
        \centering
        \includegraphics[width=1\textwidth, height=3.25cm]{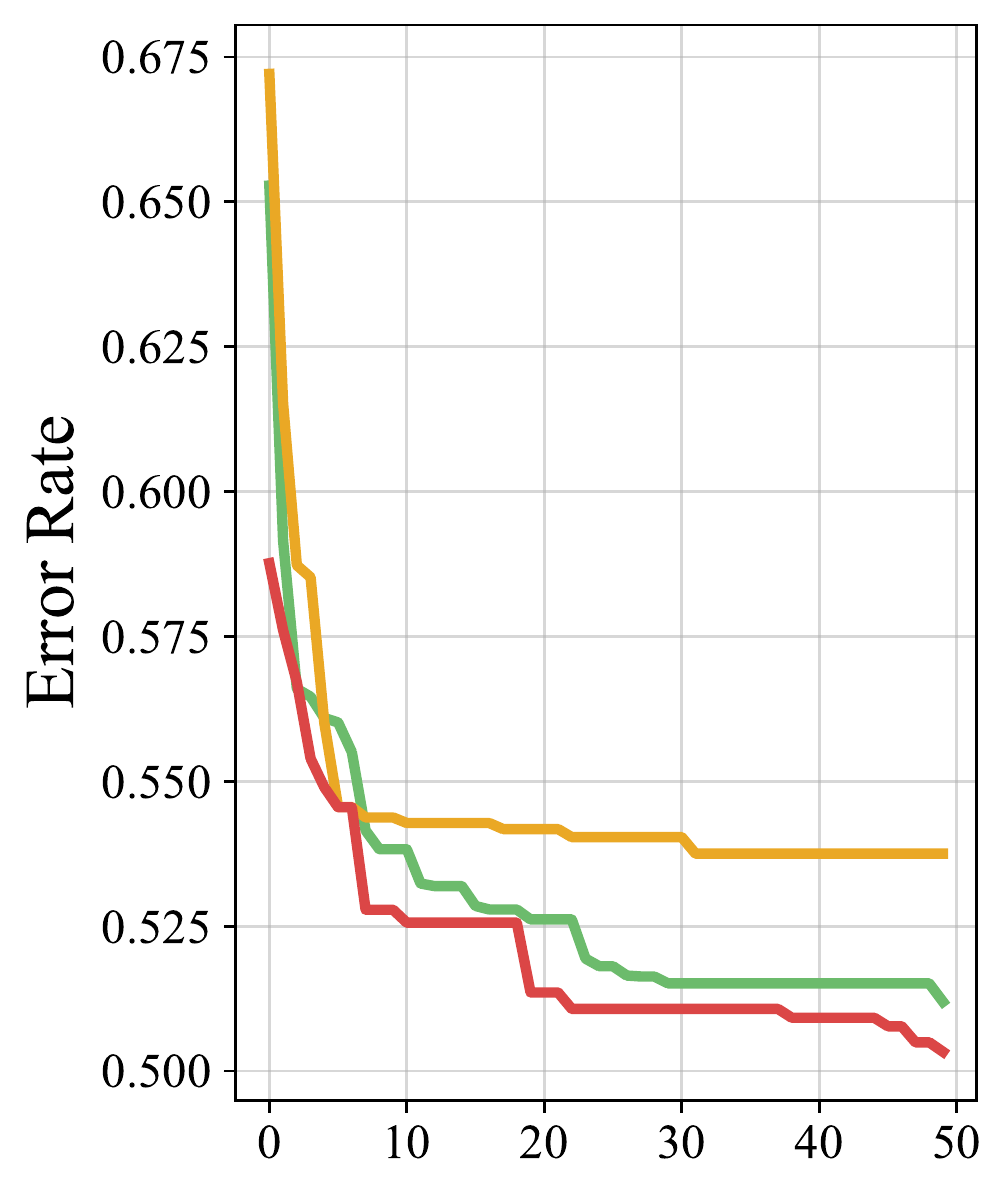}
        \caption{CIFAR-10\\SqueezeNet}
    \end{subfigure}%
    \hfill
    \begin{subfigure}{0.17\textwidth}
        \centering
        \includegraphics[width=1\textwidth, height=3.25cm]{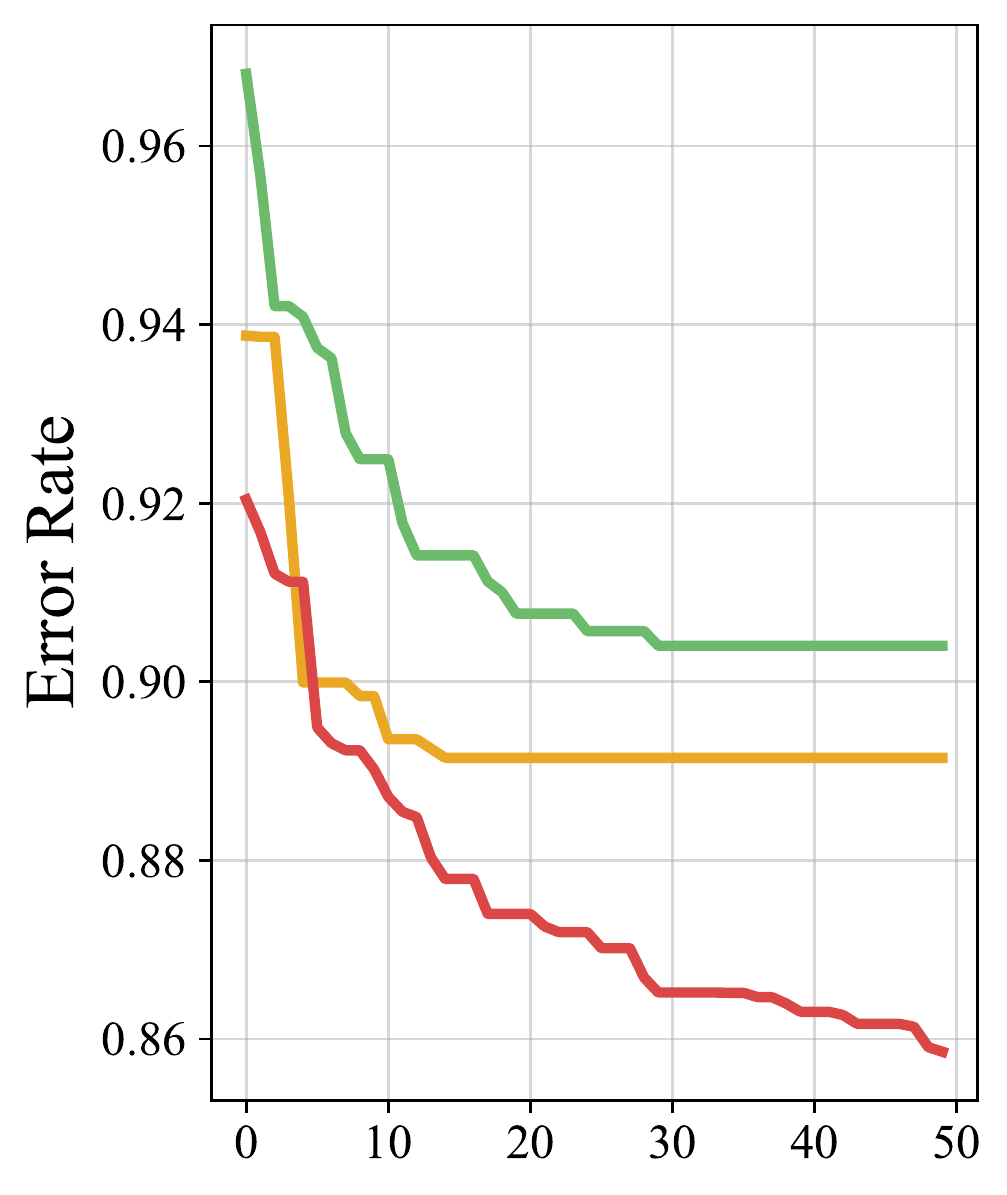}
        \caption{CIFAR-100\\WideResNet 28-10}
    \end{subfigure}%
    \hfill
    \begin{subfigure}{0.17\textwidth}
        \centering
        \includegraphics[width=1\textwidth, height=3.25cm]{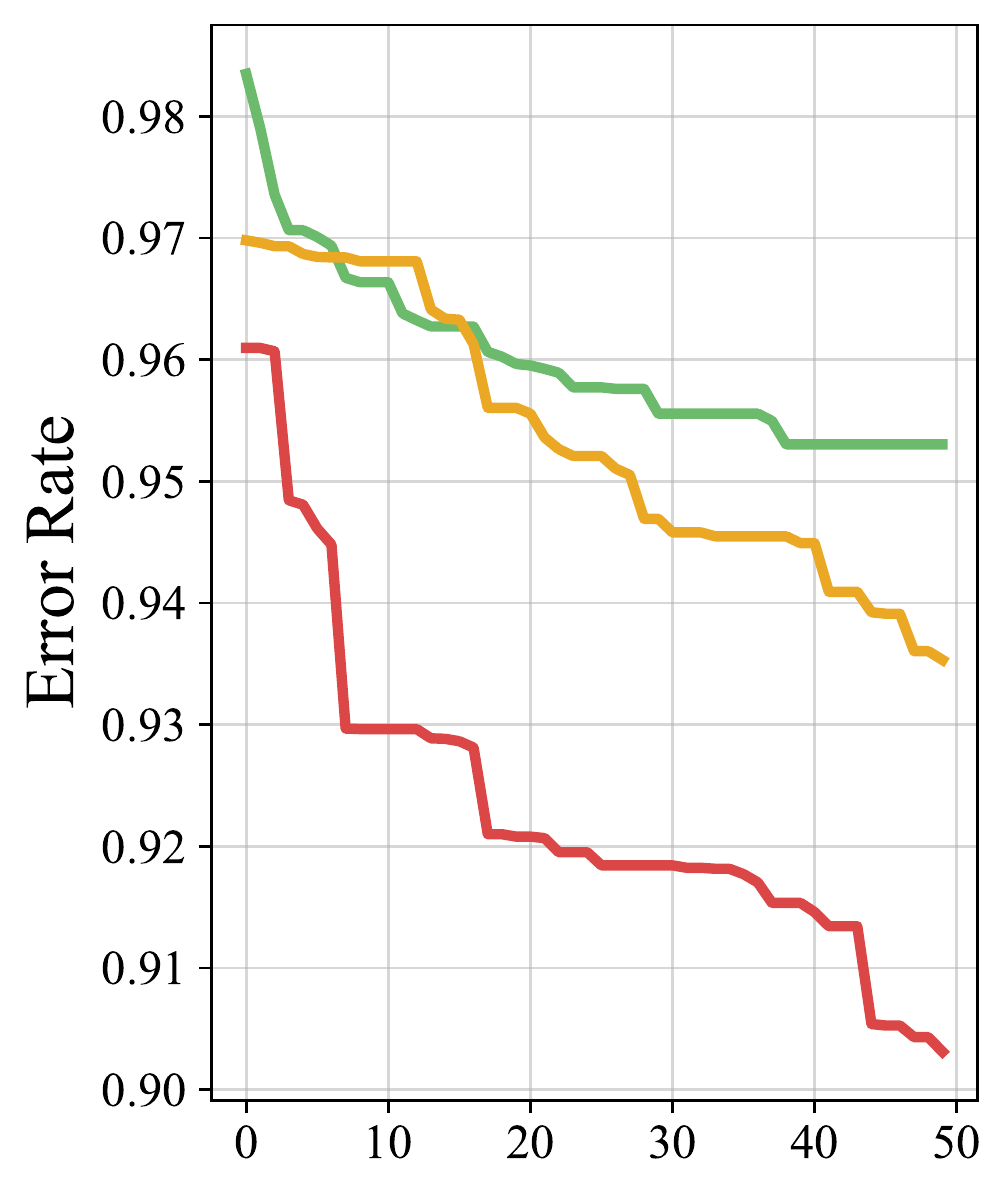}
        \caption{CIFAR-100\\PyramidNet}
    \end{subfigure}%
    \hfill
    \begin{subfigure}{0.17\textwidth}
        \centering
        \includegraphics[width=1\textwidth, height=3.25cm]{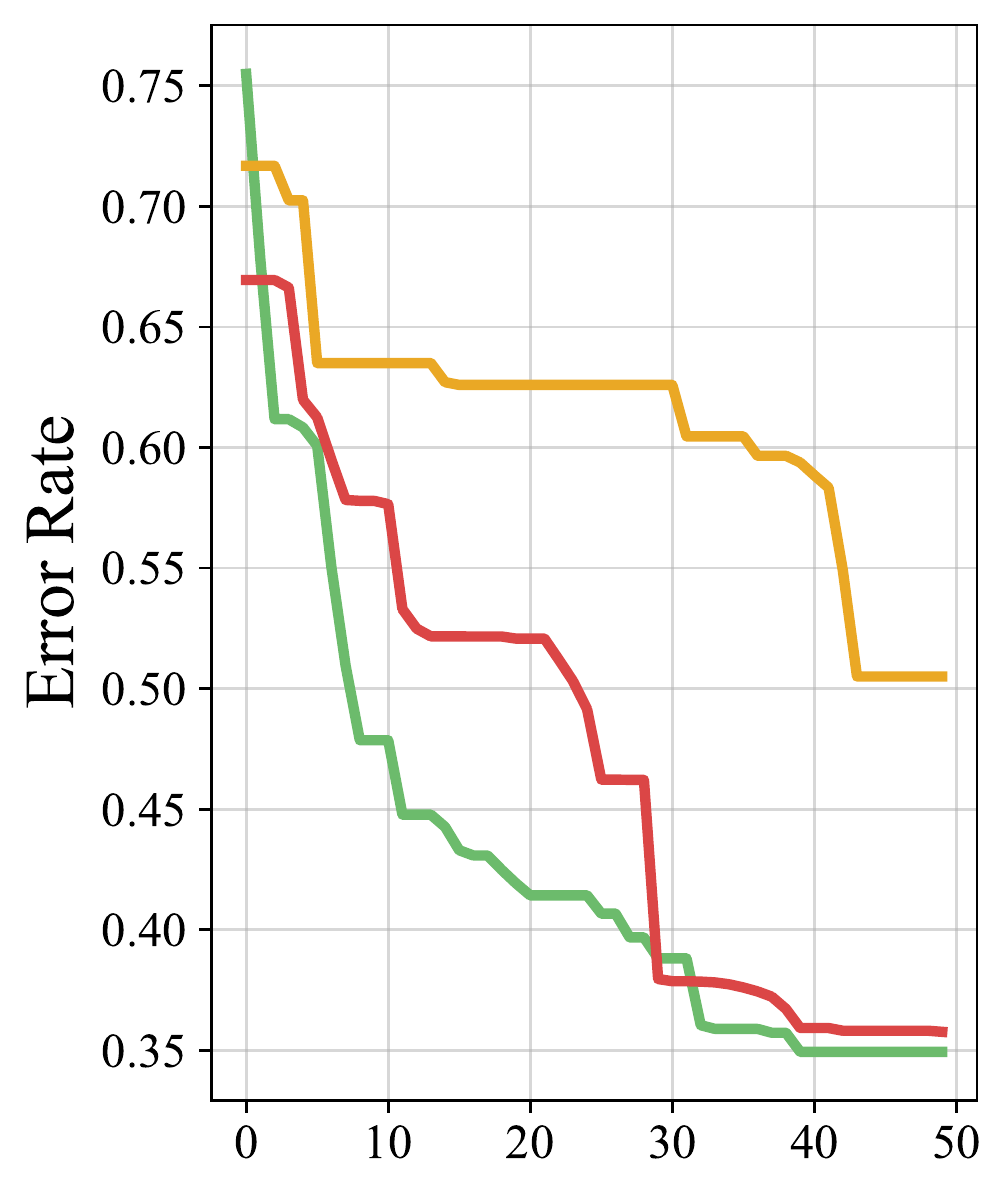}
        \caption{SVHN\\WideResNet 28-10}
    \end{subfigure}
    \begin{subfigure}{\textwidth}
    \vspace{1mm}
        \centering
        \includegraphics[width=0.35\textwidth]{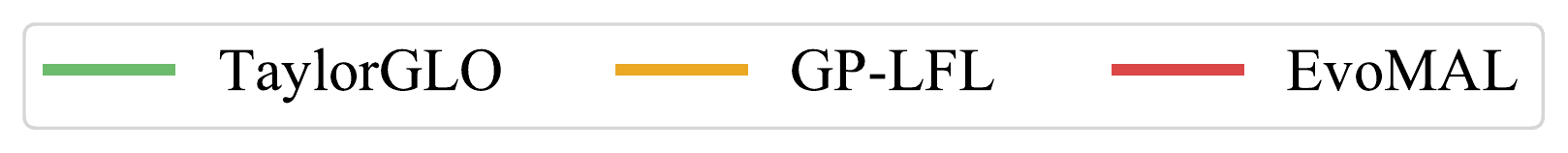}
    \end{subfigure}
    \vspace{-2mm}
    \captionsetup{justification=centering}
    \caption{Mean \textit{meta-training} learning curves across 5 independent executions of each algorithm, showing the fitness score (y-axis) against generations (x-axis), where a generation is equivalent to 25 evaluations. Best viewed in color.}
    \label{fig:meta-training-learning-curves}%
\vspace{-5mm}
\end{figure*}

To further validate the performance of EvoMAL, the much more challenging task of loss function transfer is assessed, where the meta-learned loss functions learned in one source domain are transferred to a new but related target domain. In our experiments, the meta-learned loss functions are taken directly from CIFAR-10 (in the previous section) and then transferred to CIFAR-100 with no further computational overhead, using the same model as the source. To ensure a fair comparison is made to the baseline, only the best-performing loss function found across the 5 random seeds from each method on each task + model pair are used. A summary of the final inference testing error rates when performing loss function transfer is given in Table \ref{table:meta-testing-transfer}.

The results show that even when using a single-task meta-learning setup where cross-task generalization is not explicitly optimized for, meta-learned loss functions can still be transferred with some success. In regards to meta-generalization, it is observed that EvoMAL and GP-LFL transfer their relative performance the most consistently to new tasks. In contrast, ML$^3$, which uses a neural network-based representation, often fails to generalize to the new tasks, a finding similar to \cite{hospedales2020meta} which found that symbolic representations often generalize better than sub-symbolic representations. Another notable result is that on both AlexNet and VGG-16 in the direct meta-learning setup, large gains in performance compared to the baseline are observed, as shown in Table \ref{table:meta-testing-results-classification}. Conversely, when the learned loss functions are transferred to CIFAR-100, all the loss function learning methods perform worse than the baseline as shown in Table \ref{table:meta-testing-transfer}. These results suggest that the learned loss functions have likely been meta-overfitted to the source task and that meta-regularization is an important aspect to consider for loss function transfer.

\vspace{-2mm}
\subsection{Meta-Training Performance}
\label{sec:meta-training}

The meta-training learning curves are given in Fig. \ref{fig:meta-training-learning-curves}, where the search performance of EvoMAL is compared to TaylorGLO and GP-LFL at each iteration and the performance is quantified by the fitness function using partial training sessions. Based on the results, it is very evident that adding local-search mechanisms into the EvoMAL framework dramatically increases the search effectiveness of GP-based loss function learning. EvoMAL is observed to consistently attain better performance than GP-LFL in a significantly shorter number of iterations. In almost all tasks, EvoMAL is shown to find better performing loss functions in the first 10 generations compared to those found by GP-LFL after 50. 

Contrasting EvoMAL to TaylorGLO, it is generally shown again that for most tasks, EvoMAL produces better solutions in a smaller number of iterations. Furthermore, performance does not appear to prematurely converge on the more challenging tasks of CIFAR-100 and SVHN compared to TaylorGLO and GP-LFL. Interestingly, on both CIFAR-10 AlexNet and SVHN WideResNet, TaylorGLO is able to achieve slightly better final solutions compared to EvoMAL on average; however, as shown by the final inference testing error rates in Tables \ref{table:meta-testing-results-classification} and \ref{table:meta-testing-results-regression}, these don't necessarily correspond to better final inference performance. This discrepancy between meta-training curves and final inference is also observed in the inverse case, where EvoMAL is shown to have much better learning curves than both TaylorGLO and GP-LFL, \textit{e.g.} in CIFAR-10 AllCNN-C and ResNet-18. However, the final inference error rates of EvoMAL in Table \ref{table:meta-testing-results-classification} are only marginally better than those of TaylorGLO and GP-LFL. 

This phenomenon is likely due to some of the meta-learned loss functions implicitly tuning the learning rate  (discussed further in \ref{sec:loss-functions}). Implicit learning rate tuning can result in increased convergence capabilities, \textit{i.e.}, faster learning which results in better fitness when using partial training sessions, but does not necessarily imply a strongly generalizing and robustly trained model at meta-testing time when using full training sessions.

\subsubsection{Meta-Training Run-Time}

\begin{table}
\centering
\caption{Average run-time of the \textit{meta-training} process for each of the benchmark methods. Each algorithm is run on a single Nvidia RTX A5000, and the results are reported in hours.}
\resizebox{1\columnwidth}{!}{
\begin{tabular}{lcccc}
\hline
\noalign{\vskip 1mm}
Task and Model      & ML$^3$        & TaylorGLO     & GP-LFL        & EvoMAL                \\ \hline \noalign{\vskip 1mm}
\textbf{Diabetes}   &               &               &               &                       \\
MLP                 & 0.01          & 0.83          & 0.53          & 1.93                  \\ \noalign{\vskip 1mm} \hline \noalign{\vskip 1mm} 
\textbf{Boston}     &               &               &               &                       \\
MLP                 & 0.01          & 0.85          & 0.46          & 1.59                  \\ \noalign{\vskip 1mm} \hline \noalign{\vskip 1mm} 
\textbf{California} &               &               &               &                       \\
MLP                 & 0.01          & 0.94          & 0.84          & 2.06                  \\ \noalign{\vskip 1mm} \hline \noalign{\vskip 1mm} 
\textbf{MNIST}      &               &               &               &                       \\
Logistic            & 0.02          & 1.44          & 0.61          & 2.45                  \\
MLP                 & 0.03          & 2.06          & 0.77          & 5.31                  \\
LeNet-5             & 0.03          & 2.30          & 0.82          & 3.29                  \\ \noalign{\vskip 1mm} \hline \noalign{\vskip 1mm} 
\textbf{CIFAR-10}   &               &               &               &                       \\
AlexNet             & 0.03          & 3.75          & 1.04          & 5.90                  \\
VGG-16              & 0.12          & 4.67          & 0.89          & 9.12                  \\
AllCNN-C            & 0.12          & 4.55          & 0.90          & 8.77                  \\
ResNet-18           & 0.60          & 9.14          & 1.02          & 54.22                 \\
PreResNet           & 0.41          & 8.40          & 0.96          & 41.73                 \\
WideResNet          & 0.63          & 12.85         & 0.98          & 66.72                 \\
SqueezeNet          & 0.12          & 4.95          & 0.41          & 11.18                 \\\noalign{\vskip 1mm} \hline \noalign{\vskip 1mm} 
\textbf{CIFAR-100}  &               &               &               &                       \\
WideResNet          & 0.20          & 20.34         & 1.34          & 57.61                 \\
PyramidNet          & 0.20          & 24.83         & 1.32          & 49.89                 \\\noalign{\vskip 1mm} \hline \noalign{\vskip 1mm} 
\textbf{SVHN}       &               &               &               &                       \\
WideResNet          & 0.20          & 41.57         & 1.09          & 67.28                 \\\noalign{\vskip 1mm} \hline
\end{tabular}
}
\label{table:run-time}
\end{table}

The use of a two-stage discovery process by EvoMAL enables the development of highly effective loss functions, as shown by the meta-training and meta-testing results. Producing on average models with superior inference performance compared to TaylorGLO and ML$^3$ which only optimize the coefficients/weights of fixed parametric structures, and GP-LFL which uses no local-search techniques. However, this bi-level optimization procedure where both the model structure and parameters are inferred adversely affects the computational efficiency of the meta-learning process, as shown in Table \ref{table:run-time}, which reports the average run-time (in hours) of meta-training for each loss function learning method.

The results show that EvoMAL is more computationally expensive than ML$^3$ and GP-LFL and approximately twice as expensive as TaylorGLO on average. Although EvoMAL is computationally expensive, it should be emphasized that this is still dramatically more efficient than GLO \cite{gonzalez2020improved}, the bi-level predecessor to TaylorGLO, whose costly meta-learning procedure required a supercomputer for even very simple datasets such as MNIST. 

The bi-level optimization process of EvoMAL is made computationally tractable by replacing the costly CMA-ES loss optimization stage from GLO with a significantly more efficient gradient-based procedure. On CIFAR-10, using the relatively small network, PreResNet-20 GLO required 11,120 partial training sessions and approximately 171 GPU days of computation \cite{gonzalez2020improved, gonzalez2021optimizing} compared to EvoMAL, which only needed on average 1.7 GPU days. In addition, the reduced run times of EvoMAL can also be partially attributed to the application of time-saving filters, which enables a subset of the loss optimizations and fitness evaluations to be either cached or obviated entirely. 

\begin{figure}[t!]
\centering

    \includegraphics[width=0.99\columnwidth]{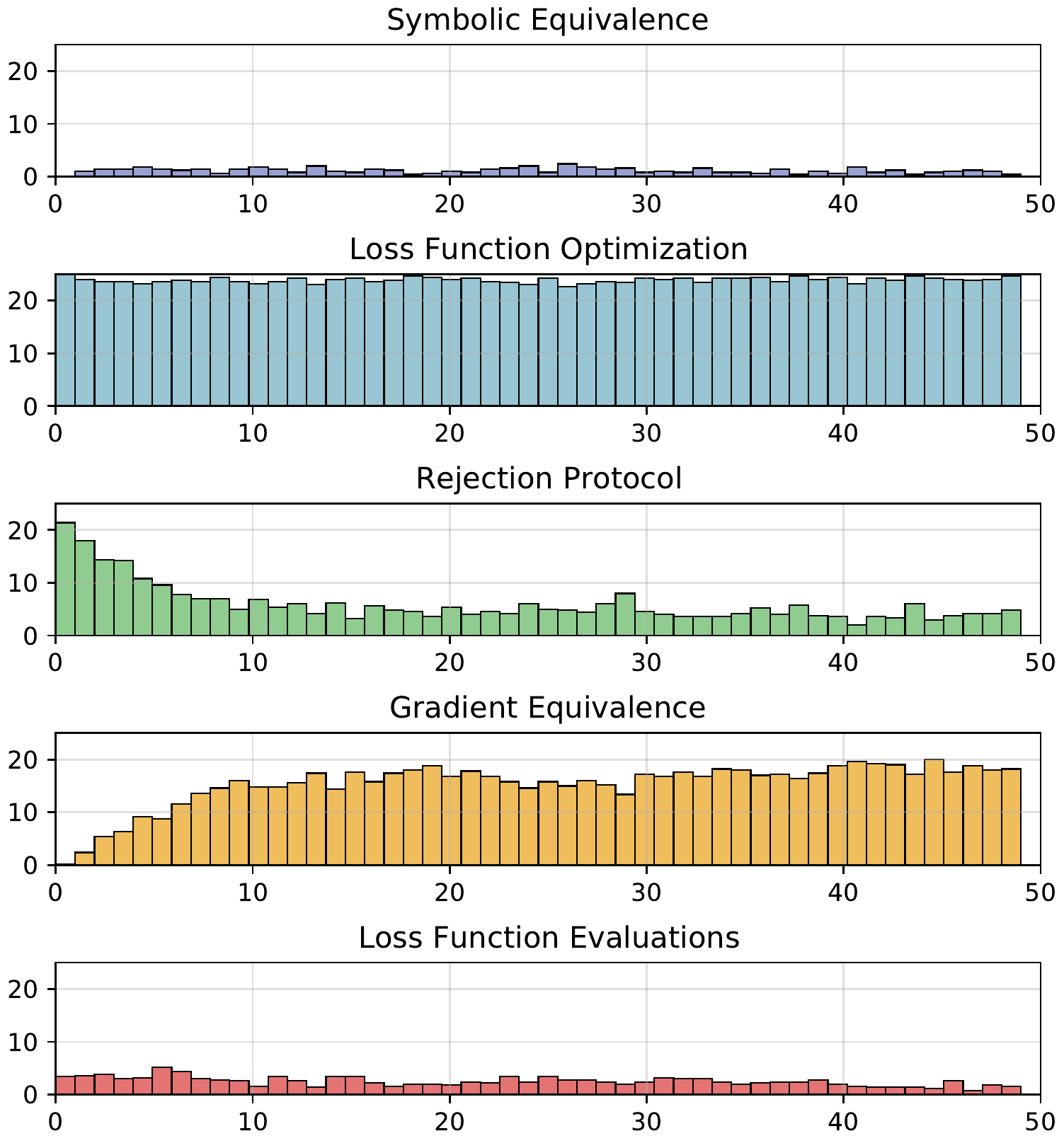}
    \caption{Frequencies of occurrence of the time saving filters, and the corresponding frequencies of loss optimization and evaluation throughout the symbolic search process. Reporting the average frequencies across all task + model pairs and independent executions of EvoMAL.}
    
\label{fig:meta-training-filters}
\vspace{-4mm}
\end{figure}

To summarize the effects of the time-saving filters, a set of histograms are presented in Fig. \ref{fig:meta-training-filters} showing the frequency of occurrence throughout the evolutionary process. Examining the symbolic equivalence filter results, it is observed that, on average, $\sim$10\% of the loss functions are identified as being symbolically equivalent at each generation; consequently, these loss functions have their fitness cached, and $\sim$90\% of the loss functions progress to the next stage and are optimized. Regarding the pre-evaluation filters, the rejection protocol initially rejects the majority of the optimized loss functions early in the search for being unpromising, automatically assigning them the worst-case fitness. In contrast, in the late stages of the symbolic search, this filter occurs incrementally less frequently, suggesting that convergence is being approached, further supported by the frequency of occurrence of the gradient equivalence filter, which caches few loss functions at the start of the search, but many near the end. Due to this aggressive filtering, only $\sim$25\% of the population at each generation have their fitness evaluated, which helps to reduce the run-time further. Note that the run-time of EvoMAL can be further reduced for large-scale optimization problems through parallelization to distribute loss function optimization and evaluation across multiple GPUs or clusters. 

\begin{figure*}
\centering
\begin{subfigure}{.5\textwidth}
  \centering
  \includegraphics[width=\textwidth, height=4.5cm]{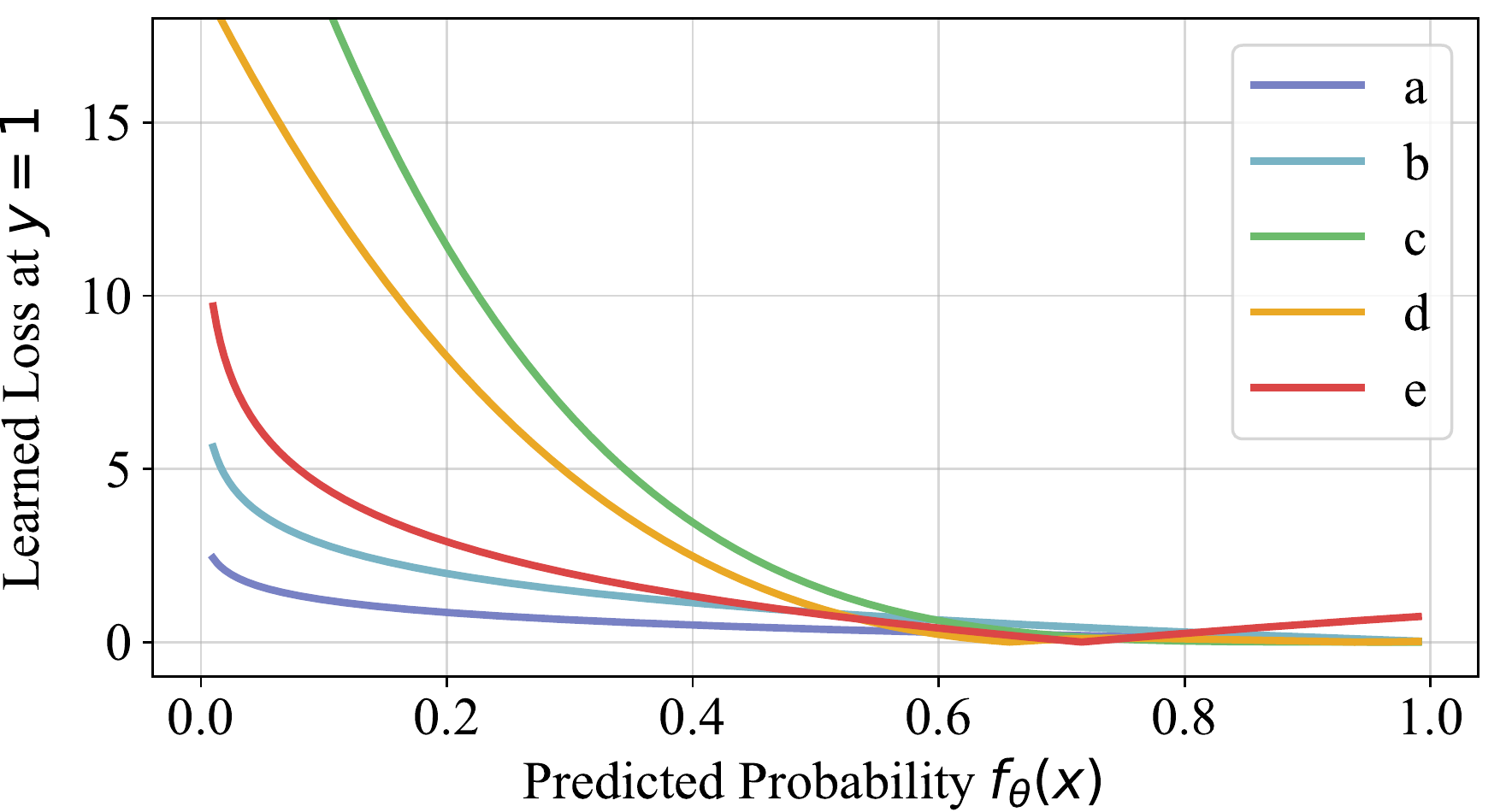}
\end{subfigure}%
\begin{subfigure}{.5\textwidth}
  \centering
  \includegraphics[width=\textwidth, height=4.5cm]{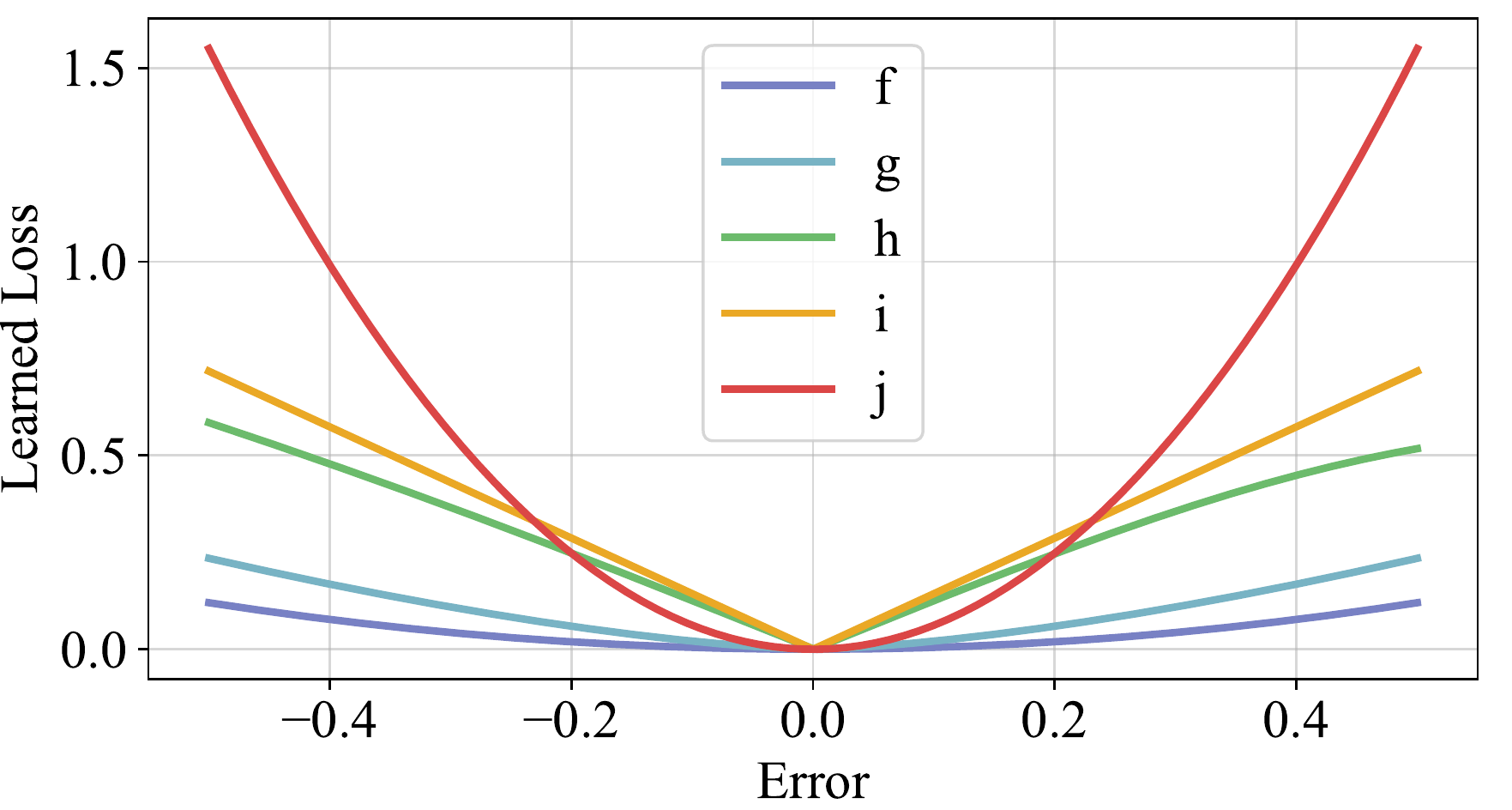}
\end{subfigure}
\captionsetup{justification=centering}
\caption{Example loss functions meta-learned by EvoMAL. \textbf{The left plot shows classification loss functions, and \\ right shows regression loss functions.} Best viewed in color.}
\label{fig:meta-learned-loss-functions}
\end{figure*}

\subsection{Meta-Learned Loss Functions}
\label{sec:loss-functions}

\renewcommand{\arraystretch}{1.2}
\begin{table}

\centering
\caption{Example loss functions meta-learned by EvoMAL, where solutions have been numerically and algebraically simplified. Furthermore, the parameters $\phi$ have also been omitted for improved clarity and parsimony.}
\vspace{-3mm}

\begin{subfigure}{.5\textwidth}
  
\begin{tabular}{l>{\centering\arraybackslash}p{0.85\columnwidth}}
\multicolumn{1}{c}{\textbf{}} &  \\ \hline \noalign{\vskip 1mm}

a. & $|\log(\sqrt{y \cdot f_{\theta}(x)} + \epsilon)|$ \\ 

b. & $\log(y \cdot f_{\theta}(x)) + \sqrt{\log(y \cdot f_{\theta}(x))}^{2}$ \\ \noalign{\vskip 1mm}

c. & $((y - f_{\theta}(x))/f_{\theta}(x))^4$ \\ \noalign{\vskip 1mm} 

d. & $y^3 + f_{\theta}(x)^3 + f_{\theta}(x) \cdot y^2 + f_{\theta}(x)^2 \cdot y$ \\ \noalign{\vskip 1mm}

e. & $|\log((y \cdot f_{\theta}(x))^2 + \epsilon)|$ \\ \noalign{\vskip 1mm} 

\hline \noalign{\vskip 1mm} \hline \noalign{\vskip 1mm}

f. & $(f_{\theta}(x) - \min(\max(y, -1), 1))^2$ \\ \noalign{\vskip 1mm}

g. & $|y \cdot (\sqrt{|y/\sqrt{1 + f_{\theta}(x)^2}|})|$ \\ \noalign{\vskip 1mm}

h. &  $|(y/(1 + \log(|y - 1|)^2)) - f_{\theta}(x)|$ \\ \noalign{\vskip 1mm}

i. & $\sqrt{|y \cdot (y - f_{\theta}(x))|}$ \\ \noalign{\vskip 1mm}

j. & $\sqrt{|y - f_{\theta}(x)|}$ \\ \noalign{\vskip 1mm} \hline

\end{tabular}
\vspace{-3mm}
 
\end{subfigure}

\label{table:meta-learned-loss-functions}
\end{table}

To better understand why the meta-learned loss functions produced by EvoMAL are so performant, an analysis is conducted on a subset of the interesting loss functions found throughout the experiments. In Fig. \ref{fig:meta-learned-loss-functions}, examples of the meta-learned loss functions produced by EvoMAL are presented. The corresponding loss functions are also given symbolically in Table \ref{table:meta-learned-loss-functions}.

\subsubsection{Learned Loss Functions for Classification Tasks}

The classification loss functions meta-learned by EvoMAL appear to converge upon three classes of loss functions. First are cross-entropy loss variants such as loss functions a) and b), which closely resemble the cross-entropy loss functionally and symbolically. Second, are loss functions that have similar characteristics to the parametric focal loss \cite{lin2017focal}, such as loss functions c) and d). These loss functions recalibrate how easy and hard samples are prioritized; in most cases, very little or no loss is attributed to high-confidence correct predictions, while significant loss is attributed to high-confidence wrong predictions. Finally, loss functions such as e) which demonstrate unintuitive behavior, such as assigning more loss to confident and correct solutions relative to unconfident and correct solutions, a characteristic which induces implicit label smoothing regularization \cite{szegedy2016rethinking, muller2019does}, see analysis in Appendix A. 

\subsubsection{Learned Loss Functions for Regression Tasks}

Analyzing the regression loss functions it is observed that there are several unique behaviors particularly around incorporating strategies for improving robustness to outliers. For example in f) the loss function takes on the form of the squared loss; however, it incorporates a thresholding operation via the $\max$ operator for directly limiting the size of the ground truth label. Alternatively in loss functions g), i), and j), there is frequent use of the square root operator, which decreases the loss attributed to increasingly large errors. Finally, we observe that many of the learned loss functions, take on shapes similar to well-known handcrafted loss functions; for example, h) appears to be a cross between the absolute loss and the Cauchy (Lorentzian) loss \cite{black1996unification}.

\subsubsection{General Observations}

In addition to the trends identified for classification and regression, respectively, there are several more noteworthy trends:

\begin{itemize}[leftmargin=*]

    \item Structurally complex loss functions typically perform worse relative to simpler loss functions, potentially due to the increased meta or base optimization difficulties. This is a similar finding to what was found when meta-learning activation functions \cite{ramachandran2017searching}. 
  
    \item Many of the learned loss functions in classification are asymmetric, producing different loss values for false positive and false negative predictions, often caused by exploiting $f_{\theta}(x)$ softmax output activation, where the sum of the class-wise outputs is required to equal to 1. 

\end{itemize}
\vspace{-2mm}

\subsection{Loss Landscapes Analysis}
\label{sec:loss-landscapes}

\begin{figure}[t!]
    \centering
    {\includegraphics[width=\columnwidth,height=3.75cm]{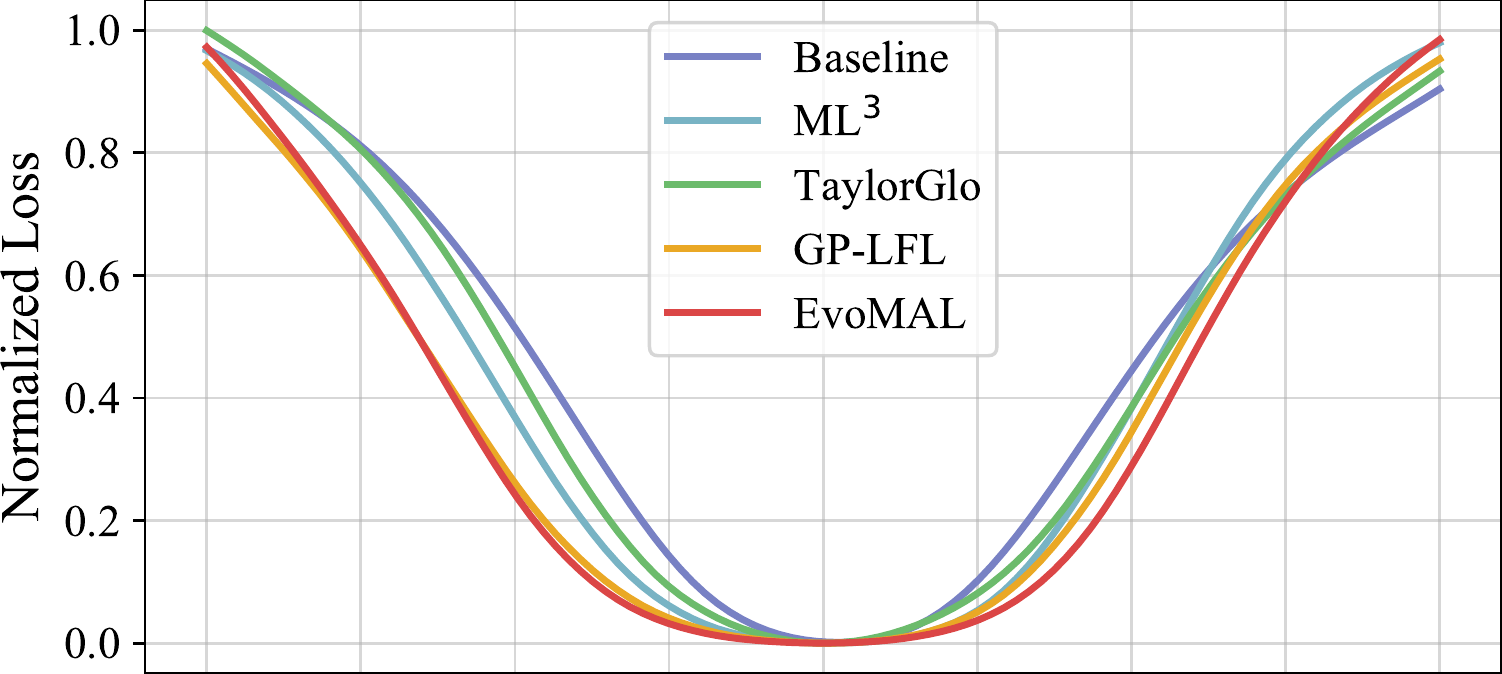}\vspace{5mm}}%
    \qquad
    {\includegraphics[width=\columnwidth, height=4.0cm]{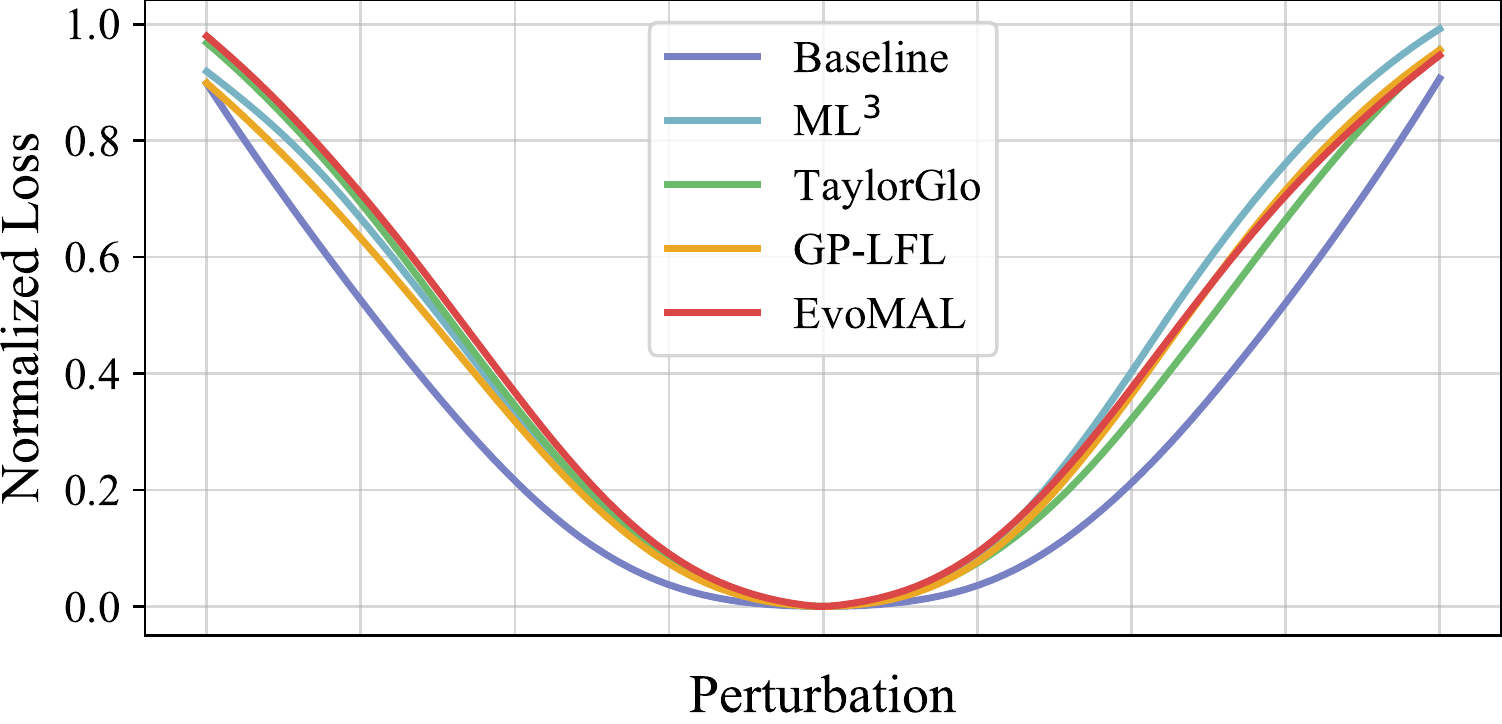}}%
    \caption{The average 1D loss landscapes generated on the CIFAR-10 dataset, the \textbf{top figure shows the loss landscapes generated on AllCNN-C, and the bottom figure shows the loss landscapes generated on AlexNet}. The landscapes show the average (mean) loss taken across the 5 independent executions of each algorithm on each task+model pair. Best viewed in color.}
    \vspace{-3mm}
    \label{fig:loss-landscapes}%
\end{figure}

Loss function learning as a paradigm has consistently shown to be an effective way of improving performance; however, it is not yet fully understood what exactly meta-learned loss functions are learning and why they are so performant compared to their handcrafted counterparts.  In \cite{gonzalez2021optimizing} and \cite{gao2022loss}, it was found that the loss landscapes of models trained with learned loss functions produce flatter landscapes relative to those trained with the cross-entropy loss. The flatness of a loss landscape has been hypothesized to correspond closely to a model's generalization capabilities \cite{hochreiter1997flat, keskar2016large, li2018visualizing, chaudhari2019entropy}; thus, they conclude that meta-learned loss functions improve generalization. These findings are independently reproduced and are shown in Fig. \ref{fig:loss-landscapes}. The loss landscapes are generated using the filter-wise normalization method \cite{li2018visualizing}, which plots a normalized random direction of the weight space $\theta$. 

The loss landscapes visualizations show that the loss functions developed by EvoMAL can produce flatter loss landscapes on average in contrast to those produced by the cross-entropy, ML$^3$, TaylorGLO, and GP-LFL, as shown in the top figure which shows the landscapes generated using AllCNN-C. In contrast to prior findings, we also find that in some cases the meta-learned loss functions produced in our experiments show relatively sharper loss landscapes compared to those produced using the cross-entropy loss, as shown in the bottom figure which shows loss landscapes generated using the base model AlexNet. These findings suggest that the relative flatness of the loss landscape does not fully explain why meta-learned loss functions can produce improved performance, especially since there is evidence that sharp minma can generalize well \textit{i.e.} “\textit{flat vs sharp}” debate \cite{dinh2017sharp}.

\subsection{Implicit Learning Rate Tuning}
\label{sec:implicit-tuning}

\begin{figure}
\centering

    \includegraphics[width=\columnwidth, height=4.5cm]{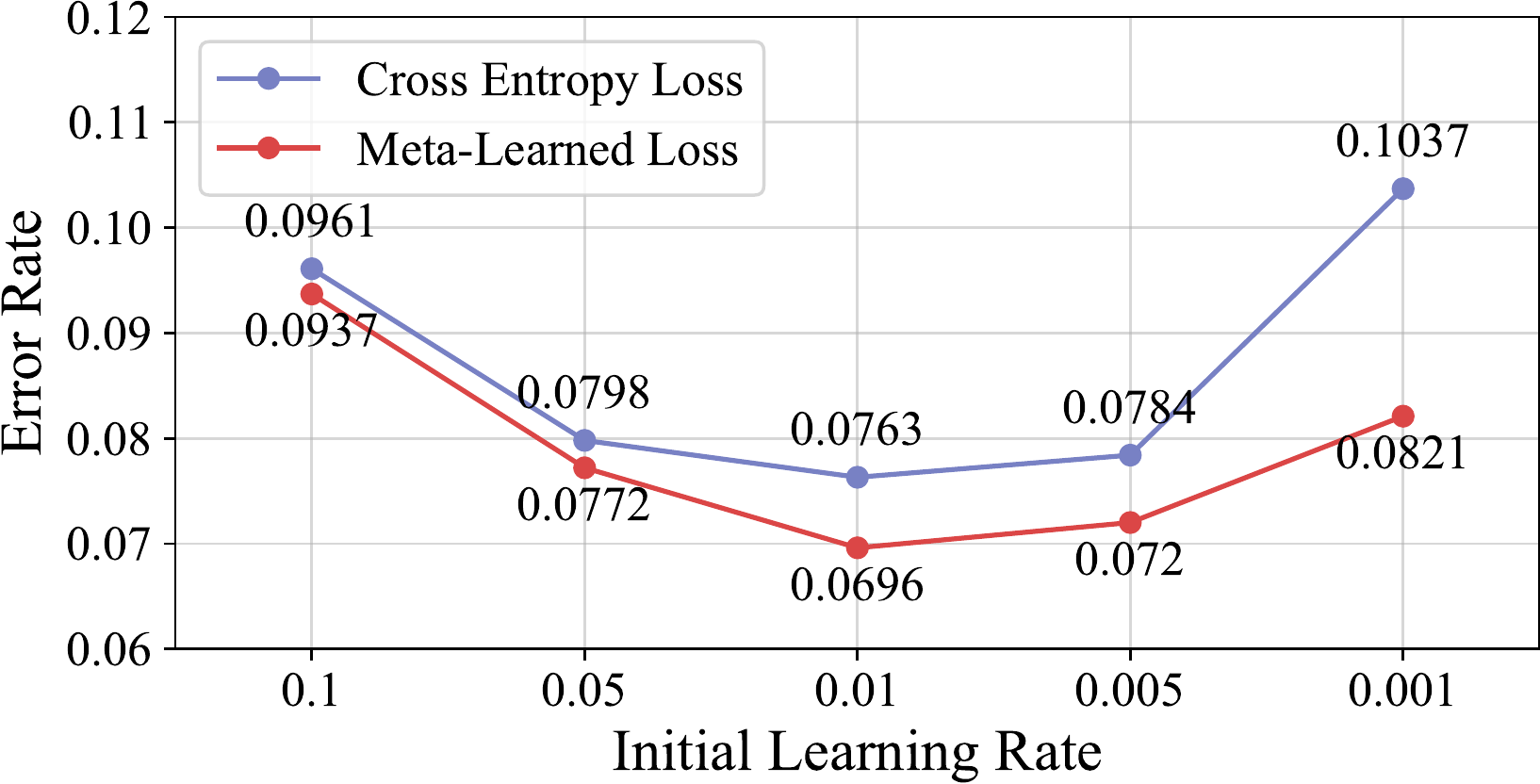}
    \caption{Grid search comparing the average error rate of the baseline cross-entropy loss and EvoMAL on CIFAR-10 AllCNN-C across a set of base learning rate values, where 5 executions of each algorithm are performed on each learning rate value.}
    \vspace{-3mm}
    
\label{fig:learning-rate}
\end{figure}

Another explanation for why meta-learned loss functions improve performance over handcrafted loss functions is that they can implicitly tune the (base) learning rate since for some suitably expressive representation of $\MetaLoss$, since
\begin{equation}\label{eq:implicit-learning-rate-tuning}
    \exists\alpha\exists\phi:\theta - \alpha\nabla_{\theta}\Loss_{\Task} \approx \theta - \nabla_{\theta}\MetaLoss_{\phi}^{\Transpose}.
\end{equation}
Thus, performance improvement when using meta-learned loss functions may be the indirect result of a change in the learning rate, scaling the resulting gradient of the loss function. To validate the implicit learning rate hypothesis, a grid search is performed over the base-learning rate using the cross-entropy loss and EvoMAL on CIFAR-10 AllCNN-C. The results are shown in Fig. \ref{fig:learning-rate}. 

The results show that the base learning rate $\alpha$ is a crucial hyper-parameter that influences the performance of both the baseline and EvoMAL. However, in the case of EvoMAL, it is found that when using a relatively small $\alpha$ value, the base learning rate is implicitly tuned, and the loss function learning algorithm achieves an artificially large performance margin compared to the baseline. Implicit learning rate tuning of a similar magnitude is also observed when using relatively large $\alpha$ values; however, the algorithm's stability is inconsistent, with some runs failing to converge. Finally, when a near-optimal $\alpha$ value is used, performance improvement is consistently better than the baseline. These results indicate two key findings: (1) meta-learned loss functions improve upon handcrafted loss functions and that the performance improvement when using meta-learned loss functions is not primarily a result of implicit learning rate tuning when $\alpha$ is tuned. (2) The base learning rate $\alpha$ can be considered as part of the initialization of the meta-learned loss function, as it determines the initial scale of the loss function.

\section{Conclusions and Future Work}
\label{sec:conclusions}

This work presents a new framework for meta-learning symbolic loss function via a hybrid neuro-symbolic search approach called Evolved Model-Agnostic Loss (EvoMAL). The proposed technique uses genetic programming to learn a set of expression tree-based loss functions, which are subsequently transformed into a new network-style representation using a newly proposed transitional procedure. This new representation enables the integration of a computationally tractable gradient-based local-search approach to enhance the search capabilities significantly. Unlike previous approaches, which stack evolution-based techniques, EvoMAL's efficient local-search enables loss function learning on commodity hardware. 

The experimental results confirm that EvoMAL consistently meta-learns loss functions which can produce more performant models compared to those trained with conventional handcrafted loss functions, as well as other state-of-the-art loss function learning techniques. Furthermore, analysis of some of the meta-learned loss functions reveals several key findings regarding common loss function structures and how they interact with the models trained by them. Finally, automating the task of loss function selection has shown to enable a diverse and creative set of loss functions to be generated, which would not be replicable through a simple grid search over handcrafted loss functions.

There are many promising future research directions as a consequence of this work. Regarding algorithmic extensions, the EvoMAL framework is general in its design. It would be interesting to extend it to different meta-learning applications such as gradient-based optimizers or activation functions. In terms of loss function learning, a natural extension of the work would be to meta-learn the loss function and other deep neural network components simultaneously, similar to the methods presented in \cite{ravi2017optimization, li2017meta, elsken2020meta, baik2021meta}. For example, the meta-learned loss functions could consider additional arguments such as the timestep or model weights, which would implicitly induce learning rate scheduling or weight regularization, respectively. Another example would be to combine neural architecture search with loss function learning, as the experiments in this work use handcrafted neural network architectures which are biased towards the squared error loss or cross-entropy loss since they were designed to optimize for that specific loss function. Larger performance gains may be achieved using custom neural network architectures explicitly designed for the meta-learned loss functions.

\bibliography{references.bib}
\bibliographystyle{IEEEtran}

\ifCLASSOPTIONcaptionsoff
  \newpage
\fi

\begin{IEEEbiography}[{\includegraphics[width=1in,height=1.25in,clip,keepaspectratio]
{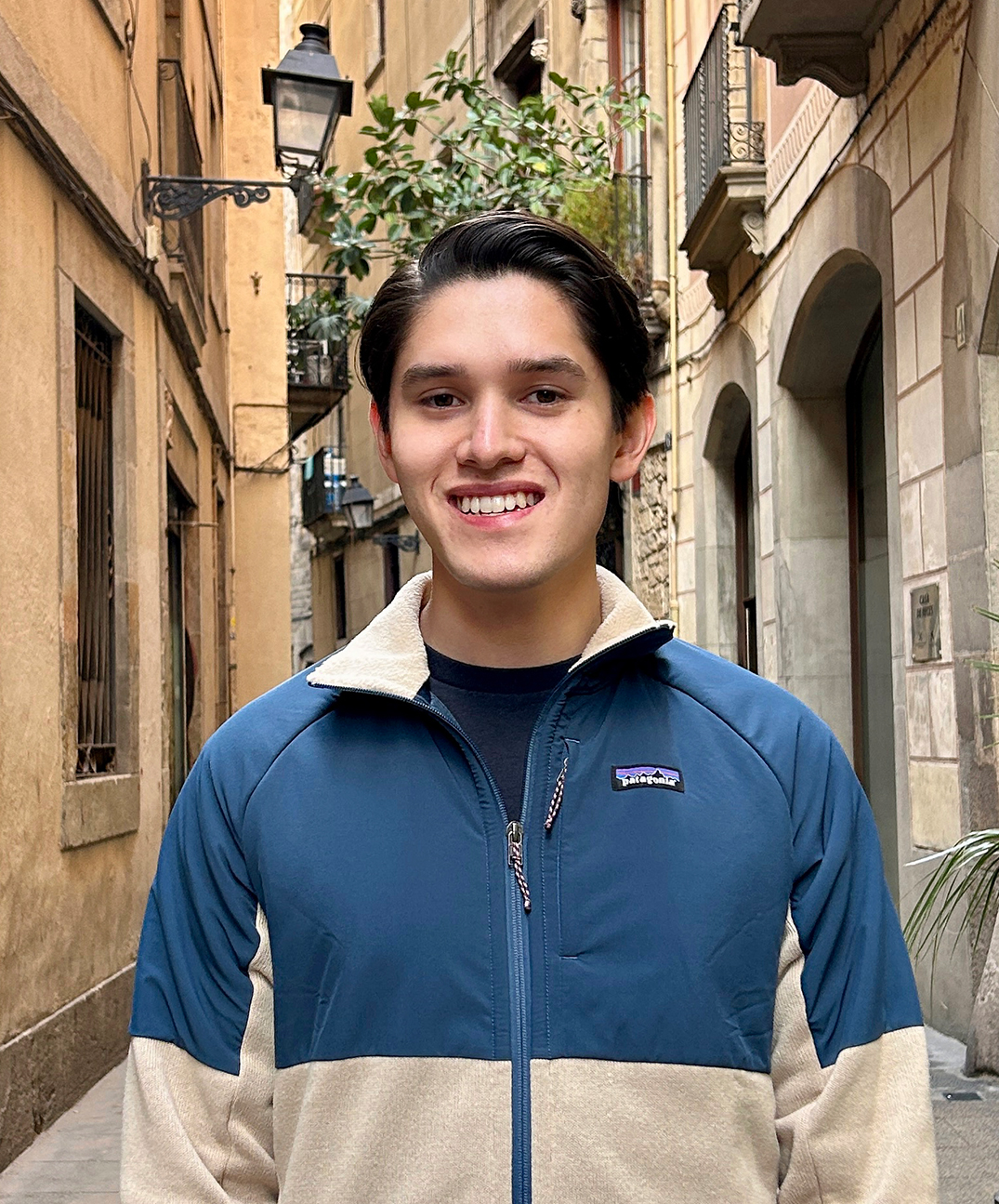}}]{Christian Raymond}
received a  B.Sc. and  B.Sc. (Hons) degrees from the School of Engineering and Computer Science, Victoria University of Wellington, New Zealand, in 2020 and 2021 respectively. He is currently pursuing a Ph.D. in Artificial Intelligence with the Evolutionary Computation Research Group (ECRG) at the Centre for Data Science and Artificial Intelligence (CDSAI) at Victoria University of Wellington. His research interests include meta-learning, hyper-parameter optimization, and few-shot learning.
\end{IEEEbiography}

\begin{IEEEbiography}[{\includegraphics[width=1in,height=1.25in,clip,keepaspectratio]
{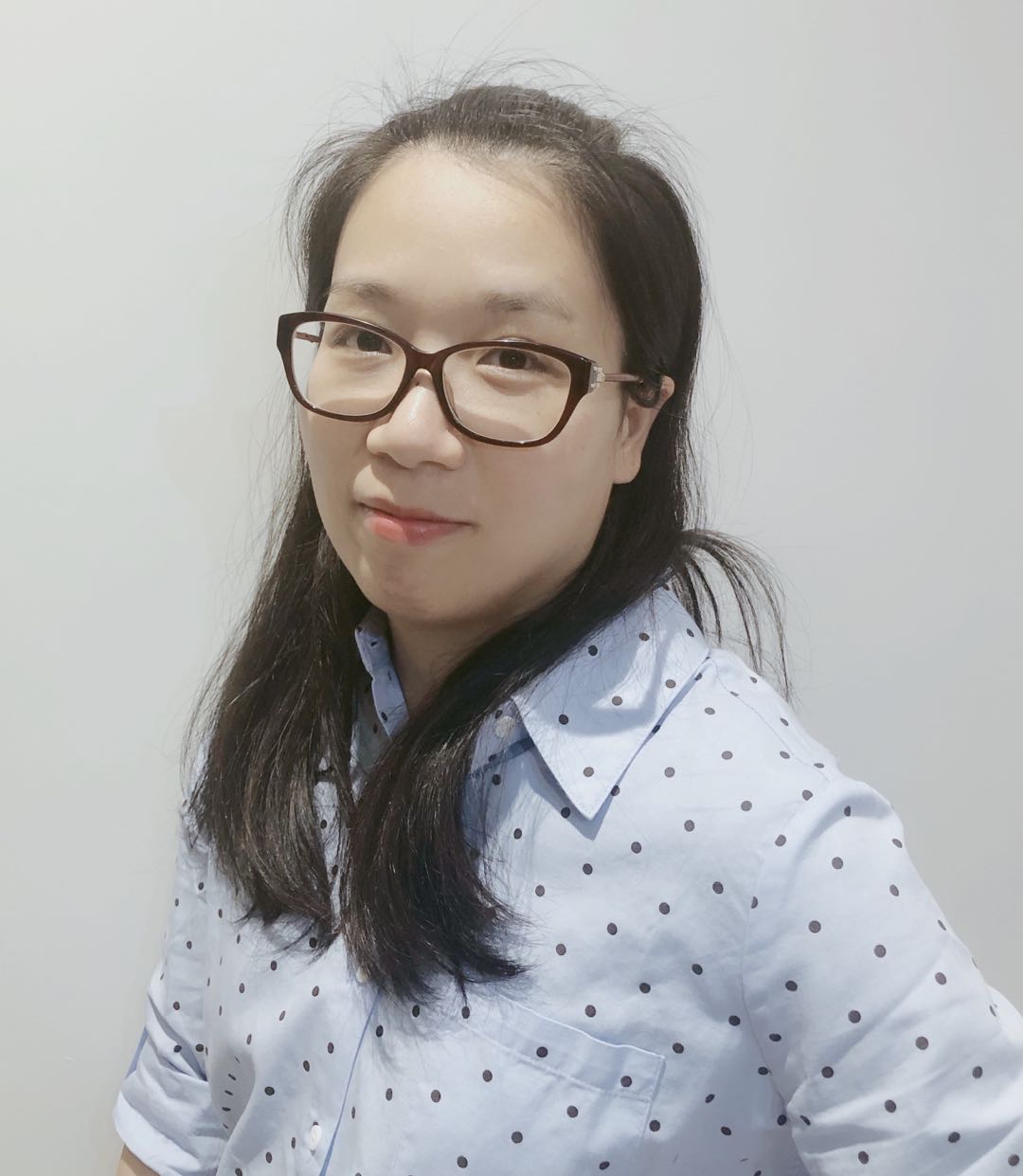}}]{Qi Chen} (M’14) received a B.E. degree in Automation from the University of South China, Hunan, China in 2005 and an M.E. degree in Software Engineering from Beijing Institute of Technology, Beijing, China in 2007, and a PhD degree in computer science in 2018 at Victoria University of Wellington, New Zealand. Since 2014, she has joined the Evolutionary Computation Research Group at Victoria University of Wellington (VUW). Currently, she is a Lecturer in the School of Engineering and Computer Science at VUW. Qi’s current research mainly focuses on genetic programming for symbolic regression. Her research interests include machine learning, evolutionary computation, feature selection, feature construction, transfer learning, domain adaptation, and statistical learning theory. She serves as a reviewer of international conferences, including IEEE Congress on Evolutionary Computation, and international journals, including IEEE Transactions on Cybernetics and IEEE Transactions on Evolutionary Computation.
\end{IEEEbiography}
\vspace{-5mm}
\begin{IEEEbiography}[{\includegraphics[width=1in,height=1.25in,clip,keepaspectratio]
{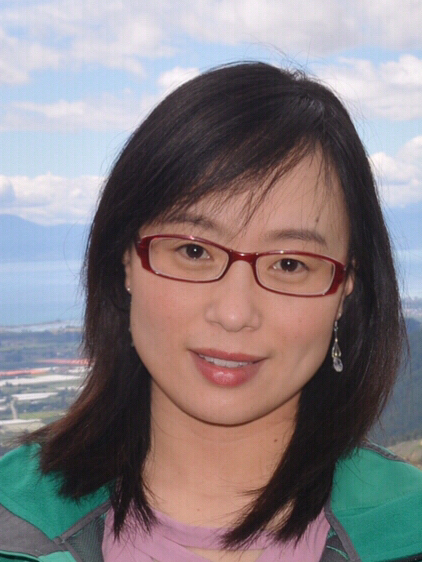}}]{Bing Xue} (M’10-SM’21) received the B.Sc. degree from the Henan University of Economics and Law, Zhengzhou, China, in 2007, the M.Sc. degree in management from Shenzhen University, Shenzhen, China, in 2010, and the Ph.D. degree in computer science in 2014 at Victoria University of Wellington (VUW), New Zealand. She is currently a Professor in Computer Science, and Program Director of Science in the School of Engineering and Computer Science at VUW. She has over 300 papers published in fully refereed international journals and conferences and her research focuses mainly on evolutionary computation, machine learning, classification, symbolic regression, feature selection, evolving deep neural networks, image analysis, transfer learning, and multi-objective machine learning. Dr. Xue is currently the Chair of IEEE Computational Intelligence Society (CIS) Task Force on Transfer Learning and Transfer Optimization, Vice-Chair of IEEE CIS Evolutionary Computation Technical Committee, Editor of IEEE CIS Newsletter, Vice-Chair of IEEE Task Force on Evolutionary Feature Selection and Construction, and Vice-Chair IEEE CIS Task Force on Evolutionary Deep Learning and Applications. She has also served asassociate editor of several international journals, such as IEEE Computational Intelligence Magazine and IEEE Transactions on Evolutionary Computation.
\end{IEEEbiography}
\vspace{-5mm}
\begin{IEEEbiography}[{\includegraphics[width=1in,height=1.25in,clip,keepaspectratio]
{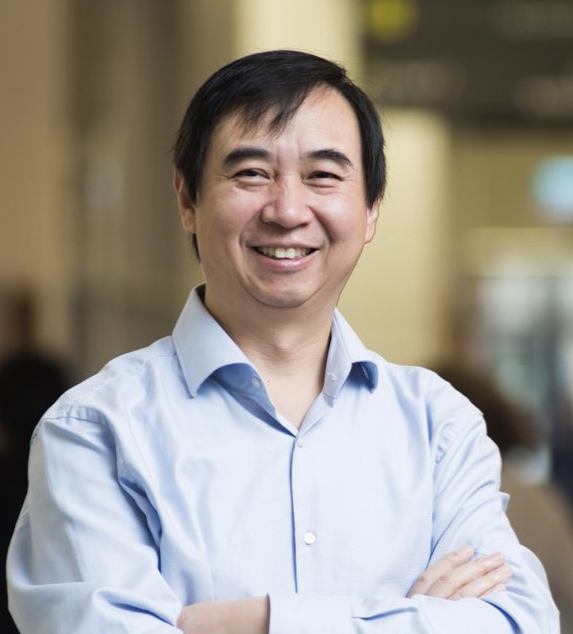}}]{Mengjie Zhang} (M’04-SM’10-F’19) received the B.E. and M.E. degrees from the Artificial Intelligence Research Center, Agricultural University of Hebei, Hebei, China, and the Ph.D. degree in computer science from RMIT University, Melbourne, VIC, Australia, in 1989, 1992, and 2000, respectively. He is currently a Professor of Computer Science, Head of the Evolutionary Computation Research Group, and the Associate Dean (Research and Innovation) in the Faculty of Engineering. His current research interests include evolutionary computation, particularly genetic programming, particle swarm optimization, and learning classifier systems with application areas of image analysis, multiobjective optimization, feature selection and reduction, job shop scheduling, and transfer learning. He has published over 600 research papers in refereed international journals and conferences. Prof. Zhang is a Fellow of the Royal Society of New Zealand and has been a Panel Member of the Marsden Fund (New Zealand Government Funding), a Fellow of IEEE, and a member of ACM. He was the chair of the IEEE CIS Intelligent Systems and Applications Technical Committee, and chair for the IEEE CIS Emergent Technologies Technical Committee, the Evolutionary Computation Technical Committee, and a member of the IEEE CIS Award Committee. He is a vice-chair of the IEEE CIS Task Force on Evolutionary Feature Selection and Construction, a vice-chair of the Task Force on Evolutionary Computer Vision and Image Processing, and the founding chair of the IEEE Computational Intelligence Chapter in New Zealand. He is also a committee member of the IEEE NZ Central Section.
\end{IEEEbiography}

\appendices

\begingroup\makeatletter\def\f@size{8}\check@mathfonts

\section{Theoretical Analysis}
\label{sec:theoretical-analysis}

To further investigate the questions of why do meta-learned loss functions perform better than handcrafted loss functions, and what are meta-learned loss functions learning, we perform a theoretical analysis inspired by the analysis performed in \cite{gonzalez2020effective}. The analysis reveals important connections between our learned loss functions and the cross-entropy loss and label smoothing regularization. 

\subsection{Meta-Learned Loss Function}\label{sec:deriving-learned}

In this theoretical analysis, we analyze the general form of loss functions a) and e) from Table \ref{table:meta-learned-loss-functions} in Section \ref{sec:loss-functions} of the main manuscript (where $\epsilon=1e-7$ is a small constant):
\begin{equation}
\mathcal{M}_{a} = |\log(\sqrt{y \cdot f_{\theta}(x) + \epsilon})|
\end{equation}
\begin{equation}
\mathcal{M}_{e} = |\log((y \cdot f_{\theta}(x) + \epsilon)^2)|
\end{equation}
where $\mathcal{M}_{a}$ and $\mathcal{M}_{e}$ are equivalent upto a scaling factor $\phi_{0}$, since
\begin{equation}
|\log((y \cdot f_{\theta}(x) + \epsilon)^{\phi_{0}})| = \phi_{0} \cdot |\log(y \cdot f_{\theta}(x) + \epsilon )|.
\end{equation}
Therefore, the general form of the learned loss functions can be given as follows:
\begin{equation}
\mathcal{M}_{\phi} = |\log(y \cdot f_{\theta}(x) + \epsilon)|
\end{equation}
where, one further simplification can be made since the non-target learned loss (\textit{i.e.} when $y=0$)  evaluates to $|\log(\epsilon)|$ which is not dependent on the base model predictions, and supplies a constant gradient. Consequently, the target label $y$ can be taken outside the $\log$ and $\epsilon$ can be dropped giving us the final learned loss function which we further refer to as the \textit{Absolute Cross-Entropy Loss} ($\Loss_{ACE}$) function:
\begin{equation}
\Loss_{ACE} = y \cdot |\log(f_{\theta}(x))|
\end{equation}
where the meta parameters $\phi_0$ and $\phi_1$ can be made explicit as they reveal key insights into why $\Loss_{ACE}$ can improve performance.
\begin{equation}\label{eq:learned-loss-function}
\Loss_{ACE} = \phi_{0} \cdot y \cdot |\log(\phi_{1} \cdot f_{\theta}(x))|
\end{equation}
Regarding $\phi_{0}$, it is straightforward to see that this corresponds to an explicit linear scaling of the loss function. This is identical in behavior to what the base learning rate $\alpha$ is doing in vanilla stochastic gradient descent and makes explicit the equality we established in Equation \eqref{eq:implicit-learning-rate-tuning}. As for $\phi_{1}$, as will be shown in the subsequent analysis, this parameter enables $\Loss_{ACE}$ to regularize overconfident predictions in a similar fashion label smoothing regularization, resulting in improved generalization performance onto out-of-sample instances.

\subsection{Learning Rule Decomposition}\label{sec:decomposition}

First, we decompose our learning rule to isolate the contribution of the loss function. For simplicity, we will consider the simple case of using vanilla stochastic gradient descent (SGD):
\begin{equation}
    \theta_{t+1} \leftarrow \theta_{t} - \alpha \nabla_{\theta_{t}}\Big[\mathcal{L}(y, f_{\theta_{t}}(x))\Big]
\end{equation}
where $\alpha$ is the (base) learning rate and $\mathcal{L}$ is the loss function, which takes as arguments the true label $y$ and the base model predictions $f_{\theta}(x)$. Consequently, the learning rule for the base model weights $\theta_{t}$ based on the  $i^{th}$ model output $f_{\theta}(x)_{i}$ and target $y_{i}$ can be described as follows:

\begin{equation}\label{eq:general-form}
\begin{split}
\theta_{t+1} 
& \leftarrow \theta_{t} - \alpha \frac{\partial}{\partial \theta_{t}} \Big[\mathcal{L}(y_{i}, f_{\theta_{t}}(x)_{i})\Big] \\
& = \theta_{t} - \alpha \bigg[\frac{\partial}{\partial f}\mathcal{L}(y_{i}, f_{\theta_{t}}(x)_{i}) \cdot \frac{\partial}{\partial \theta_{t}} f_{\theta}(x)_{i}\bigg] \\
& = \theta_{t} + \alpha \bigg[\underbrace{-\frac{\partial}{\partial f}\bigg(\mathcal{L}(y_{i}, f_{\theta_{t}}(x)_{i})\bigg)}_{\delta} \cdot \frac{\partial}{\partial \theta_{t}} f_{\theta_{t}}(x)_{i}\bigg]
\end{split}
\end{equation}
Given this general form, we can now substitute any loss function into Equation \eqref{eq:general-form} to obtain a unique expression $\delta$ which describes the behavior of the loss function with respect to the probabilistic output. Through substitution of the $\Loss_{ACE}$ into the SGD learning rule we get
\begin{equation}
\theta_{t+1} \leftarrow \theta_{t} + \alpha \bigg[\underbrace{-\frac{\partial}{\partial f} \bigg(y \cdot |\log(\phi_{1} \cdot f_{\theta}(x))| \bigg)}_{\delta_{\Loss_{ACE}}} \cdot \frac{\partial}{\partial \theta_{j}} f_{\theta}(x)_{i}\bigg],
\end{equation}
where the behavior $\delta_{\Loss_{ACE}}$ is defined as follows:
\begin{equation}
\begin{split}
\delta_{\Loss_{ACE}} 
&= - \frac{\partial}{\partial f} \bigg(y \cdot |\log(\phi_{1} \cdot f_{\theta}(x))|\bigg) \\
&= - y \cdot \frac{\log(\phi_{1} \cdot f_{\theta}(x))}{|\log(\phi_{1} \cdot f_{\theta}(x))|} \cdot \frac{\partial}{\partial f} \bigg(\log(\phi_{1} \cdot f_{\theta}(x))\bigg) \\
&= - y \cdot \frac{\log(\phi_{1} \cdot f_{\theta}(x))}{|\log(\phi_{1} \cdot f_{\theta}(x))|} \cdot \frac{1}{\phi_{1} f_{\theta}(x)} \cdot \frac{\partial}{\partial f} \bigg(\phi_{1} \cdot f_{\theta}(x)\bigg) \\
&= - y \cdot \frac{\log(\phi_{1} \cdot f_{\theta}(x))}{|\log(\phi_{1} \cdot f_{\theta}(x))|} \cdot \frac{1}{\phi_{1} f_{\theta}(x)} \cdot \phi_{1} \\
&= - \frac{\phi_{1} \cdot y \cdot \log(\phi_{1} \cdot f_{\theta}(x))}{\phi_{1} \cdot f_{\theta}(x) \cdot |\log(\phi_{1} \cdot f_{\theta}(x))|} \\
&= - \frac{y \cdot \log(\phi_{1} \cdot f_{\theta}(x))}{f_{\theta}(x) \cdot |\log(\phi_{1} \cdot f_{\theta}(x))|} \\
\end{split}
\end{equation}

\subsection{Dual Point Analysis}

Now that a framework for deriving the behavior of a loss function has been established, we can analyze $\Loss_{ACE}$ from two unique and significant points in the learning process: (1) initial learning behavior at the null epoch when learning begins under a random initialization, and (2) in the zero training error regime where a loss function's regularization behaviors can be observed when there is nothing new to learn from the training data.

\subsubsection{Behavior at the Null Epoch}\label{sec:null}

First, consider the case where the base model weights $\theta_{0}$ are randomly initialized (\textit{i.e.}, the null epoch before any learned has happened) such that the expected value of the $i^{th}$ output of the base model is given as
\begin{equation}
\forall k \in [1, \mathcal{C}], \text{where } \mathcal{C} > 2: \mathbb{E} [f_{\theta}(x)_{i}] = \mathcal{C}^{-1}
\end{equation}
where $\mathcal{C}$ is the number of classes. Note, that the output $f_{\theta}(x)_{i}$ represents the post-softmax output which converts the raw logits into probabilities. Hence, the predicted probabilities are uniformly distributed among the classes $\frac{1}{\mathcal{C}} = \mathcal{C}^{-1}$.

\subsubsection{Behavior at Zero Training Error}\label{sec:zero}

Second, the zero training error regime is considered, which explores what happens when there is nothing left to learn from the training data. This regime is of great interest as the regularization behavior (or lack thereof) can be observed and contrasted. The most obvious way to analyze the zero training error case is to substitute the true target $y_{i}$ for the predicted output $f_{\theta}(x)_{i}$; however, this can sometimes result in indeterminates (\textit{i.e.}, $0/0$). Therefore, we instead consider the case where we approach the zero training error regime, \textit{i.e.}, where we get $\epsilon$ close to the true label. Since all the outputs sum to $1$, let

\begin{equation}
\mathbb{E}[f_{\theta}(x)_i] =
\begin{cases}
\epsilon, & y_{i} = 0 \\
1 - \epsilon(\mathcal{C} - 1), & y_{i} = 1
\end{cases}
\end{equation}

\subsection{Relationship to Cross-Entropy Loss}

\begin{figure*}[t!]
\centering
\begin{subfigure}{0.4\textwidth}
\centering
\includegraphics[width=1\textwidth]{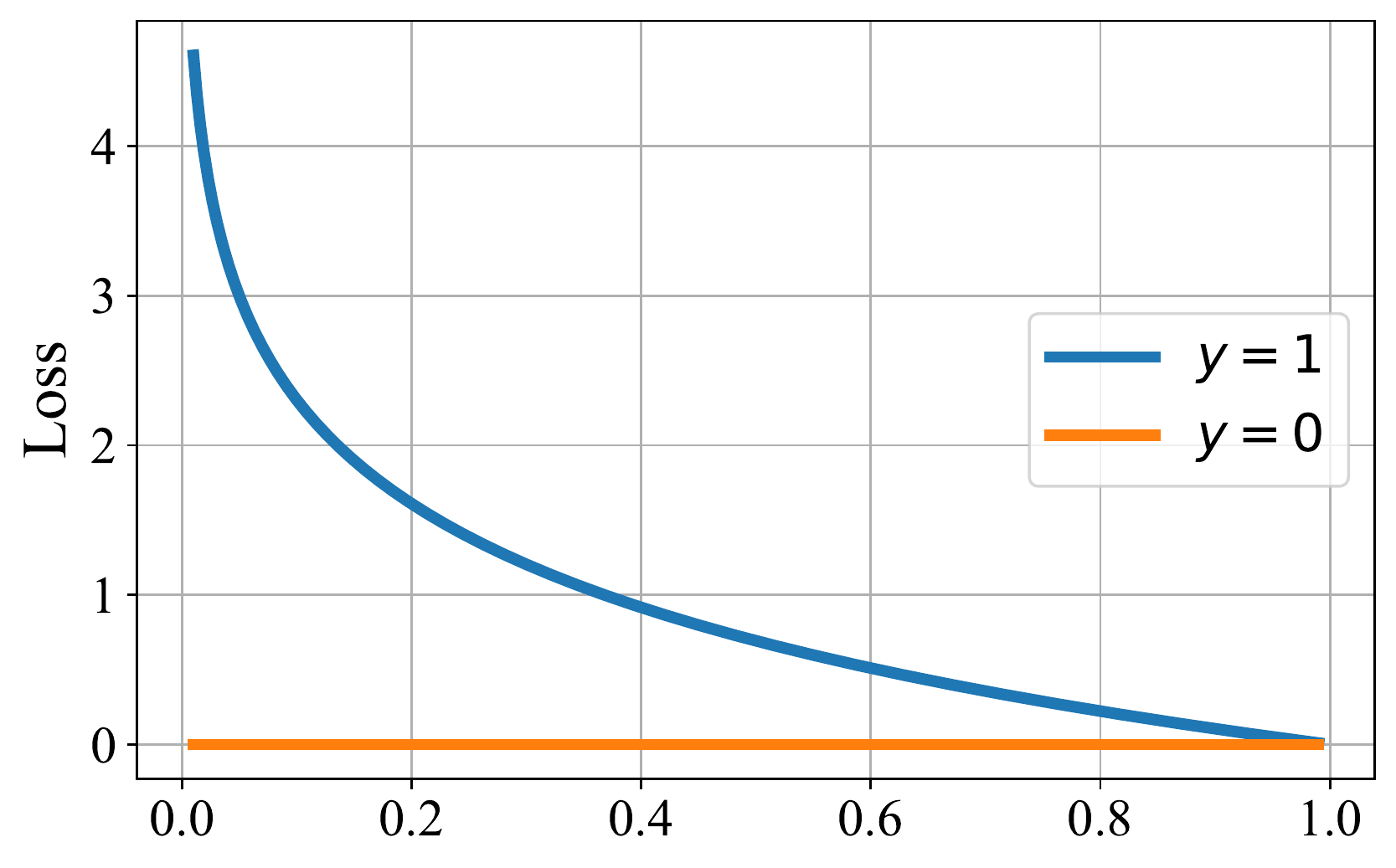}
\end{subfigure}
\begin{subfigure}{0.4\textwidth}
\centering
\includegraphics[width=1\textwidth]{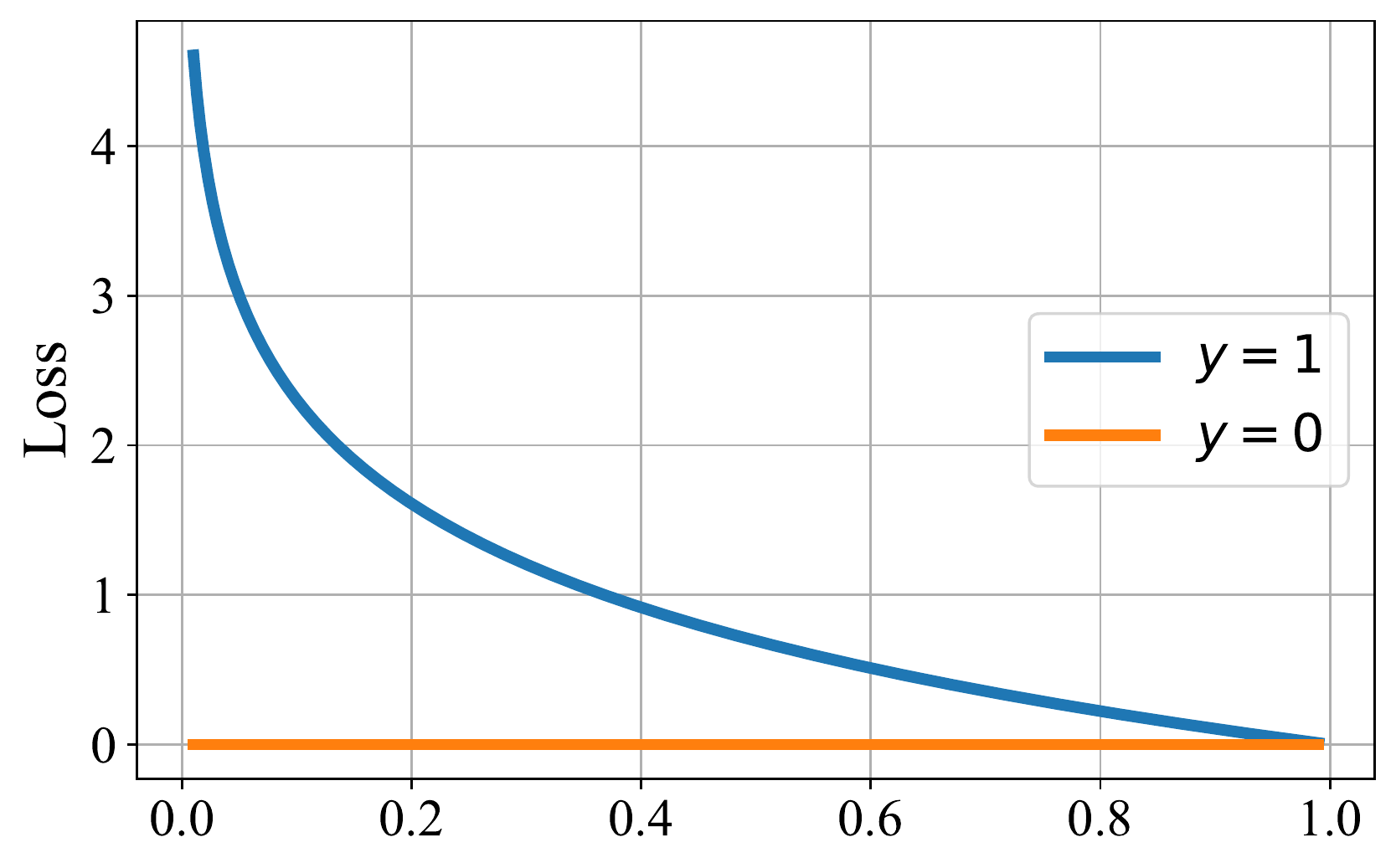}
\end{subfigure}

\captionsetup{justification=centering}
\caption{Contrasting the \textit{Cross-Entropy Loss} (left) to the \textit{Absolute Cross-Entropy Loss} (right), \\ where the x-axis represents the model's predictions $f_{\theta}(x)$.}
\label{fig:learned-loss-function}
\end{figure*}

From Figure \ref{fig:meta-learned-loss-functions} in Section \ref{sec:loss-functions} it is shown that the absolute cross-entropy loss ($\Loss_{ACE}$) bears a close resemblance to the widely used and \textit{de facto} standard cross-entropy loss ($\mathcal{L}_{CE}$). However, as we will show they are not just similar, they are equivalent when using a default parameterization, \textit{i.e.}, $\phi_0, \phi_1 = 1$, at the null epoch and when approaching the zero training error regime.

\subsubsection{Cross-Entropy Loss Decomposition}

\noindent
First decomposing the cross-entropy learning rule via substitution of the loss into Equation \eqref{eq:general-form} we get the following rule: 
\begin{equation}
\theta_{t+1} \leftarrow \theta_{t} + \alpha \bigg[\underbrace{\frac{\partial}{\partial f} \bigg(y_{i} \cdot \log(f_{\theta}(x)_{i})\bigg)}_{\delta_{\mathcal{L}_{CE}}} \frac{\partial}{\partial \theta_{j}} f_{\theta}(x)_{i}\bigg]
\end{equation}
where the behavior is defined as
\begin{equation}
    \delta_{\mathcal{L}_{CE}} = \frac{y_{i}}{f_{\theta}(x)_{i}}.
\end{equation}

\subsubsection{Behavior at the Null Epoch}

Subsequently, behavior at the null epoch can then be defined for $\mathcal{L}_{CE}$ piecewise for target vs non-target outputs as
\begin{equation}
\begin{split}
\delta_{\mathcal{L}_{CE}} &=
\begin{cases}
\frac{0}{\mathcal{C}^{-1}}, & y_{i} = 0 \\
\frac{1}{\mathcal{C}^{-1}}, & y_{i} = 1
\end{cases} \\
&=
\begin{cases}
0, & y_{i} = 0 \\
\mathcal{C}, & y_{i} = 1
\end{cases}
\end{split}
\label{eq:ce-behavior-null}
\end{equation}
This shows that when $y_{i}=1$ the target output value is maximized, while the non-target output values remain the same; however, due to the base model's softmax output activation function, the target output value being maximized will in turn minimize the non-target outputs. Notably, this is identical behavior at the null epoch to $\Loss_{ACE}$,
\begin{equation}
\begin{split}
\delta_{\Loss_{ACE}} &= 
\begin{cases}
-\frac{0}{\mathcal{C}^{-1} \cdot |\log(\phi_{1} \cdot \mathcal{C}^{-1})|}, & y_{i} = 0 \\
-\frac{\log(\phi_{1} \cdot \mathcal{C}^{-1})}{\mathcal{C}^{-1} \cdot |\log(\phi_{1} \cdot \mathcal{C}^{-1})|}, & y_{i} = 1
\end{cases} \\
&= 
\begin{cases}
-\frac{0}{\mathcal{C}^{-1} \cdot |\log(\mathcal{C}^{-1})|}, & y_{i} = 0 \\
\frac{-1 \cdot \log(\mathcal{C}^{-1})}{\mathcal{C}^{-1} \cdot (-1 \cdot \log(\mathcal{C}^{-1}))}, & y_{i} = 1
\end{cases} \\
&= 
\begin{cases}
0, & y_{i} = 0 \\
\mathcal{C}, & y_{i} = 1
\end{cases}
\end{split}
\label{eq:learned-behavior-null}
\end{equation}

\subsubsection{Behavior at Zero Training Error}

On the other extreme, the behavior of the cross-entropy loss ($\mathcal{L}_{CE}$) when approaching zero training error can be defined piecewise for target vs non-target outputs as follows
\begin{equation}
\begin{split}
\delta_{\mathcal{L}_{CE}} &= 
\begin{cases}
y_{i}/\epsilon, & y_{i} = 0 \\
y_{i}/(1 - \epsilon(\mathcal{C} - 1)), & y_{i} = 1
\end{cases} \\
&=\lim_{\epsilon\rightarrow 0}
\begin{cases}
0, & y_{i} = 0 \\
1, & y_{i} = 1
\end{cases}
\end{split}
\label{eq:ce-behavior-zero}
\end{equation}
The target output value is again maximized, while the non-target output values are minimized due to the softmax. This behavior is replicated by $\Loss_{ACE}$:
\begin{equation}
\begin{split}
\delta_{\Loss_{ACE}} &= 
\begin{cases}
-\frac{0}{\epsilon \cdot |\log(\phi_{1} \cdot \epsilon)|}, & y_{i} = 0 \\
-\frac{\log(\phi_{1} \cdot (1 - \epsilon(\mathcal{C} - 1))))}{(1 - \epsilon(\mathcal{C} - 1)) \cdot |\log(\phi_{1} \cdot (1 - \epsilon(\mathcal{C} - 1)))|}, & y_{i} = 1
\end{cases} \\
&= 
\begin{cases}
-\frac{0}{\epsilon \cdot |\log(\phi_{1} \cdot \epsilon)|}, & y_{i} = 0 \\
\frac{-1 \cdot \log(\phi_{1} \cdot (1 - \epsilon(\mathcal{C} - 1))))}{(1 - \epsilon(\mathcal{C} - 1)) \cdot (-1 \cdot \log(\phi_{1} \cdot (1 - \epsilon(\mathcal{C} - 1))))}, & y_{i} = 1
\end{cases} \\
&=\lim_{\epsilon\rightarrow 0}
\begin{cases}
0, & y_{i} = 0 \\
1, & y_{i} = 1
\end{cases}
\end{split}
\label{eq:learned-behavior-zero}
\end{equation}

\subsubsection{Result Discussion}

\noindent
These results show that in the extreme limits of training (\textit{i.e.}, very start and very end of training), the absolute cross-entropy loss with a default parameterization $\phi_{0}, \phi_{1} = 1$ is equivalent in learning behavior to the cross-entropy loss. At the null epoch, both loss functions will maximize the target and minimize the non-target (due to the softmax). Approaching zero training error, this behavior will continue, with the target output value being maximized and the non-target outputs minimized. This result can be validated visually as shown in Figure \ref{fig:learned-loss-function}. This is a notable result as it shows that EvoMAL was able to discover the cross-entropy loss function directly from the data.

\subsection{Relationship to Label Smoothing Regularization}
\label{sec:label-smoothing-regularization}

\begin{figure*}[t!]
\centering
\begin{subfigure}{0.4\textwidth}
\centering
\includegraphics[width=1\textwidth]{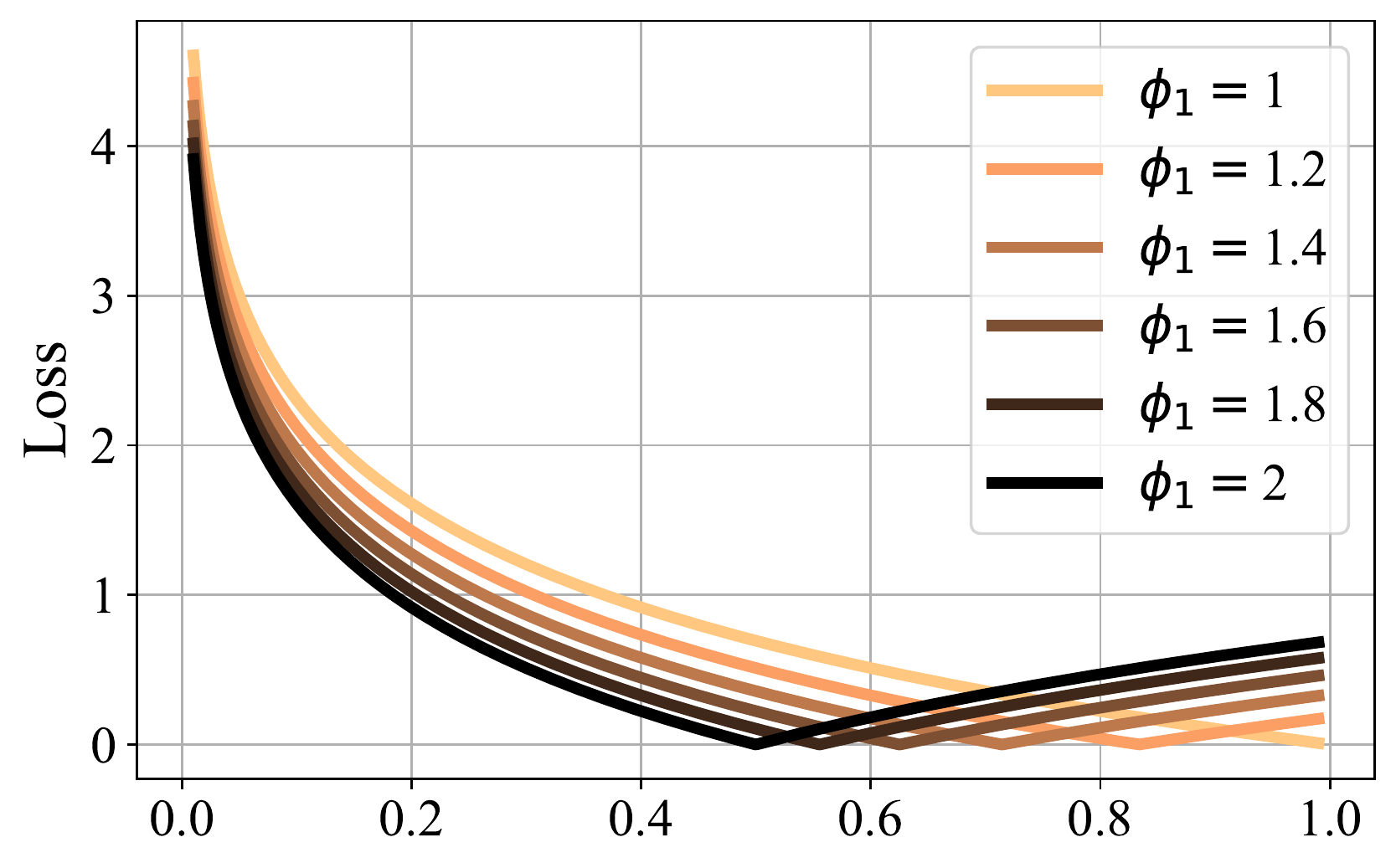}
\end{subfigure}
\begin{subfigure}{0.4\textwidth}
\centering
\includegraphics[width=1\textwidth]{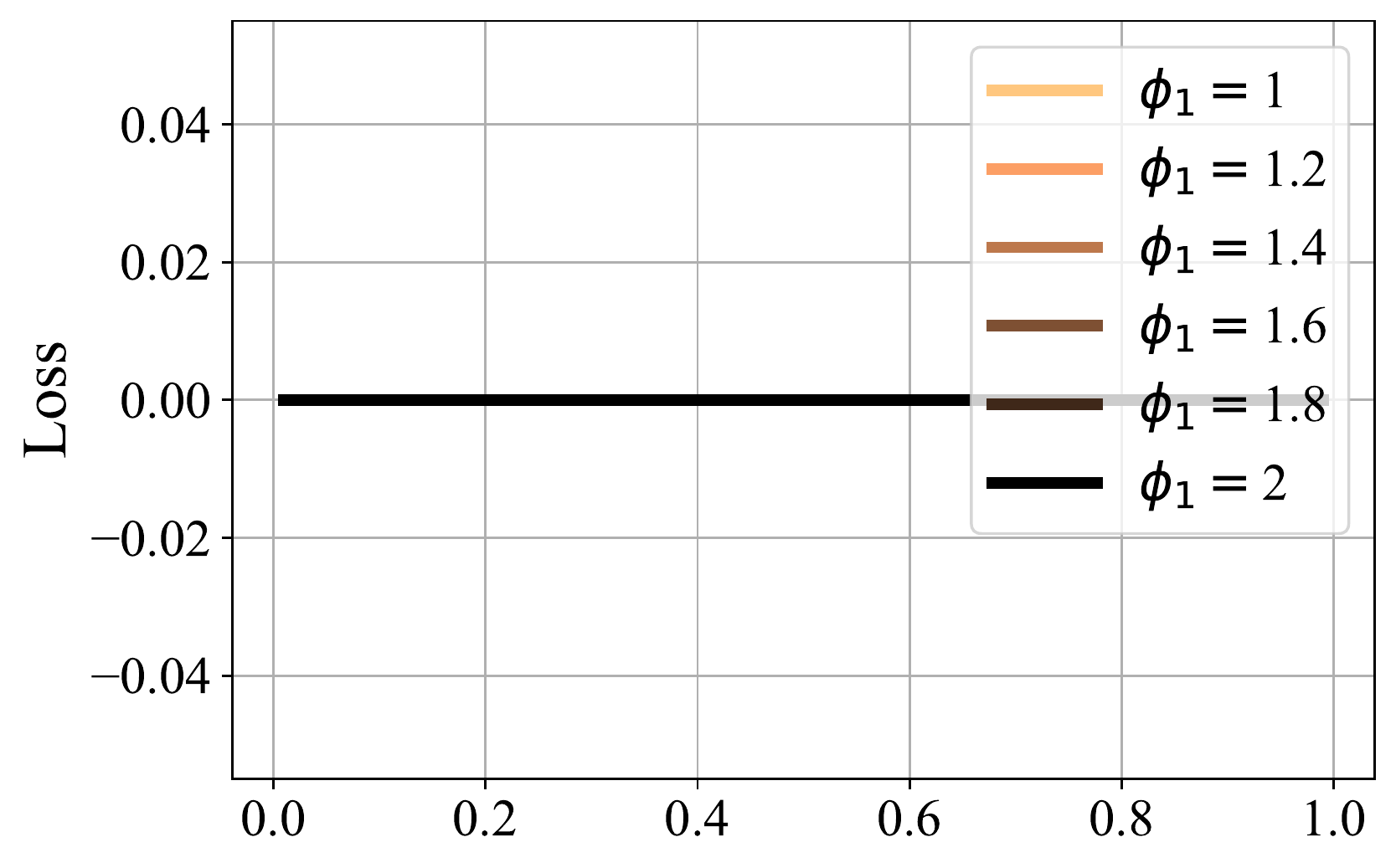}
\end{subfigure}
\captionsetup{justification=centering}
\caption{Visualizing the proposed \textit{Absolute Cross-Entropy Loss}, with varying values of $\phi_1$, where the left figure \\ shows the target loss ($y_i = 1$), and the right figure shows the non-target loss ($y_i = 0$).}
\label{fig:varying-learned-loss-function}
\end{figure*}

\begin{figure*}[t!]
\centering
\begin{subfigure}{0.4\textwidth}
\centering
\includegraphics[width=1\textwidth]{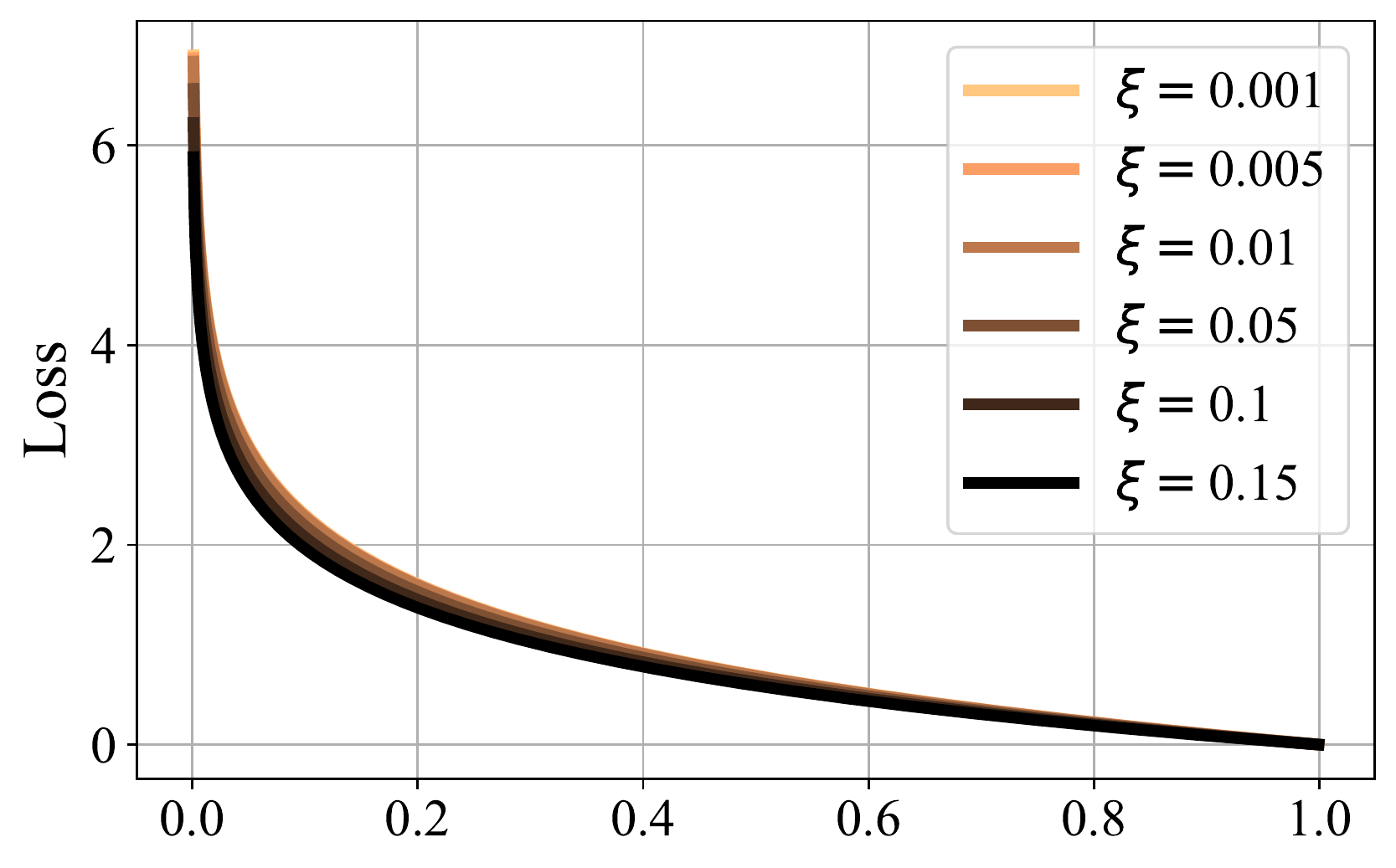}
\end{subfigure}%
\hspace{5mm}
\begin{subfigure}{0.4\textwidth}
\centering
\includegraphics[width=1\textwidth]{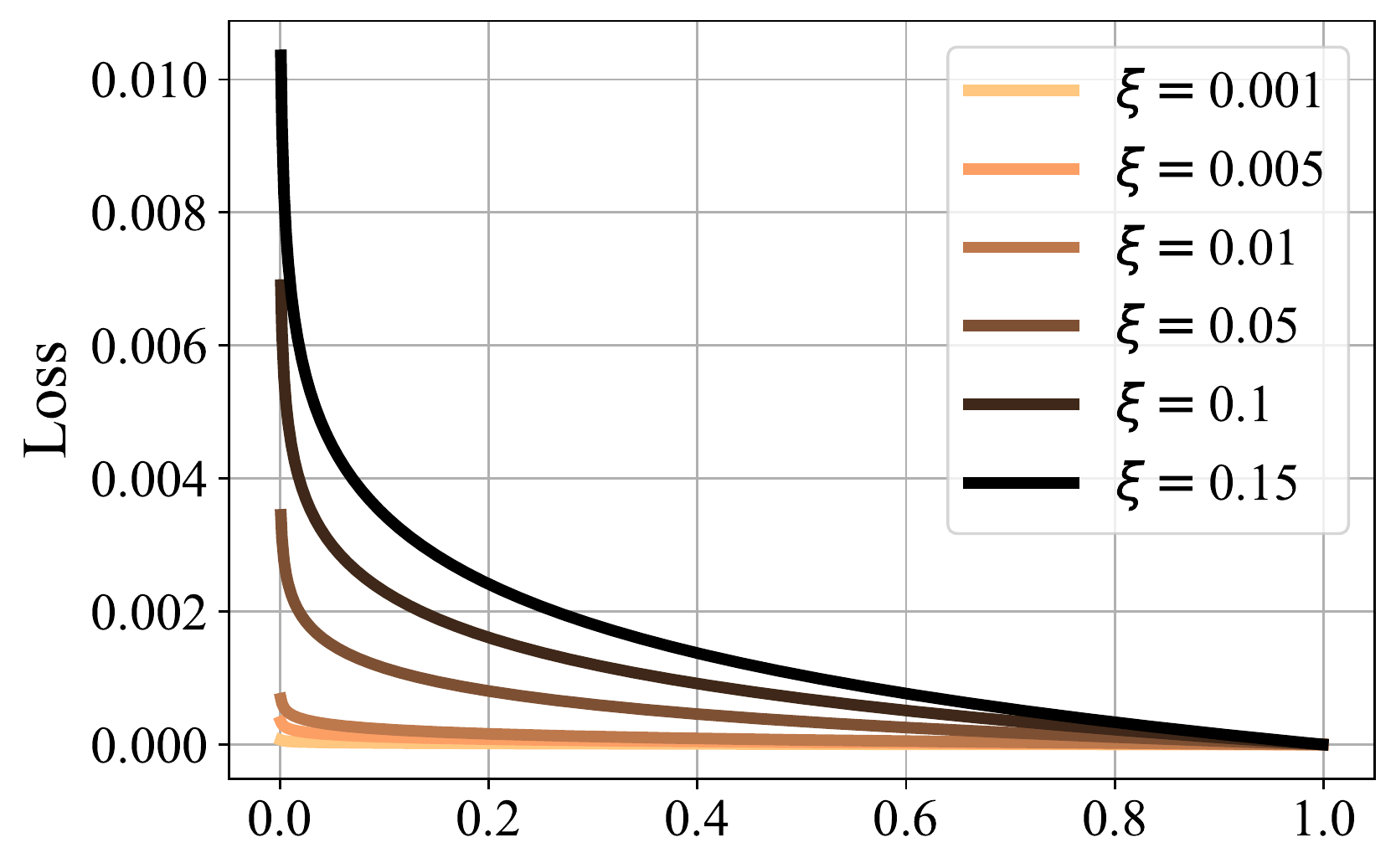}
\end{subfigure}%

\captionsetup{justification=centering}
\caption{Visualizing \textit{Label Smoothing Regularization Loss} with varying smoothing values $\xi$, where the left figure \\ shows the target loss ($y_i = 1$), and the right figure shows the non-target loss ($y_i = 0$).}
\label{fig:label-smoothing-loss}
\end{figure*}

A key aspect of the absolute cross-entropy loss, particularly evident in loss function (e) in Figure \ref{fig:meta-learned-loss-functions}, is its perplexing tendency to counterintuitively push the base model's prediction $f_{\theta}(x)_i$ away from 1 as it approaches close to 1. Functionally, this phenomenon aims to prevent a model from becoming too overconfident in its predictions, which can lead to poor generalization on new unseen testing data. Mitigating overconfidence is crucial for developing robust models that can generalize outside the training data.

This behavior is very similar to \textit{Label Smoothing Regularization} ($\Loss_{LSR}$) \cite{szegedy2016rethinking, muller2019does}, a widely-used regularization technique used in deep neural networks that transforms hard targets into soft targets. Label smoothing regularization adjusts target labels according to the formula $y_{i} \leftarrow y_{i}(1 - \xi) + \xi/\mathcal{C}$, where $1 > \xi > 0$ is the smoothing coefficient used to calibrate the regularization pressure. By softening the target labels $\Loss_{LSR}$ reduces the confidence of the model, which consequently helps to mitigate overfitting and enhance generalization. This technique has been used to improve model robustness and generalization across various domains such as computer vision and natural language processing.

As shown in the forthcoming derivations, the absolute cross-entropy loss and label smoothing regularization ($\Loss_{LSR}$) exhibit similarities in their effects on model predictions. Both techniques maximize the target outputs and minimize the non-target outputs, particularly at the initial training stage (null epoch) and as the training error approaches zero. These effects are observed under the conditions where $\phi_0=1$ and $2 > \phi_1 > 1$

\subsubsection{Label Smoothing Regularization Decomposition}

First decomposing the label smoothing learning rule via substitution of the loss into Equation \eqref{eq:general-form} we get the following rule:
\begin{equation}
\theta_{t+1} \leftarrow \theta_{t} + \alpha \bigg[\underbrace{\frac{\partial}{\partial f} \bigg(\left(y_{i}(1-\xi) + \frac{\xi}{\mathcal{C}}\right) \log(f_{\theta}(x)_{i})\bigg)}_{\delta_{\mathcal{L}_{LSR}}} \frac{\partial}{\partial \theta_{j}} f_{\theta}(x)_{i}\bigg]
\end{equation}
where the behavior is defined as follows:
\begin{equation}
\begin{split}
\delta_{\mathcal{L}_{LSR}} 
&= \frac{\partial}{\partial f} \left(\left(y_{i}(1-\xi) + \frac{\xi}{\mathcal{C}}\right) \log(f_{\theta}(x)_{i})\right)\\
&= \left(y_{i}(1-\xi) + \frac{\xi}{\mathcal{C}}\right) \cdot \frac{\partial}{\partial f} \left(\log(f_{\theta}(x)_{i})\right)\\
&= \left(y_{i}(1-\xi) + \frac{\xi}{\mathcal{C}}\right) \cdot \frac{1}{f_{\theta}(x)_i}\\
&= \frac{y_{i}(1 - \xi) + \xi/\mathcal{C}}{f_{\theta}(x)_{i}} \\
\end{split}
\end{equation}

\subsubsection{Behavior at the Null Epoch}

Behavior at the null epoch can then be defined for label smoothing regularization ($\mathcal{L}_{LSR}$) piecewise for target vs non-target outputs as
\begin{equation}\label{eq:label-smoothing-null-behavior}
\begin{split}
\delta_{\mathcal{\Loss}_{LSR}} &= 
\begin{cases}
\frac{\xi/\mathcal{C}}{\mathcal{C}^{-1}}, & y_{i} = 0 \\
\frac{1 - \xi + \xi/\mathcal{C}}{\mathcal{C}^{-1}}, & y_{i} = 1
\end{cases} \\
&= 
\begin{cases}
\xi, & y_{i} = 0 \\
\mathcal{C} - \mathcal{C}\xi + \xi, & y_{i} = 1
\end{cases}
\end{split}
\end{equation}
where the $y_{i}=1$ case is shown to dominate $y_{i}=0$ since
\begin{equation}
\xi < \mathcal{C} - \mathcal{C}\xi + \xi.
\end{equation}
This result shows that when $y_{i}=1$ the target output value will be maximized, while the non-target output values will be minimized due to the base model's softmax activation function. This is as expected, both consistent with the behavior of the cross-entropy, Equation \eqref{eq:ce-behavior-null}, as well as with the absolute cross-entropy, where $\phi_1=1$, Equation \eqref{eq:learned-behavior-null}. Furthermore, as shown this behavior is also consistent with the behavior of the absolute cross-entropy where $2>\phi_1>1$, since

\begin{equation}
\begin{split}
\delta_{\Loss_{ACE}} &= 
\begin{cases}
-\frac{0}{\mathcal{C}^{-1} \cdot |\log(\phi_{1} \cdot \mathcal{C}^{-1})|}, & y_{i} = 0 \\
-\frac{\log(\phi_{1} \cdot \mathcal{C}^{-1})}{\mathcal{C}^{-1} \cdot |\log(\phi_{1} \cdot \mathcal{C}^{-1})|}, & y_{i} = 1
\end{cases} \\
&= 
\begin{cases}
-\frac{0}{\mathcal{C}^{-1} \cdot |\log(\phi_{1} \cdot \mathcal{C}^{-1})|}, & y_{i} = 0 \\
\frac{-1 \cdot \log(\phi_{1} \cdot \mathcal{C}^{-1})}{\mathcal{C}^{-1} \cdot (-1 \cdot \log(\phi_{1} \cdot \mathcal{C}^{-1}))}, & y_{i} = 1
\end{cases} \\
&= 
\begin{cases}
0, & y_{i} = 0 \\
\mathcal{C}, & y_{i} = 1
\end{cases}
\end{split}
\end{equation}

\subsubsection{Behavior at Zero Training Error}

As label smoothing regularization is a regularization technique that penalizes overconfidence, we expect to see noteworthy results when approaching zero training error. The regularization behavior of $\delta_{LSR}$ as the zero training error case is approached:
\begin{equation}\label{eq:label-smoothing-behavior}
\begin{split}
\delta_{\mathcal{L}_{LSR}} &= 
\begin{cases}
\frac{\xi/\mathcal{C}}{\epsilon}, & y_{i} = 0 \\
\frac{1 - \xi + \xi/\mathcal{C}}{1 - \epsilon(\mathcal{C} - 1)}, & y_{i} = 1
\end{cases} \\
&=\lim_{\epsilon\rightarrow 0}
\begin{cases}
+\infty, & y_{i} = 0 \\
1 - \xi + \frac{\xi}{\mathcal{C}}, & y_{i} = 1
\end{cases}
\end{split}
\end{equation}
where the $y_{i}=0$ case is shown to dominate $y_{i}=1$ since
\begin{equation}
+\infty > 1 - \xi + \frac{\xi}{\mathcal{C}}.
\end{equation}
Contrary to the prior results we can see that as the zero training error case is approached the non-target output is maximized while the target output is minimized. This result clearly demonstrates the underlying mechanism for how label smoothing regularization penalizes overconfidence and improves generalization. A noteworthy discovery that we put forward is that this underlying mechanism is also apparent in $\Loss_{ACE}$. Let $\phi_0=1$ and $2 > \phi_1 > 1$, then:
\begin{equation}
\begin{split}
\delta_{\Loss_{ACE}} &= 
\begin{cases}
-\frac{0}{\epsilon|\log(\phi_{1} \cdot \epsilon)|}, & y_{i} = 0 \\
-\frac{\log(\phi_{1}(1 - \epsilon(\mathcal{C} - 1) + \epsilon))}{(1 - \epsilon(\mathcal{C} - 1) + \epsilon)|\log(\phi_{1}(1 - \epsilon(\mathcal{C} - 1) + \epsilon))|}, & y_{i} = 1 \\
\end{cases} \\
&= 
\begin{cases}
-\frac{0}{\epsilon|\log(\phi_{1} \cdot \epsilon)|}, & y_{i} = 0 \\
\frac{-1 \cdot \log(\phi_{1}(1 - \epsilon(\mathcal{C} - 1) + \epsilon))}{(1 - \epsilon(\mathcal{C} - 1) + \epsilon)\cdot\log(\phi_{1}(1 - \epsilon(\mathcal{C} - 1) + \epsilon))}, & y_{i} = 1 \\
\end{cases} \\
&=\lim_{\epsilon\rightarrow 0}
\begin{cases}
0, & y_{i} = 0 \\
-1, & y_{i} = 1 \\
\end{cases}
\end{split}
\end{equation}

\subsubsection{Results Discussion}

\begin{figure*}[t!]
\centering

\begin{subfigure}{0.35\textwidth}
\includegraphics[width=1\textwidth]{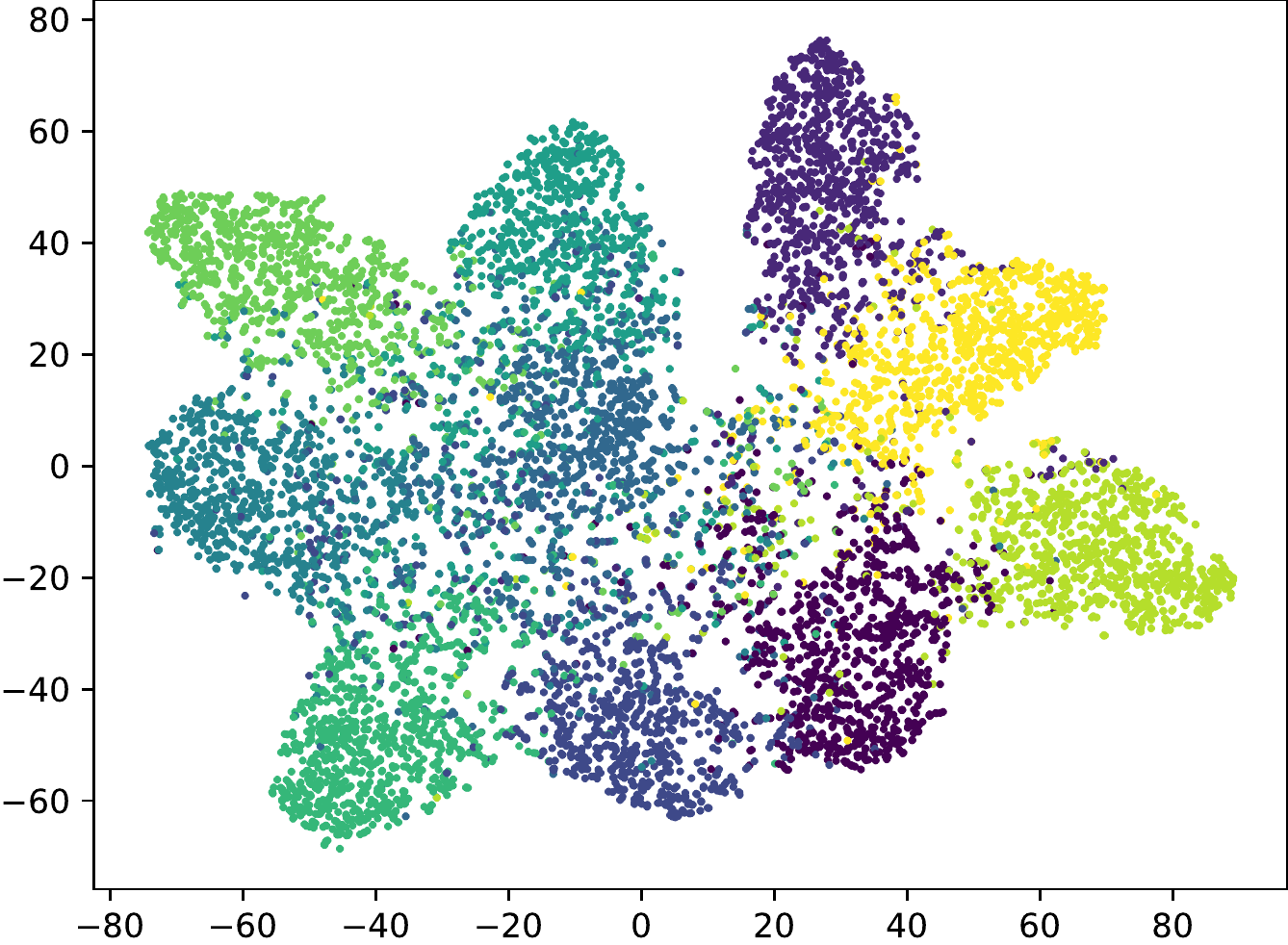}
\caption{Cross-Entropy -- 15.66\%.}
\end{subfigure}%
\hspace{5mm}
\begin{subfigure}{0.35\textwidth}
\includegraphics[width=1\textwidth]{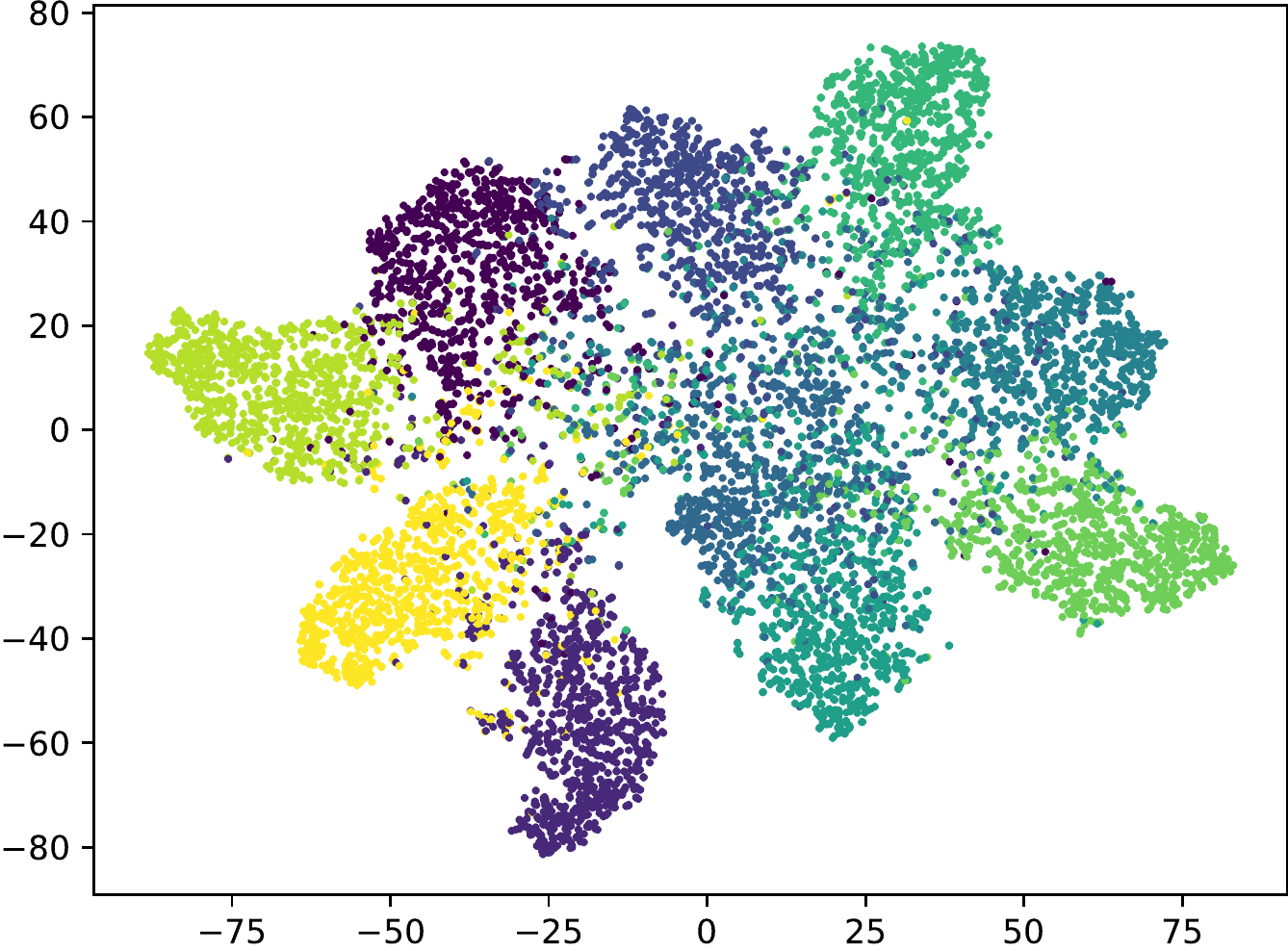}
\caption{Absolute CE ($\phi_{1}=1$) -- 15.43\%.}
\end{subfigure}%
\par\bigskip
\begin{subfigure}{0.35\textwidth}
\includegraphics[width=1\textwidth]{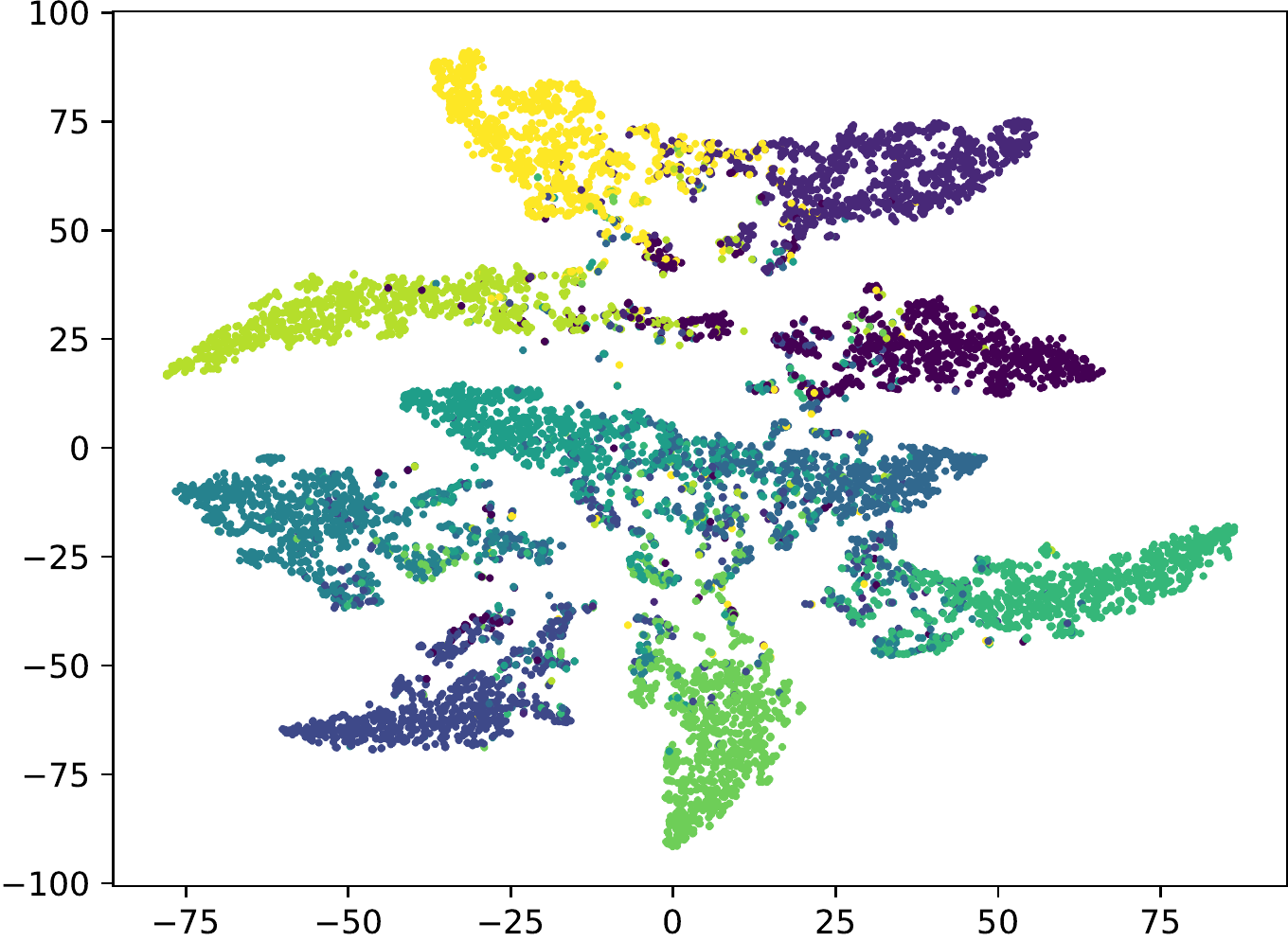}
\caption{Label Smoothing ($\xi = 0.1$) -- 14.99\%}
\end{subfigure}
\hspace{5mm}
\begin{subfigure}{0.35\textwidth}
\includegraphics[width=1\textwidth]{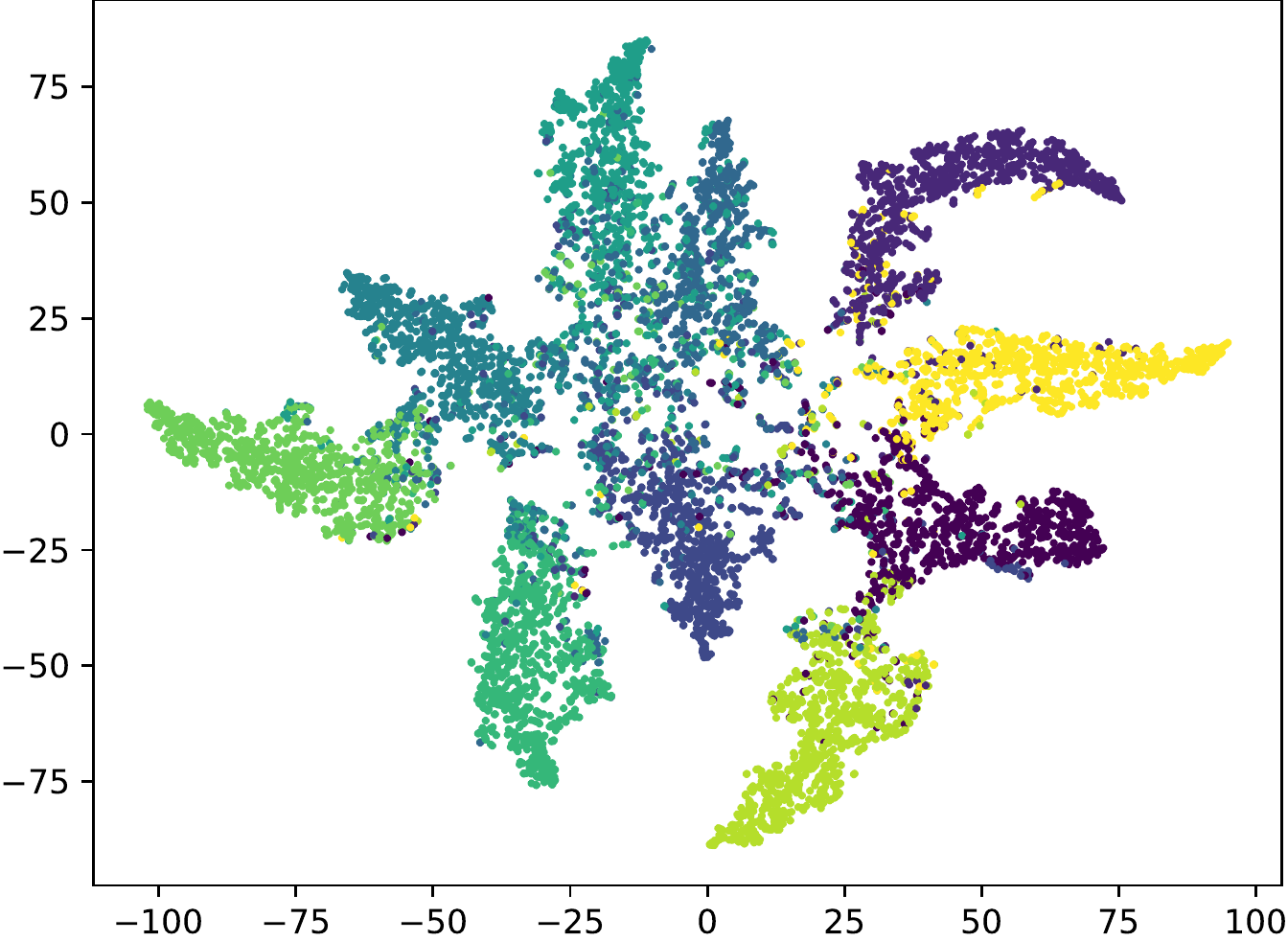}  
\caption{Absolute CE ($\phi_{1}=1.1$) -- 14.92\%}
\end{subfigure}

\captionsetup{justification=centering}
\caption{The penultimate layer representations on AlexNet CIFAR-10, using t-SNE for dimensionality reduction. \\ Comparing \textit{Absolute Cross-Entropy Loss} (Absolute CE) where $\phi_1 = \{1, 1.1\}$, to \textit{Cross-Entropy Loss} and \\ \textit{Label Smoothing Regularization Loss}. Color represents the class of an instance.}
\label{fig:learned-respresentations}
\end{figure*}

The results show the functional equivalence in behavior between label smoothing regularization and the absolute cross-entropy loss where $2 > \phi_{1} > 1$. At the start of training where the model weights $\theta$ are randomized $\delta_{\Loss_{ACE}}$ is shown to be identical to $\delta_{\Loss_{LSR}}$ and $\delta_{\Loss_{CE}}$, maximizing the target and due to the softmax minimizing the non-target. In contrast, the results presented in Equation \eqref{eq:learned-behavior-zero} show that when approaching zero training error $\delta_{\Loss_{ACE}}$ has the same unique behavior as $\delta_{\Loss_{LSR}}$, forcing the non-target outputs to be maximized while the target output is minimized. Contrary to a typical loss function, $\Loss_{LSR}$ and $\Loss_{ACE}$ aim to maximize the divergence between overly confident prediction between the predicted distribution and the true target distribution.

The parity in behavior between $\Loss_{LSR}$ and $\Loss_{ACE}$ can be seen in Figures \ref{fig:varying-learned-loss-function} and \ref{fig:label-smoothing-loss}, respectively. This is a notable result as it shows that EvoMAL was able to discover label smoothing regularization directly from the data, with no prior knowledge of the regularization technique being given to the method. Furthermore, although similar in behavior, it is noteworthy that $\Loss_{ACE}$ is “\textit{sparse}” as it only needs to be computed on the target output \textit{i.e.}, constant time and space complexity $\Theta(1)$, with respect to the number of classes $\mathcal{C}$, as it is 0 for all non-target outputs, as shown in Equations \eqref{eq:learned-behavior-null} and \eqref{eq:learned-behavior-zero} when $y_i=0$. In contrast, label smoothing regularization must be applied to both the target output and all non-target outputs \textit{i.e.}, linear time and space complexity $\Theta(\mathcal{C})$, with respect to the number of classes $\mathcal{C}$, as shown in Equation \eqref{eq:label-smoothing-behavior}. Hence, the learned loss function can be computed significantly faster than the cross-entropy with label smoothing.

\subsection{Further Analysis}

To validate our theoretical findings we visualize and analyze the penultimate layer representations of AlexNet on CIFAR-10 using t-distributed Stochastic Neighbor Embedding (t-SNE) \cite{van2008visualizing}, when trained with the cross-entropy, label smoothing regularization, and the absolute cross-entropy with $\phi_{1}=\{1, 1.1\}$. The results are presented in Figure \ref{fig:learned-respresentations}. It shows that the representations learned by (a) the cross-entropy and (b) absolute cross-entropy when $\phi=1$ are visually very similar, supporting the findings in Sections \ref{sec:null} and \ref{sec:zero}. Furthermore, analyzing the learned representations of (c) label smoothing regularization where $\xi = 0.1$ and (d) the absolute cross-entropy where $\phi_{1} = 1.1$ it is shown that both the learned representations have better discriminative inter-class representations and tighter intra-class representations. This empirical analysis supports our findings from the theoretical analysis in Section \ref{sec:theoretical-analysis}, showing consistent regularization results between label smoothing regularization and the absolute cross-entropy loss when $\phi_{1} > 1$.

The results of our meta-learned loss function thus far have shown to be very positive; nonetheless, we would like to draw our readers' attention to two prominent limitations of the absolute cross-entropy loss function. Firstly, $\Loss_{ACE}$ has shown to be equivalent in behavior to $\Loss_{LSR}$ at the null epoch and when approaching zero training error, \textit{i.e.}, at the start and very end of training; however, for intermediate points, the behavior is not identical. Secondly, $\Loss_{ACE}$ is not smooth as shown in Figure \ref{fig:varying-learned-loss-function} when transitioning between minimizing and maximizing the target output. Consequently, this can cause training to become unstable, disrupting the learning process and negatively affecting the generalization of the base model. In the following section, we apply our findings and discoveries from the absolute cross-entropy loss to propose a new loss function that resolves both of these issues while maintaining the key properties of $\Loss_{ACE}$, specifically, its constant time and space complexity due to target-only loss computation.

\subsection{Sparse Label Smoothing Regularization}
\label{sec:sparse-label-smoothing-regularization}

Label Smoothing Regularization (LSR) \cite{goodfellow2016deep, muller2019does} is a popular and effective technique for preventing overconfidence in a classification model. This technique achieves regularization by penalizing overconfident predictions through adjusting the target label $y$ during training such that a $y_{i} \leftarrow y_{i}(1 - \xi) + \xi/\mathcal{C}$, where $1 > \xi > 0$ is the smoothing coefficient. LSR forces the development of more robust and generalizable features within the model, as it discourages the model from being overly confident in its predictions and forces it to consider alternative possibilities, ultimately improving generalization performance. The cross-entropy loss when combined with label smoothing regularization is defined as follows:
\begin{equation}
\resizebox{0.9\columnwidth}{!}{$
\begin{split}
    \Loss_{LSR}  
    &= - \sum_{i=1}^{\mathcal{C}} \left(y_{i} \cdot (1 - \xi) + \frac{\xi}{\mathcal{C}}\right) \cdot \log(f_{\theta}(x)_{i}) \\
    &= - \underbrace{\sum_{i=1, y_{i} = 1}^{\mathcal{C}} \left(1 - \xi + \frac{\xi}{\mathcal{C}}\right) \cdot \log(f_{\theta}(x)_{i})}_{\text{Target Loss}} + \underbrace{\sum_{i=1, y_{i} = 0}^{\mathcal{C}} \frac{\xi}{\mathcal{C}} \cdot \log(f_{\theta}(x)_{i})}_{\text{Non-Target Losses}} \\
\end{split}
$}
\label{eq:label-smoothing-regularization}
\end{equation}
In contrast to the cross-entropy loss which is typically implemented as a “\textit{sparse}” loss function, \textit{i.e.}, only computes the loss on the target output as the non-target loss is $0$, label smoothing regularization is a “\textit{non-sparse}” loss function as it has a non-zero loss for both the target and non-target outputs. Consequently, this results in having to do a scalar expansion on the target\footnote{In automatic differentiation/neural network libraries such as \texttt{PyTorch} the target labels are by default represented as a vector of target indexes $y \in \mathbb{Z}^{B} \cap [0, \mathcal{C}]$; therefore, non-sparse loss functions such as label smoothing regularization require one-hot encoding of the target prior to computing the loss function.}, or more commonly a vector to matrix (column) expansion when considering a batch of instances, by one-hot encoding the labels. 

The label smoothing loss can subsequently be calculated using the one-hot encoded target matrix $y \in \{0, 1\}^{\mathcal{C} \times B}$ and the model predictions $f_{\theta}(x) \in (0, 1)^{\mathcal{C} \times B}$, where the summation of the resulting losses is taken across the $\mathcal{C}$ rows before being reduced in the $B$ batch dimension using some aggregation function such as the arithmetic mean. 

\subsection{Derivation of Sparse Label Smoothing}

Due to its non-sparse nature, label smoothing regularization has a time and space complexity of $\Theta(\mathcal{C})$, as opposed to $\Theta(1)$ for the cross-entropy (ignoring the $B$ batch dimensions, as the concept of loss is defined on a singular instance). However, by utilizing an innovative technique inspired by the behavior of the absolute cross-entropy loss analyzed in Section \ref{sec:theoretical-analysis}, which we call the “\textit{redistributed loss trick}”, we show a way to compute what we further refer to as \textit{Sparse Label Smoothing Regularization} (SparseLSR) loss, which is as the name implies sparse, and has a time and space complexity of $\Theta(1)$.

The key idea behind sparse label smoothing regularization is to utilize the redistributed loss trick, which redistributes the expected non-target loss into the target loss, obviating the need to calculate the loss on the non-target outputs. The redistributed loss trick can retain near identical behavior due to the output softmax function redistributing the gradients back into the non-target outputs during backpropagation. The sparse label smoothing regularization loss is defined as follows:
\begin{equation}
\begin{split}
    \Loss_{SparseLSR} 
    &= - \sum_{i=1, y_{i} = 1}^{\mathcal{C}} \left(1 - \xi + \frac{\xi}{\mathcal{C}}\right) \cdot \log(f_{\theta}(x)_{i}) \\ & + \mathbb{E} \left[\sum_{j=1, y_{i} = 0}^{\mathcal{C}} \frac{\xi}{\mathcal{C}} \cdot \log(f_{\theta}(x)_{j})\right] \\
\end{split}
\end{equation}
where the expectation is distributed across the summation, yielding the following simplified form:
\begin{equation}
\begin{split}
    \Loss_{SparseLSR} 
    &= - \sum_{i=1, y_{i} = 1}^{\mathcal{C}} \left(1 - \xi + \frac{\xi}{\mathcal{C}}\right) \cdot \log(f_{\theta}(x)_{i}) \\ & + \sum_{j=1, y_{j} = 0}^{\mathcal{C}} \frac{\xi}{\mathcal{C}} \cdot \mathbb{E} \left[\log(f_{\theta}(x)_{j})\right]. \\
\end{split}
\end{equation}
Following this, the expectation of the model's non-target output $\mathbb{E} \left[\log(f_{\theta}(x)_{j})\right]$ is approximated via a first-order Taylor-expansion, \textit{i.e.}, a linear approximation.
\begin{equation}
\resizebox{0.9\columnwidth}{!}{$
\begin{split}
     \mathbb{E} \left[\log(f_{\theta}(x)_{j})\right]
     &\approx \mathbb{E} \left[\log(\mathbb{E}\left[f_{\theta}(x)_{j}\right]) + \frac{1}{\mathbb{E}\left[f_{\theta}(x)_{j}\right]}(f_{\theta}(x)_{j} - \mathbb{E}\left[f_{\theta}(x)_{j}\right])\right] \\
     &= \mathbb{E} \left[\log(\mathbb{E}\left[f_{\theta}(x)_{j}\right])\right] + \mathbb{E} \left[\frac{1}{\mathbb{E}\left[f_{\theta}(x)_{j}\right]}(f_{\theta}(x)_{j} - \mathbb{E}\left[f_{\theta}(x)_{j}\right])\right] \\
     &= \log(\mathbb{E}\left[f_{\theta}(x)_{j}\right]) + \frac{1}{\mathbb{E}\left[f_{\theta}(x)_{j}\right]}\mathbb{E}\left[(f_{\theta}(x)_{j} - \mathbb{E}\left[f_{\theta}(x)_{j}\right])\right] \\
     &= \log(\mathbb{E}\left[f_{\theta}(x)_{j}\right])
\end{split}
$}
\label{eq:linear-approximation-of-log}
\end{equation}
This lets us rewrite the expectation in terms of $f_{\theta}(x)_j$ as follows:
\begin{equation}
\begin{split}
    \Loss_{SparseLSR} 
    &\approx - \sum_{i=1, y_{i} = 1}^{\mathcal{C}} \left(1 - \xi + \frac{\xi}{\mathcal{C}}\right) \cdot \log(f_{\theta}(x)_{i}) \\ & + \sum_{j=1, y_{i} = 0}^{\mathcal{C}} \frac{\xi}{\mathcal{C}} \cdot \log(\mathbb{E}\left[f_{\theta}(x)_{j}\right]) \\
\end{split}
\end{equation}
As defined by the softmax activation function, the summation of the model's output predictions is $\sum_{i = 1}^{\mathcal{C}}f_{\theta}(x)_i = 1$; therefore, the expected value of the non-target output prediction $\mathbb{E}[f_{\theta}(x)_j]$ where $y_j = 0$ can be given as $1-f_{\theta}(x)_i$ where $y_i = 1$ normalized over the number of non-target outputs $\mathcal{C}-1$. Substituting this result back into our expression gives the following:
\begin{equation}
\begin{split}
    \Loss_{SparseLSR} 
    &\approx - \sum_{i=1, y_{i} = 1}^{\mathcal{C}} \left(1 - \xi + \frac{\xi}{\mathcal{C}}\right) \cdot \log(f_{\theta}(x)_{i}) \\ & + \sum_{j=1, y_{i} = 0}^{\mathcal{C}} \frac{\xi}{\mathcal{C}} \cdot \log\left(\frac{1 - f_{\theta}(x)_i}{\mathcal{C}-1}\right) \\
\end{split}
\end{equation}
where the first conditional summation can be removed to make explicit that $\Loss_{SparseLSR}$ is only non-zero for the target output, \textit{i.e.}, where $y_i = 1$, and the second conditional summation can be removed to obviate recomputation of the non-target segment of the loss which is currently defined as the summation of a constant. The final definition of Sparse Label Smoothing Regularization ($\Loss_{SparseLSR}$) is \footnote{Note, the newly proposed sparse label smoothing can be simplified to $ - \sum_{i=1}^{\mathcal{C}} y_{i}\left[\log(f_{\theta}(x)_{i}) + \xi \cdot \log\left(1 - f_{\theta}(x)_i\right)\right]$ by dropping the dependence on the number of classes $\mathcal{C}$, \textit{i.e.} multiplicative constants which scale based on the number of classes. This is a more relaxed approximation of traditional label smoothing regularization, but in principle has the same identical behavior of regularizing overconfident predictions.}:
\begin{equation}\label{eq:sparse-label-smoothing-regularization}
\begin{split}
    \Loss_{SparseLSR} 
    &\approx - \sum_{i=1}^{\mathcal{C}} y_{i}\left[\left(1 - \xi + \frac{\xi}{\mathcal{C}}\right) \cdot \log(f_{\theta}(x)_{i}) \right. \\ & \left. + \frac{\xi(\mathcal{C}-1)}{\mathcal{C}} \cdot \log\left(\frac{1 - f_{\theta}(x)_i}{\mathcal{C}-1}\right)\right] 
\end{split}
\end{equation}
The proposed loss function $\Loss_{SparseLSR}$ is visualized in Figure \ref{fig:fast-label-smoothing-loss}. As shown it resolves the key limitations of $\Loss_{ACE}$; namely, the lack of smoothness and its divergent behavior from $\Loss_{LSR}$ for intermediate points between the null epoch and approaching zero training error. Importantly, $\Loss_{SparseLSR}$ maintains identical time and space complexity due to target-only loss computation, as shown in Figure \ref{fig:fast-label-smoothing-loss} where all non-target losses are zero.

Note, that the approximation of the non-target losses used in $\Loss_{SparseLSR}$ assumes that the remaining output probabilities are uniformly distributed among the non-target outputs. Although this is a crude assumption, as demonstrated in Section \ref{sec:sparse-label-smoothing-regularization-experiments} this does not affect the regularization performance. Intuitively, a small or even moderate amount of error in the approximation of the non-target loss in label smoothing regularization is insignificant, as we are approximating the value of an error term.

\begin{figure*}[t!]
\centering

\begin{subfigure}{0.4\textwidth}
\includegraphics[width=1\textwidth]{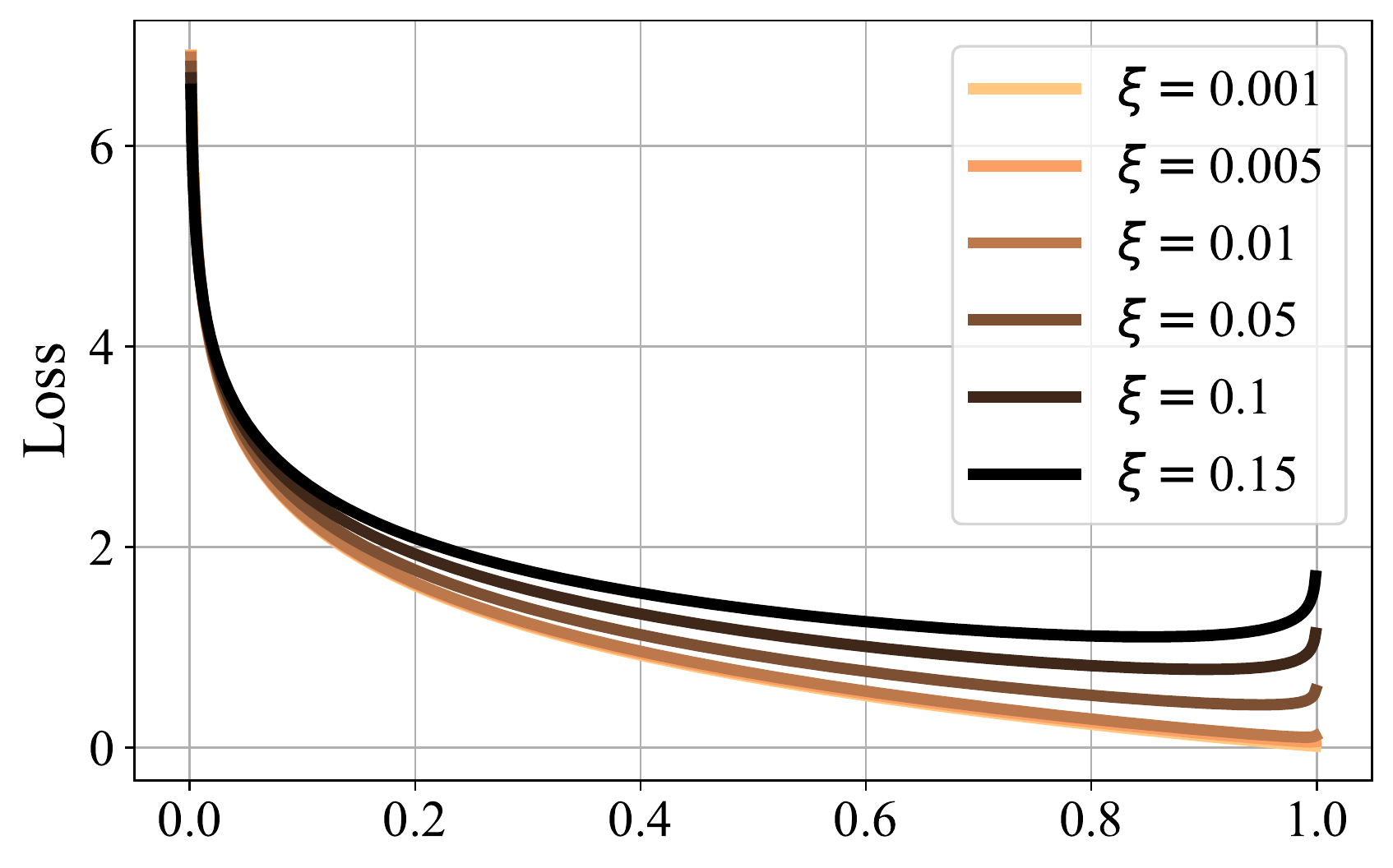}
\end{subfigure}%
\hspace{5mm}
\begin{subfigure}{0.4\textwidth}
\includegraphics[width=1\textwidth]{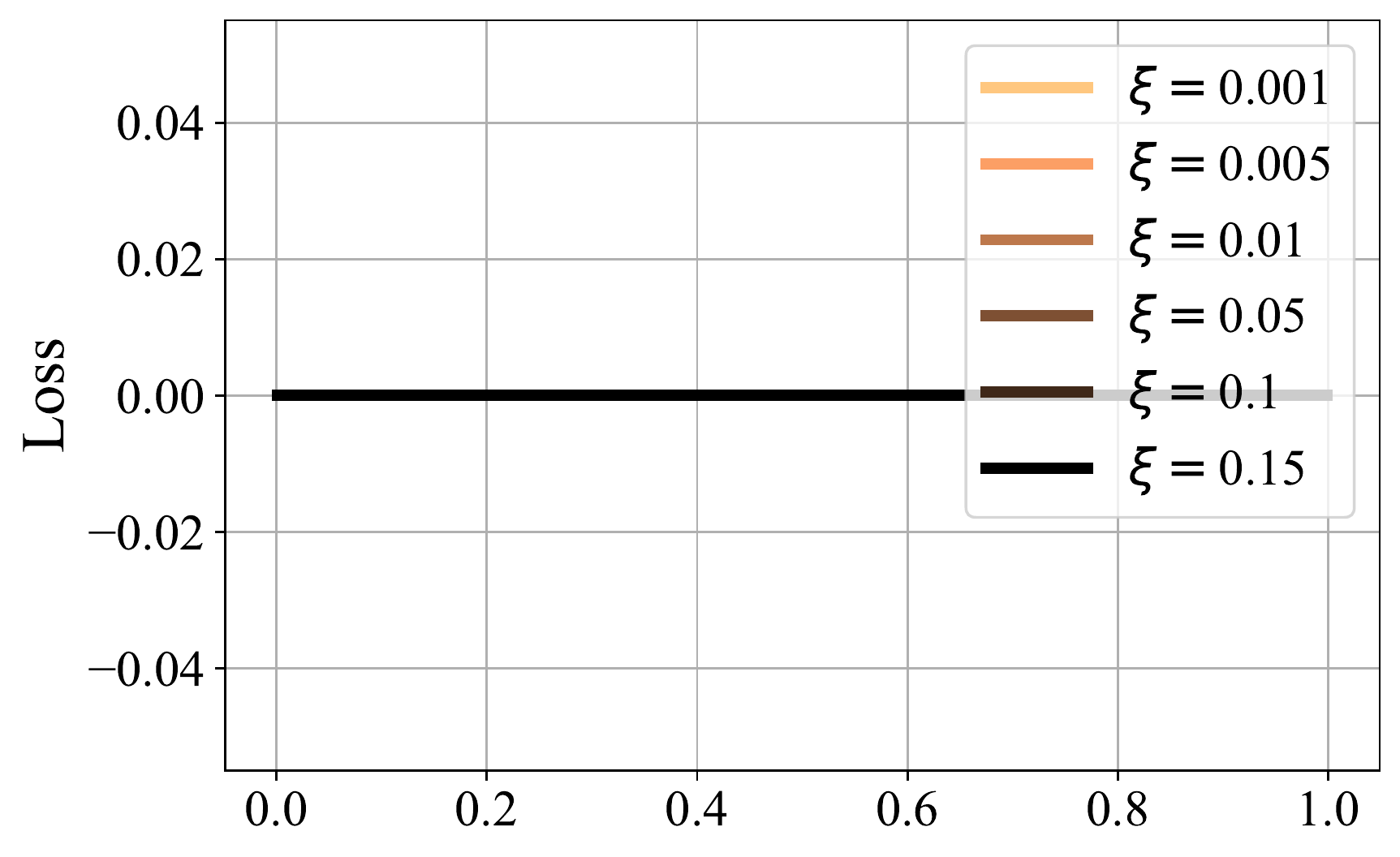}
\end{subfigure}%

\captionsetup{justification=centering}
\caption{Visualizing the proposed \textit{Sparse Label Smoothing Regularization} (SparseLSR) Loss with varying smoothing values $\xi$, where the left figure shows the target loss (\textit{i.e.}, $y_i = 1$), and the right figure shows the non-target loss (\textit{i.e.}, $y_i = 0$).}
\label{fig:fast-label-smoothing-loss}
\end{figure*}

\subsection{Numerical Stability}

The proposed sparse label smoothing regularization loss is prone to numerical stability issues, analogous to the cross-entropy loss, when computing logarithms and exponentials (exponentials are taken in the softmax when converting logits into probabilities) causing under and overflow. In particular, the following expressions are prone to causing numerical stability issues:
\begin{equation}
\begin{split}
\Loss_{SparseLSR} &\approx - \sum_{i=1}^{\mathcal{C}} y_{i}\bigg[\left(1 - \xi + \frac{\xi}{\mathcal{C}}\right) \cdot \overbrace{\log\bigg(f_{\theta}(x)_{i}\bigg)}^{\substack{\text{numerically} \\ \text{unstable}}} \\ & + \frac{\xi(\mathcal{C}-1)}{\mathcal{C}} \cdot \underbrace{\log\left(\frac{1 - f_{\theta}(x)_i}{\mathcal{C}-1}\right)}_{\substack{\text{numerically} \\ \text{unstable}}}\bigg] 
\end{split}
\end{equation}
In order to attain numerical stability when computing $log(f_{\theta}(x)_i)$ the well known \textit{log-sum-exp trick} is employed to stably convert the pre-activation logit $z_i$ into a log probability which we further denote as $\widetilde{f_{\theta}}(x)_i$:
\begin{equation}
\begin{split}
    \widetilde{f_{\theta}}(x)_i
    &= \log(f_{\theta}(x)_i) \\ 
    &= \log\left(\frac{e^{z_{i}}}{\sum_{j=1}^{\mathcal{C}}e^{z_{j}}}\right) \\
    &= \log\left(e^{z_{i}}\right) - \log\left(\textstyle\sum_{j=1}^{\mathcal{C}}e^{z_{j}}\right) \\
    &= z_{i} - \underbrace{\left(\max(z) + \log\left(\textstyle\sum_{j=1}^{\mathcal{C}}e^{z_{j} - \max(z)}\right)\right)}_{\text{LogSumExp}} \\
\end{split}
\end{equation}
%
%
Regarding the remaining numerically unstable term, this can also be computed stably via the log-sum-exp trick; however, it would require performing the log-sum-exp operation an additional time, which would negate the time and space complexity savings over the non-sparse implementation of label smoothing regularization. Therefore, we propose to instead simply take the exponential of the target log probability to recover the raw probability and then add a small constant $\epsilon=1e-7$ to avoid the undefined $\log(0)$ case. The numerically stable sparse label smoothing loss is defined as:
\begin{equation}
\begin{split}
    \Loss_{SparseLSR} 
    &= - \sum_{i=1}^{\mathcal{C}} y_{i}\left[\left(1 - \xi + \frac{\xi}{\mathcal{C}}\right) \cdot \widetilde{f_{\theta}}(x)_{i} \right. \\ & \left. + \frac{\xi(\mathcal{C}-1)}{\mathcal{C}} \cdot \log\left(\frac{1 - e^{\widetilde{f_{\theta}}(x)_i} + \epsilon}{\mathcal{C}-1}\right)\right]
\end{split}
\end{equation}

\vspace{10mm}

\subsection{Time and Space Complexity Analysis}
\label{sec:time-and-space-complexity}

The proposed sparse label smoothing loss function is a sparse method for computing the prototypically non-sparse label smoothing loss given in Equation \eqref{eq:label-smoothing-regularization}. Sparse label smoothing attributes zero losses for the non-target outputs. This makes it invariant to the number of classes $\mathcal{C}$; therefore, the loss function is only calculated and stored on the target while all non-target outputs/losses can be ignored, which gives a computational and storage complexity of $\Theta(1)$ compared to the $\Theta(\mathcal{C})$ for non-sparse label smoothing. To empirically validate the computational complexity of $\Loss_{SparseLSR}$ the runtime is tracked across a series of experiments using a batch of 100 randomly generated targets and logits where the number of classes is gradually increased $\mathcal{C} \in \mathbb{Z}: \mathcal{C} \in [3, 10000]$. The results are presented in Figure \ref{fig:fastlsr-runtime}, where the left figure shows the runtime with the log-softmax operation and the right figure shows without. 

The results confirm that the sparse label smoothing loss function is significantly faster to compute compared to the non-sparse variant. When omitting the softmax-related computation, it is straightforward to see that the non-sparse label smoothing loss function scales linearly in the number of classes $\mathcal{C}$ while the sparse version does not. Although the time savings are admittedly modest in isolation, the loss function is frequently computed in the hundreds of thousands or even millions of times when training deep neural networks; therefore, the time savings when deployed at scale are notable. For example, let $f_{\theta}$ be trained for $1,000,000$ gradient steps for a $\mathcal{C}=10,000$ class classification problem with a batch size $\mathcal{B} = 100$. From Figure \ref{fig:fastlsr-runtime} spare label smoothing saves approximately $0.3$ seconds per batch compared to the non-sparse variant, so when run over a training session of $1,000,000$ gradient steps the time savings correspond to $(0.3 \times 1,000,000)/(60^2) \approx$ $83.3$ hours, which is a significant amount of time.

\subsection{Visualizing Learned Representations}
\label{sec:sparse-label-smoothing-regularization-experiments}

\begin{figure*}[t!]
\centering

\begin{subfigure}{0.4\textwidth}
\includegraphics[width=1\textwidth]{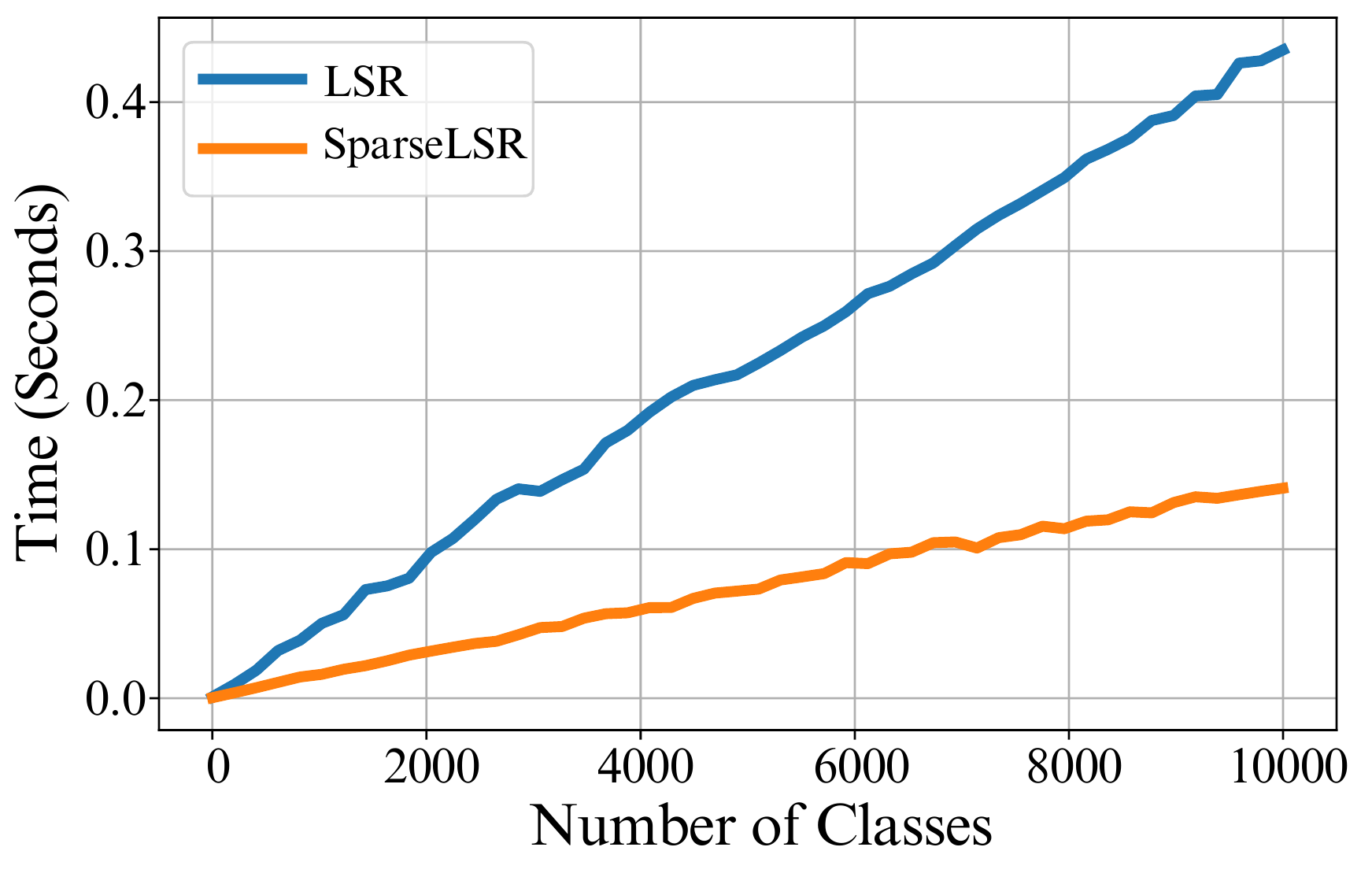}
\end{subfigure}%
\hspace{5mm}
\begin{subfigure}{0.4\textwidth}
\includegraphics[width=1\textwidth]{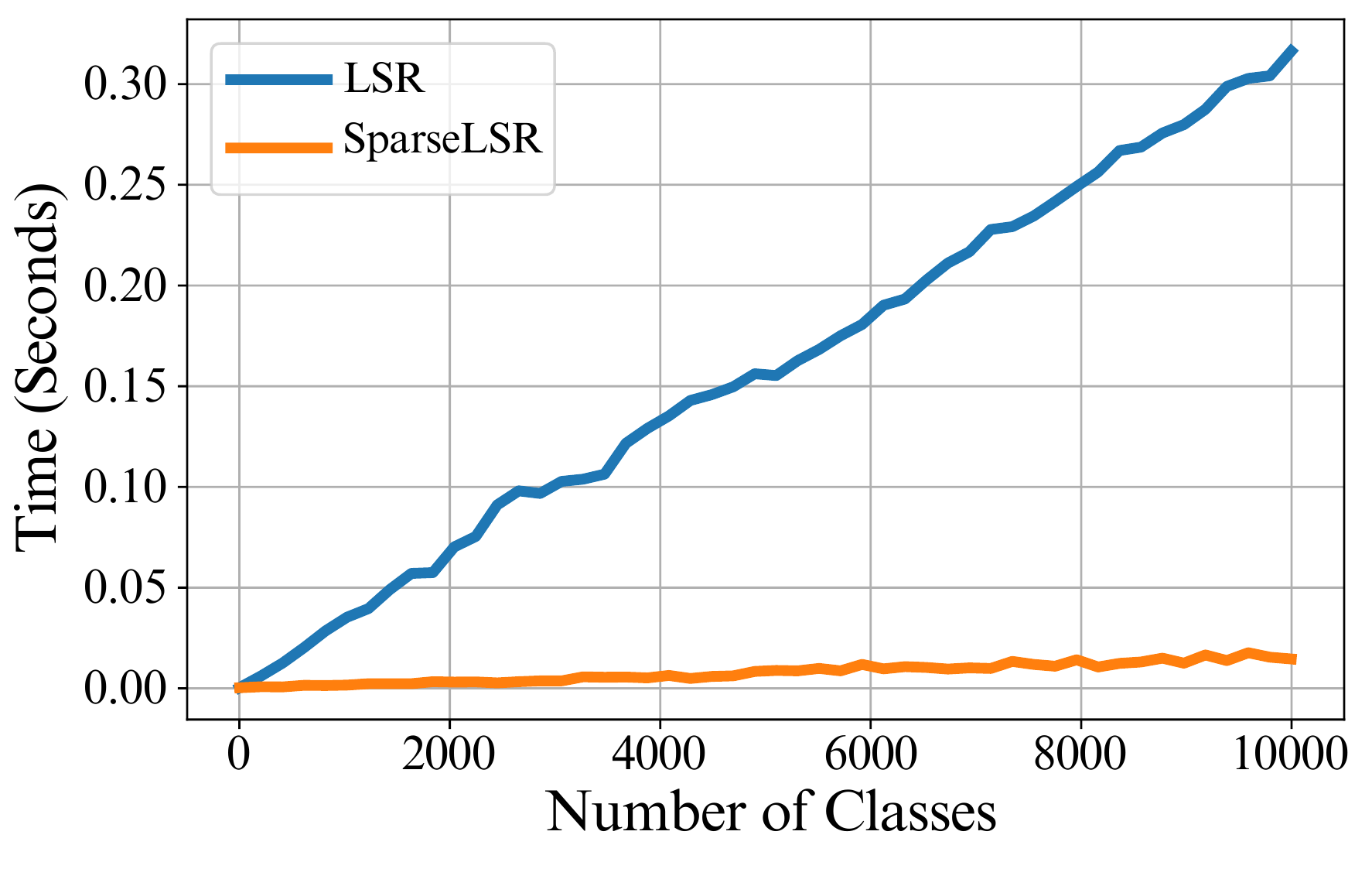}
\end{subfigure}%
\captionsetup{justification=centering}
\caption{Comparing the runtime of Label Smoothing Regularization (LSR) Loss and Sparse Label Smoothing Regularization (SparseLSR) Loss. The left figure shows the runtime with the log softmax operation and the right figure shows without.}
\label{fig:fastlsr-runtime}
\end{figure*}

To further validate the parity of the regularization behavior of our sparse label smoothing loss function compared to non-sparse label smoothing we again visualize and analyze the penultimate layer representations of AlexNet on CIFAR-10 using t-SNE \cite{van2008visualizing}. The results are presented in Figure \ref{fig:fast-label-smoothing-regularization-1}, and they show that the learned representations between the two loss functions are very similar, with often near identical inter-class separations and inter-class overlaps between the same classes. These results empirically confirm that our sparse label smoothing loss function which redistributes the expected non-target loss into the target has faithfully retrained the original behavior of label smoothing regularization. 

A surprising result from our experiments was that for larger values of $\xi$, there is slightly improved inter-class separation and more compact intra-class representation when using the sparse version of label smoothing regularization. Consequently, our sparse label smoothing loss function has slightly improved final inference performance relative to non-sparse label smoothing, often showing a consistent $0.5-1\%$ improvement in error rate, which was not intentional or expected. We hypothesize that this is an effect caused by the model uniformly penalizing all the non-target outputs for overconfident target predictions during backpropagation, enabling the model to learn more robust decision boundaries and internal representations. In non-sparse label smoothing, non-target outputs are not penalized uniformly; instead, they are penalized on a case-by-case basis. 

\subsection{Training Dynamics}

Finally, one of the key motivations for designing sparse label smoothing regularization was to address the non-smooth behavior of the absolute cross-entropy loss when transitioning between minimizing and maximizing the target output, which causes poor training dynamics. To demonstrate the improved training dynamics of sparse label smoothing regularization we visualize the training learning curves in Figure \ref{fig:proposed-loss-learning-curves}, when using the different loss functions. The results show that the absolute cross-entropy loss often does not converge early in the training (likely due to it being non-smooth). In contrast, both sparse and non-sparse label smoothing regularization demonstrate smooth and rapid convergence, with their learning curves closely overlapping. These findings suggest that the linear approximation utilized in Equation \eqref{eq:linear-approximation-of-log} for sparse label smoothing does not adversely affect the learning behavior.

\begin{figure}[t!]
\centering
\includegraphics[width=0.8\columnwidth]{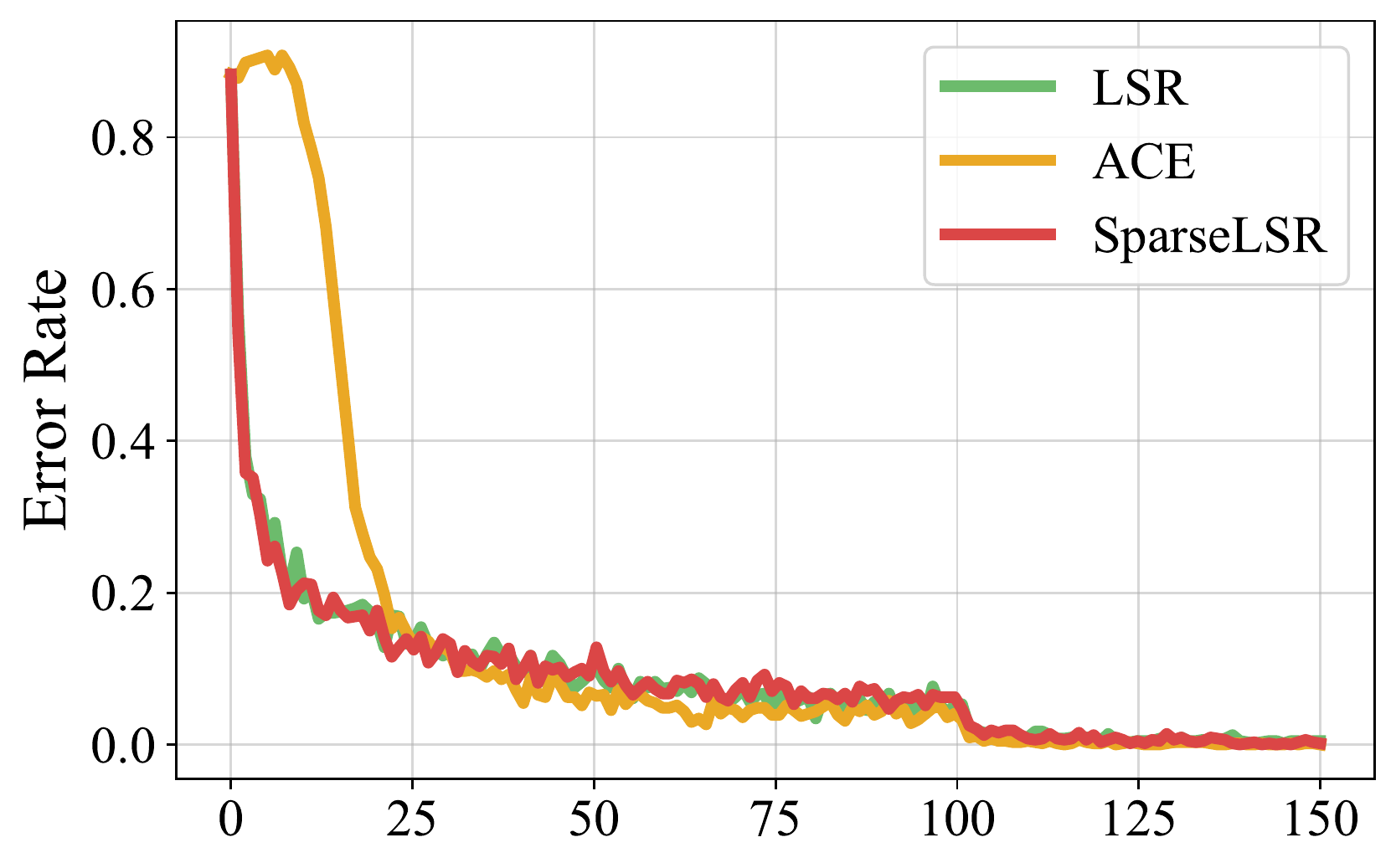}
\caption{The mean training curves on AlexNet CIFAR-10 when using Label Smoothing Regularization, Absolute Cross-Entropy Loss, and Sparse Label Smoothing Regularization. The x-axis shows the number of gradient steps in thousands.}
\label{fig:proposed-loss-learning-curves}
\end{figure}

\subsection{Extensions}

\begin{figure*}[t!]
\centering
\includegraphics[width=0.85\textwidth]{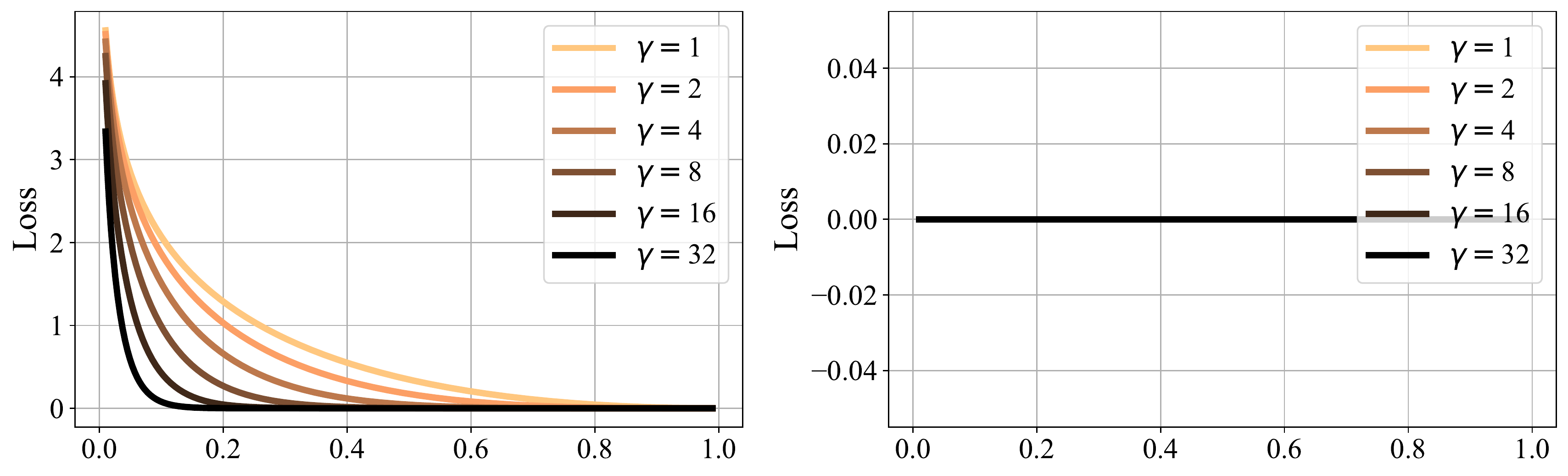}
\captionsetup{justification=centering}
\caption{Visualizing the \textit{Focal Loss} with varying focusing values $\gamma$, where the left figure shows the \\ target loss (\textit{i.e.}, $y_i = 1$), and the right figure shows the non-target loss (\textit{i.e.}, $y_i = 0$).}
\label{fig:focal_loss}
\end{figure*}

\begin{figure*}[t!]
\centering
\includegraphics[width=0.85\textwidth]{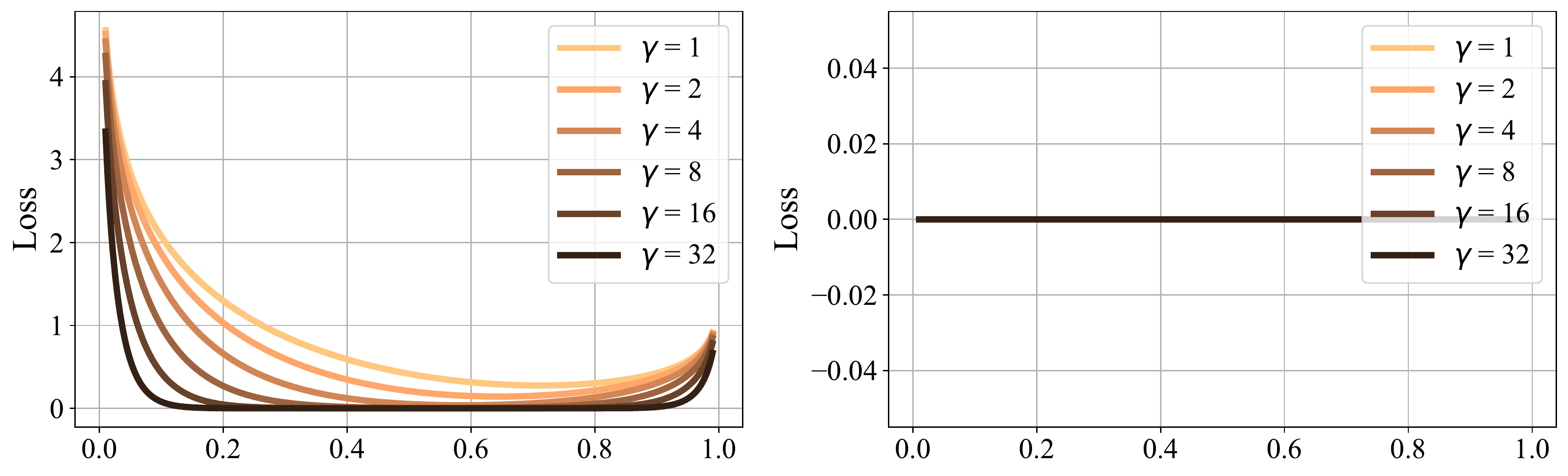}
\captionsetup{justification=centering}
\caption{Visualizing the proposed \textit{Focal Loss with Sparse Label Smoothing Regularization} (Focal+SparseLSR) \\ with varying focusing values $\gamma$ and a smoothing coefficient value $\xi=0.2$, where the left figure \\ shows the target loss (\textit{i.e.}, $y_i = 1$), and the right figure shows the non-target loss (\textit{i.e.}, $y_i = 0$).}
\label{fig:focal_sparse_label_smoothing_loss}
\vspace{2mm}
\end{figure*}

The redistributed loss trick described in Section \ref{sec:sparse-label-smoothing-regularization} is general and can be applied straightforwardly to any softmax-based classification loss function. To illustrate this, we demonstrate how the Focal Loss, a popular classification loss function, can be extended to integrate sparse label smoothing regularization. The Focal Loss, introduced by \cite{lin2017focal}, extends the standard cross-entropy loss by incorporating a term $(1 - f_{\theta}(x)_i)^{\gamma}$, where $\gamma \geq 0$ is the ``focusing'' hyperparameter. The focusing term reduces the loss for well-classified examples, thereby emphasizing harder misclassified examples during training, as shown in Figure \ref{fig:focal_loss}. 
\begin{equation}
    \mathcal{L}_{Focal} = - \sum_{i=1}^{\mathcal{C}} y_{i} \bigg[(1 - f_{\theta}(x)_{i} )^{\gamma} \cdot \log\left(f_{\theta}(x)_{i}\right)\bigg]
\end{equation}
The focal loss can be combined with sparse label smoothing regularization by incorporating the terms $(1 - f(x)_i)^{\gamma}$ and its shifted reflection $(f(x)_i)^{\gamma}$ into Equation \eqref{eq:sparse-label-smoothing-regularization} as follows:
\begin{equation}
\begin{split}
    \mathcal{L}_{Focal+SparseLSR} &= - \sum_{i=1}^{\mathcal{C}} y_{i}\bigg[\bigg(1 - f_{\theta}(x)_i \bigg)^{\gamma} \cdot \left(1 - \xi + \frac{\xi}{\mathcal{C}}\right) \\ & \cdot \log(f_{\theta}(x)_{i}) + \bigg(f_{\theta}(x)_i \bigg)^{\gamma} \cdot \frac{\xi(\mathcal{C}-1)}{\mathcal{C}} \\ & \cdot \log\left(\frac{1 - f_{\theta}(x)_i}{\mathcal{C}-1}\right)\bigg]
\end{split}
\end{equation}
As illustrated in Figure \ref{fig:focal_sparse_label_smoothing_loss}, the \textit{Focal Loss with Sparse Label Smoothing Regularization} (Focal+SparseLSR) can capture two key behaviors commonly seen in meta-learned loss functions. First, it can increase and decrease the relative loss assigned to well-classified versus poorly-classified examples. Second, it can penalize overly confident predictions through sparse label smoothing regularization. Importantly, $\Loss_{Focal+SparseLSR}$ can achieve both of these behaviors while remaining sparse, thereby benefiting from the computational and memory advantages outlined in Section \ref{sec:time-and-space-complexity}.

\vspace{10mm}
\balance

\subsection{Conclusion}

In conclusion, we have performed an empirical and theoretical analysis of some of the meta-learned loss functions by EvoMAL. Our analysis began with a brief overview of some of the meta-learned loss functions learned by EvoMAL, followed by a critical assessment of some of the previous hypotheses for “why meta-learned loss functions perform better than handcrafted loss functions” and “what are meta-learned loss functions learning”. In a quest to answer these questions, we performed a theoretical analysis of the \textit{Absolute Cross-Entropy} (ACE) Loss, one of the learned loss functions from EvoMAL. The analysis revealed that it has identical behavior at the start and very end of training to Label Smoothing Regularization (LSR), a popular regularization technique for penalizing overconfident predictions, which explains why it was able to improve performance.

Although effective, the absolute cross-entropy loss function is not smooth and has divergent behavior to label smoothing regularization at intermediate points in the training. To resolve these issues a new novel loss function was proposed inspired by our theoretical findings called \textit{Sparse Label Smoothing Regularization} (SparseLSR). The proposed loss function is similar to non-sparse label smoothing regularization; however, it utilizes the redistributed loss trick which takes the expected non-target loss and redistributes it into the target loss, obviating the need to calculate the loss on the non-target outputs. Our experimental results show that the proposed loss function is significantly faster to compute, with the time and space complexity being reduced from linear to constant time with respect to the number of classes being considered. The \texttt{PyTorch} code for the proposed Sparse Label Smoothing Loss can be found at: \href{https://github.com/Decadz/Sparse-Label-Smoothing-Regularization}{https://github.com/Decadz/Sparse-Label-Smoothing-Regularization}

\begin{figure*}
\centering

\begin{subfigure}{0.35\textwidth}
\includegraphics[width=1\textwidth]{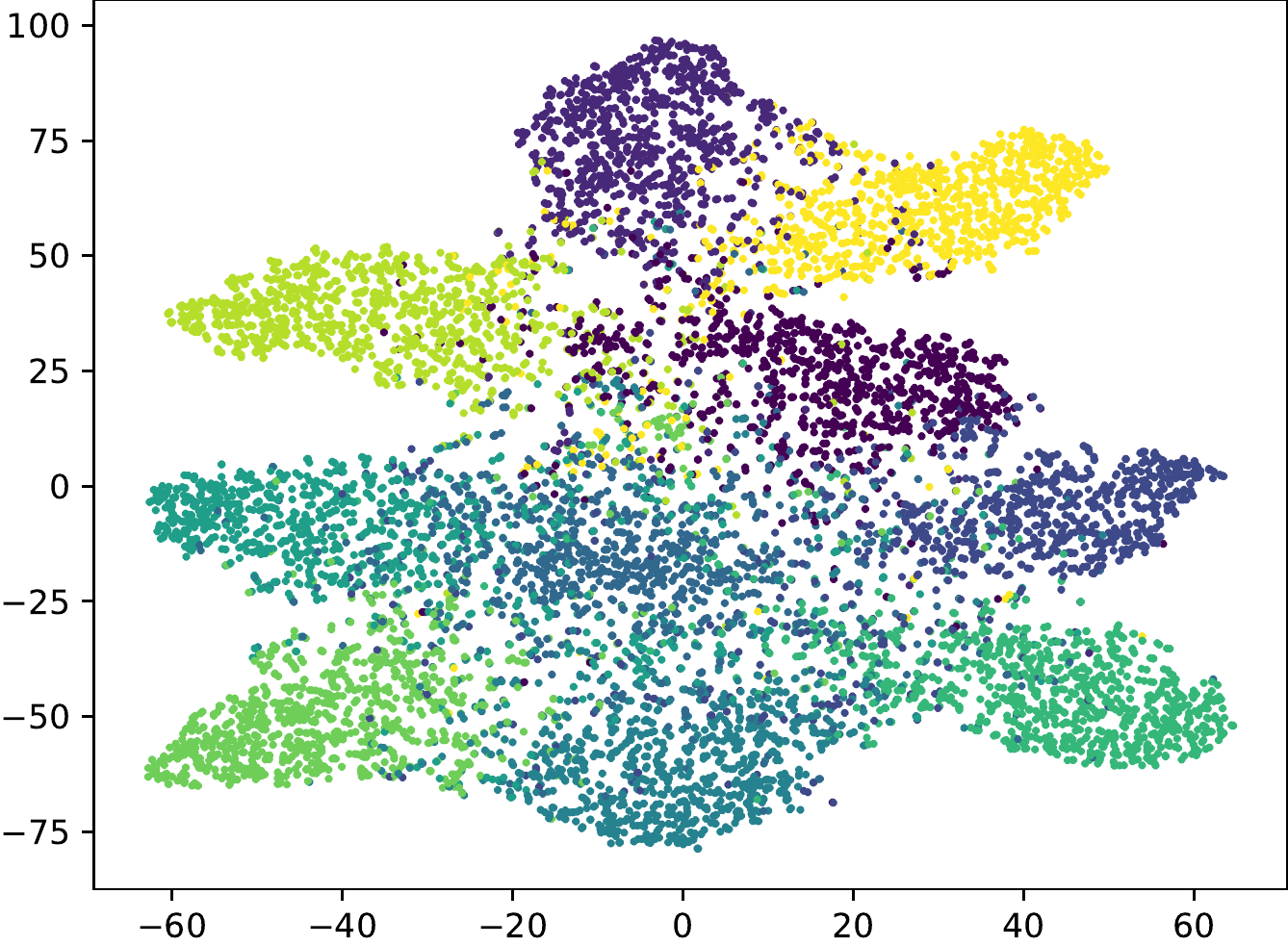}
\caption{LSR ($\xi=0$) -- 15.50\%.}
\end{subfigure}%
\hspace{5mm}
\begin{subfigure}{0.35\textwidth}
\includegraphics[width=1\textwidth]{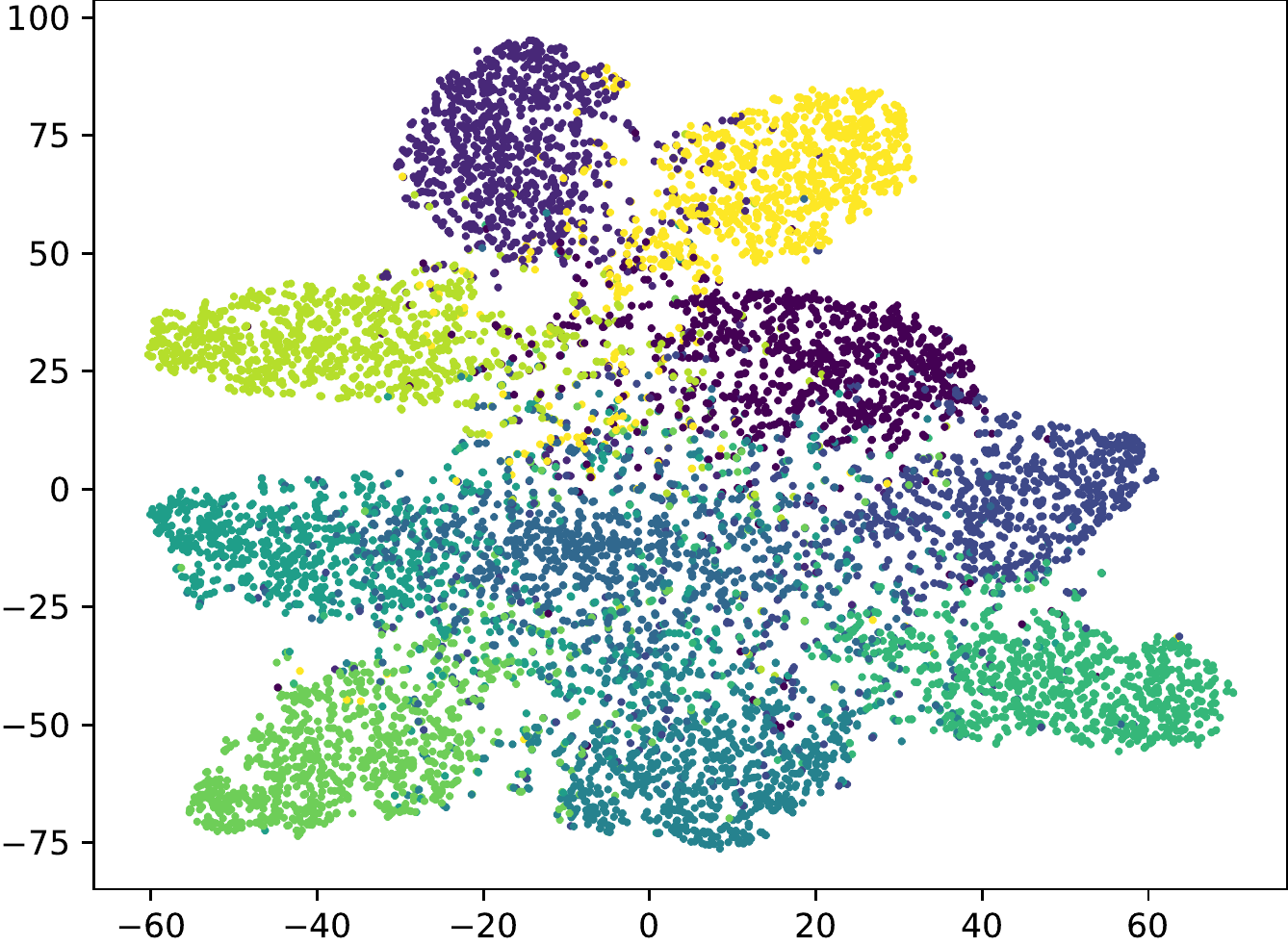}
\caption{SparseLSR ($\xi=0$) -- 15.23\%.}
\end{subfigure}%

\par\bigskip

\begin{subfigure}{0.35\textwidth}
\includegraphics[width=1\textwidth]{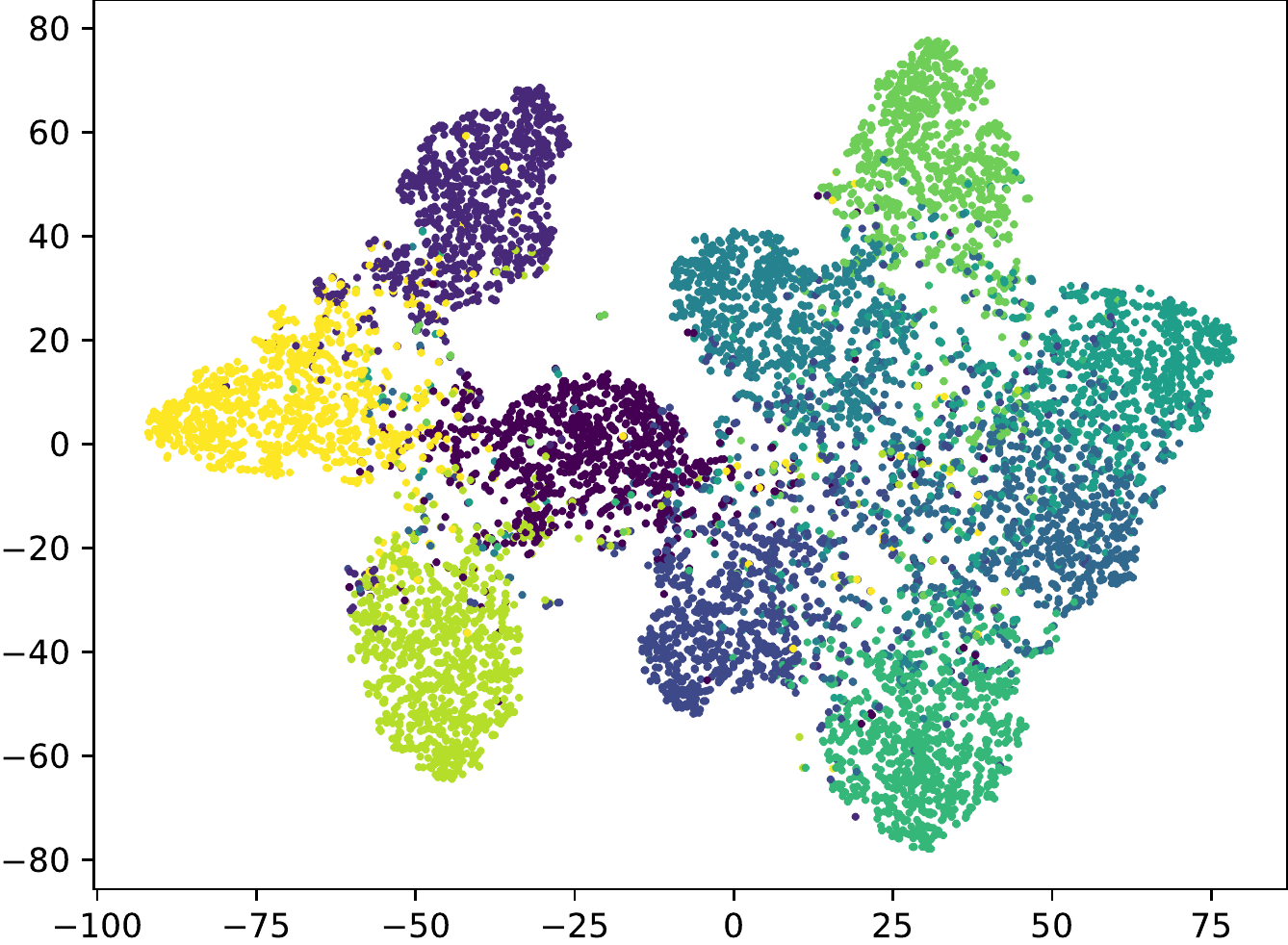}
\caption{LSR ($\xi=0.001$) -- 15.53\%.}
\end{subfigure}%
\hspace{5mm}
\begin{subfigure}{0.35\textwidth}
\includegraphics[width=1\textwidth]{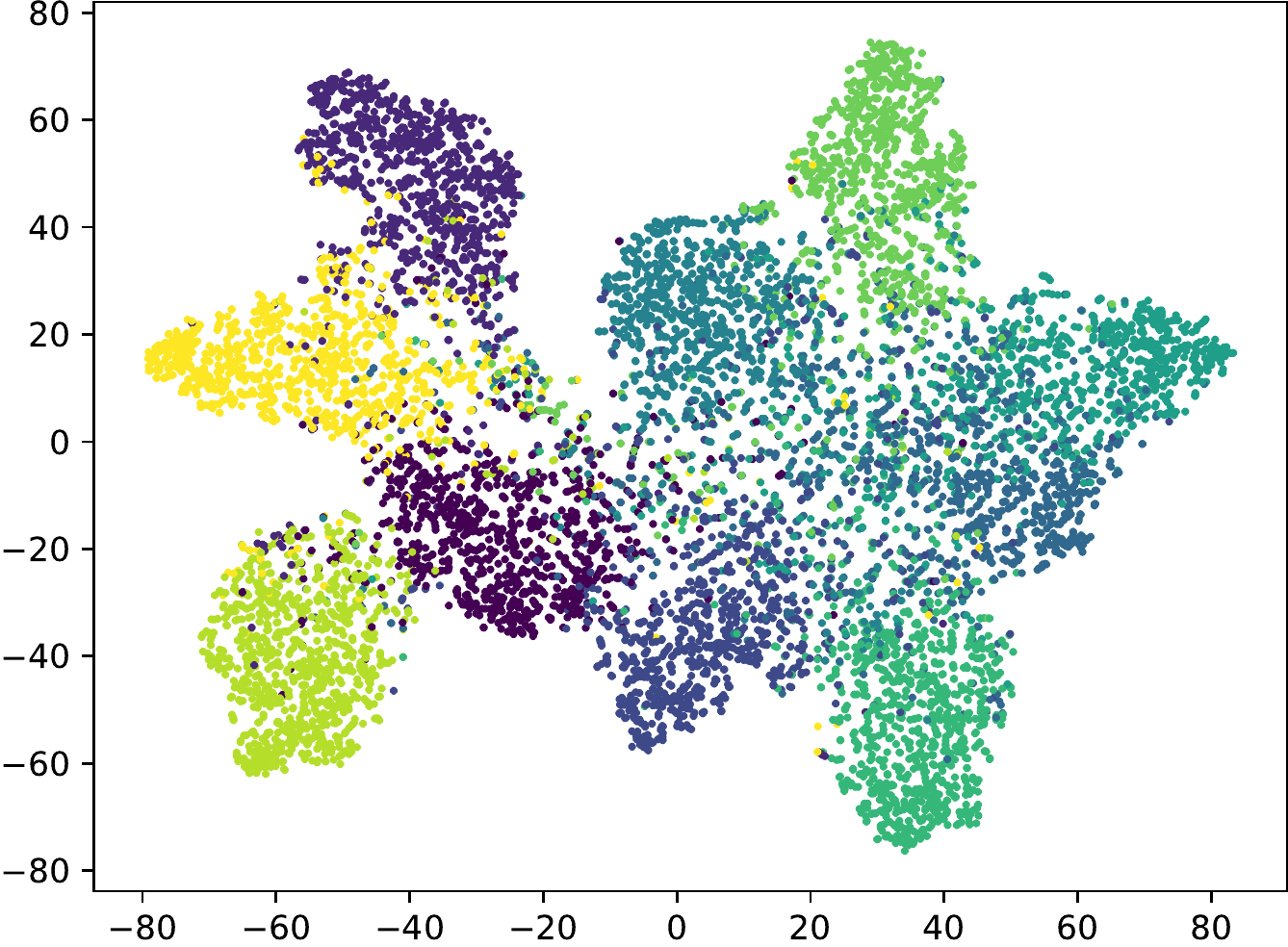}
\caption{SparseLSR ($\xi=0.001$) -- 15.47\%.}
\end{subfigure}%

\par\bigskip

\begin{subfigure}{0.35\textwidth}
\includegraphics[width=1\textwidth]{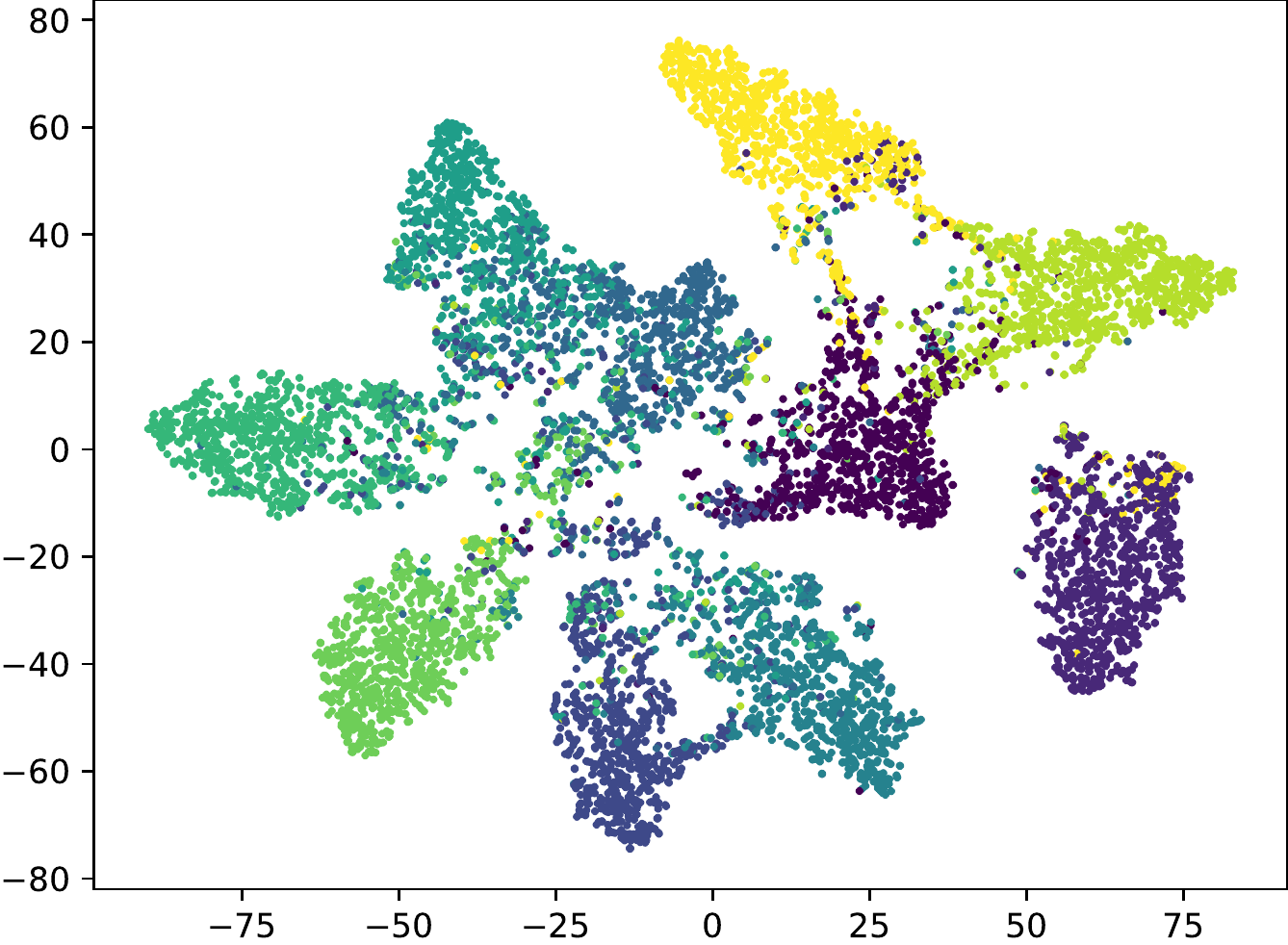}
\caption{LSR ($\xi=0.005$) -- 15.31\%.}
\end{subfigure}%
\hspace{5mm}
\begin{subfigure}{0.35\textwidth}
\includegraphics[width=1\textwidth]{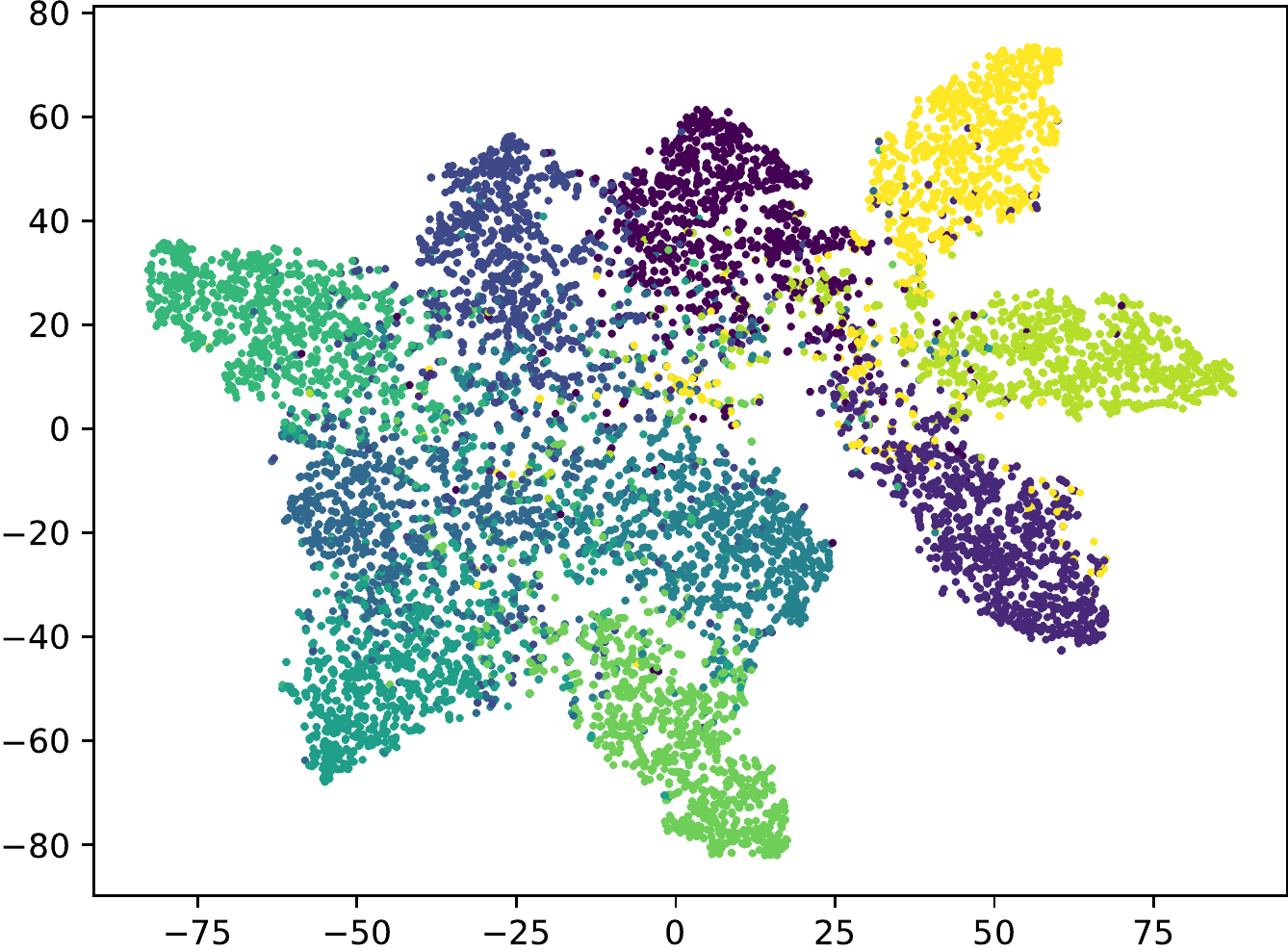}
\caption{SparseLSR ($\xi=0.005$) -- 15.66\%.}
\end{subfigure}%

\par\bigskip

\begin{subfigure}{0.35\textwidth}
\includegraphics[width=1\textwidth]{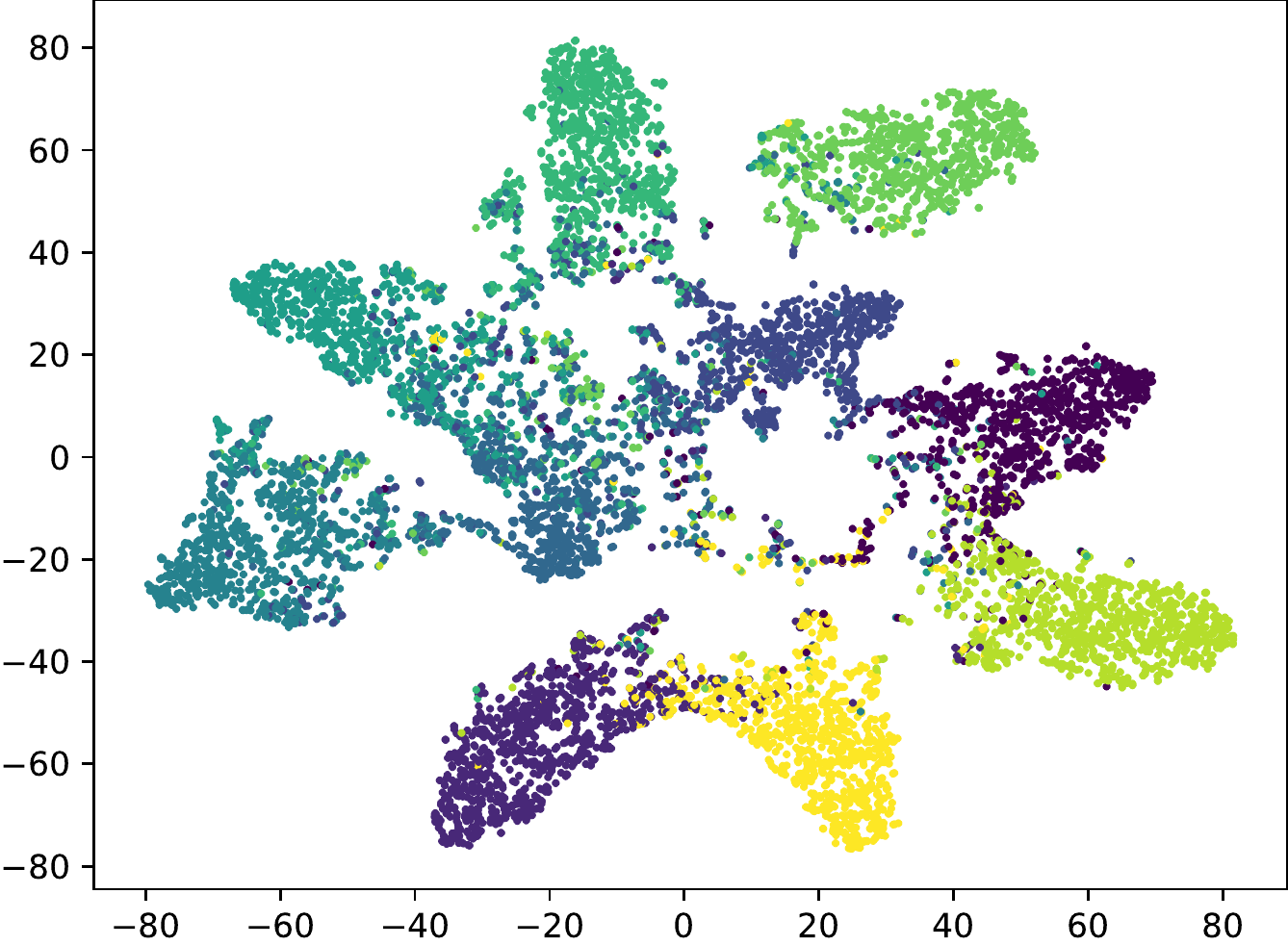}
\caption{LSR ($\xi=0.01$) -- 15.77\%.}
\end{subfigure}%
\hspace{5mm}
\begin{subfigure}{0.35\textwidth}
\includegraphics[width=1\textwidth]{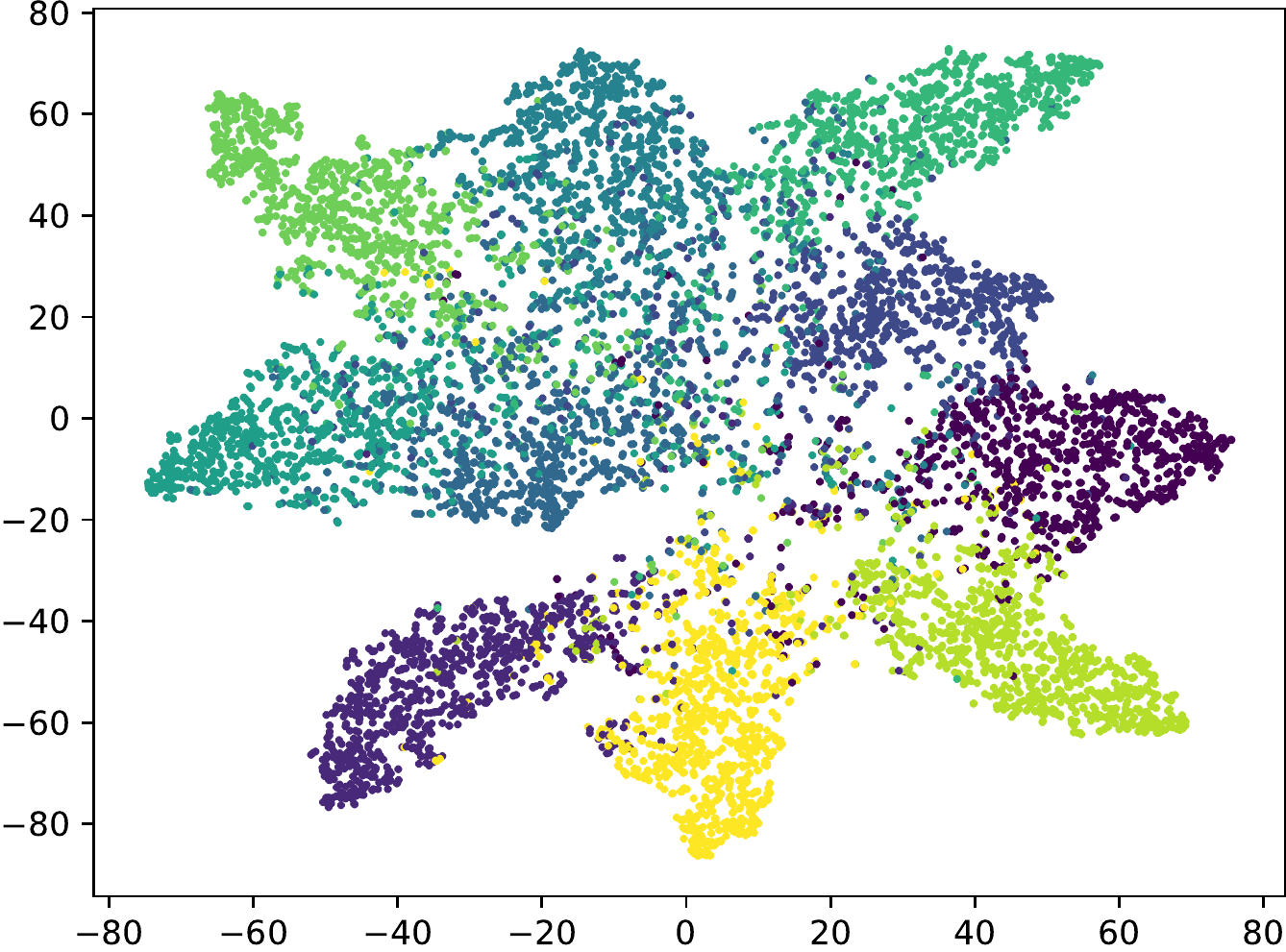}
\caption{SparseLSR ($\xi=0.01$) -- 15.42\%.}
\end{subfigure}%

\captionsetup{justification=centering}
\caption{Visualizing the penultimate layer representations on AlexNet CIFAR-10. Comparing \textit{Label Smoothing Regularization} (LSR) to our proposed \textit{Sparse Label Smoothing Regularization} (SparseLSR).}
\label{fig:fast-label-smoothing-regularization-1}
\end{figure*}

\begin{figure*}
\centering

\ContinuedFloat

\begin{subfigure}{0.35\textwidth}
\includegraphics[width=1\textwidth]{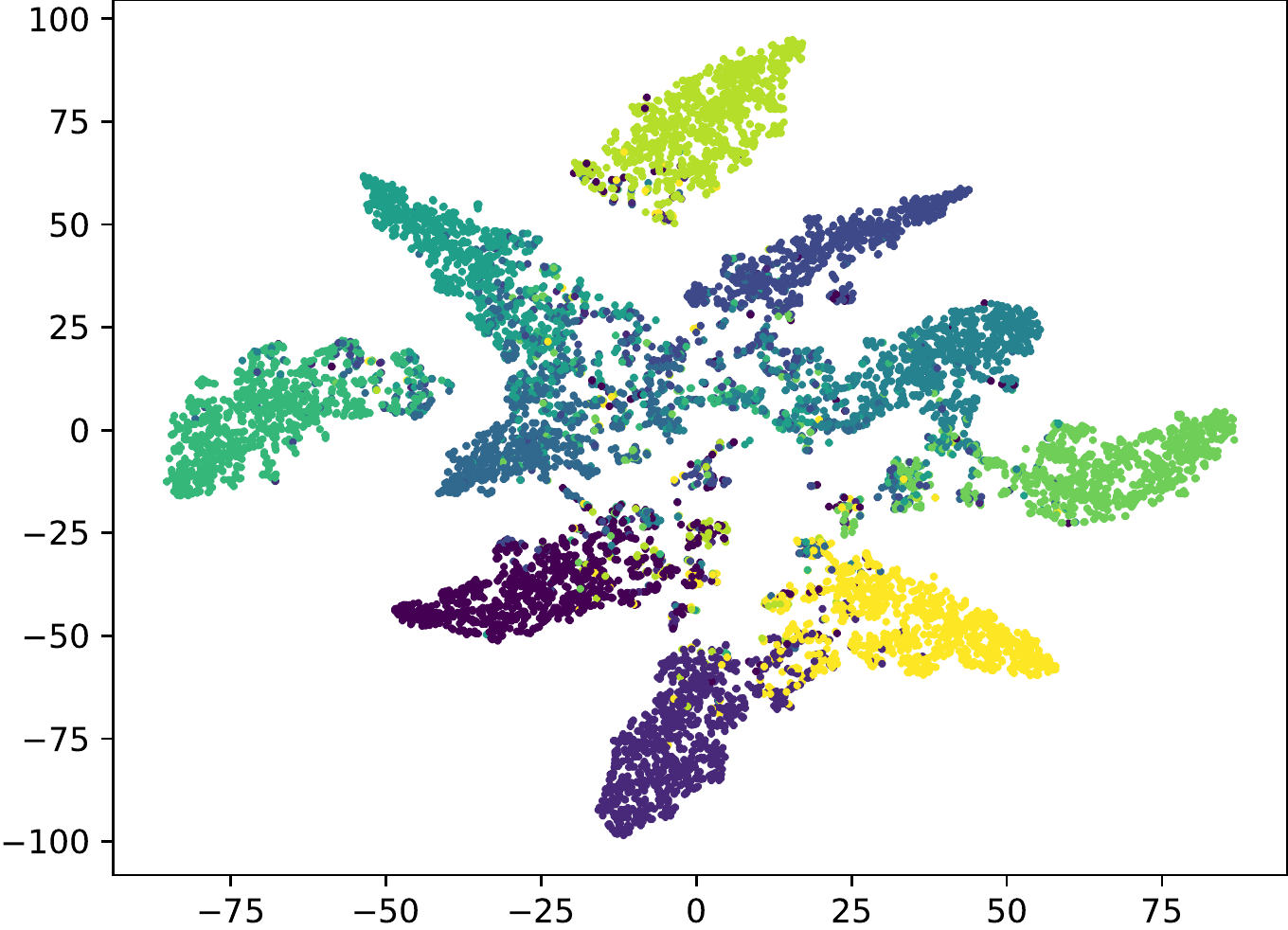}
\caption{LSR ($\xi=0.05$) -- 15.33\%.}
\end{subfigure}%
\hspace{5mm}
\begin{subfigure}{0.35\textwidth}
\includegraphics[width=1\textwidth]{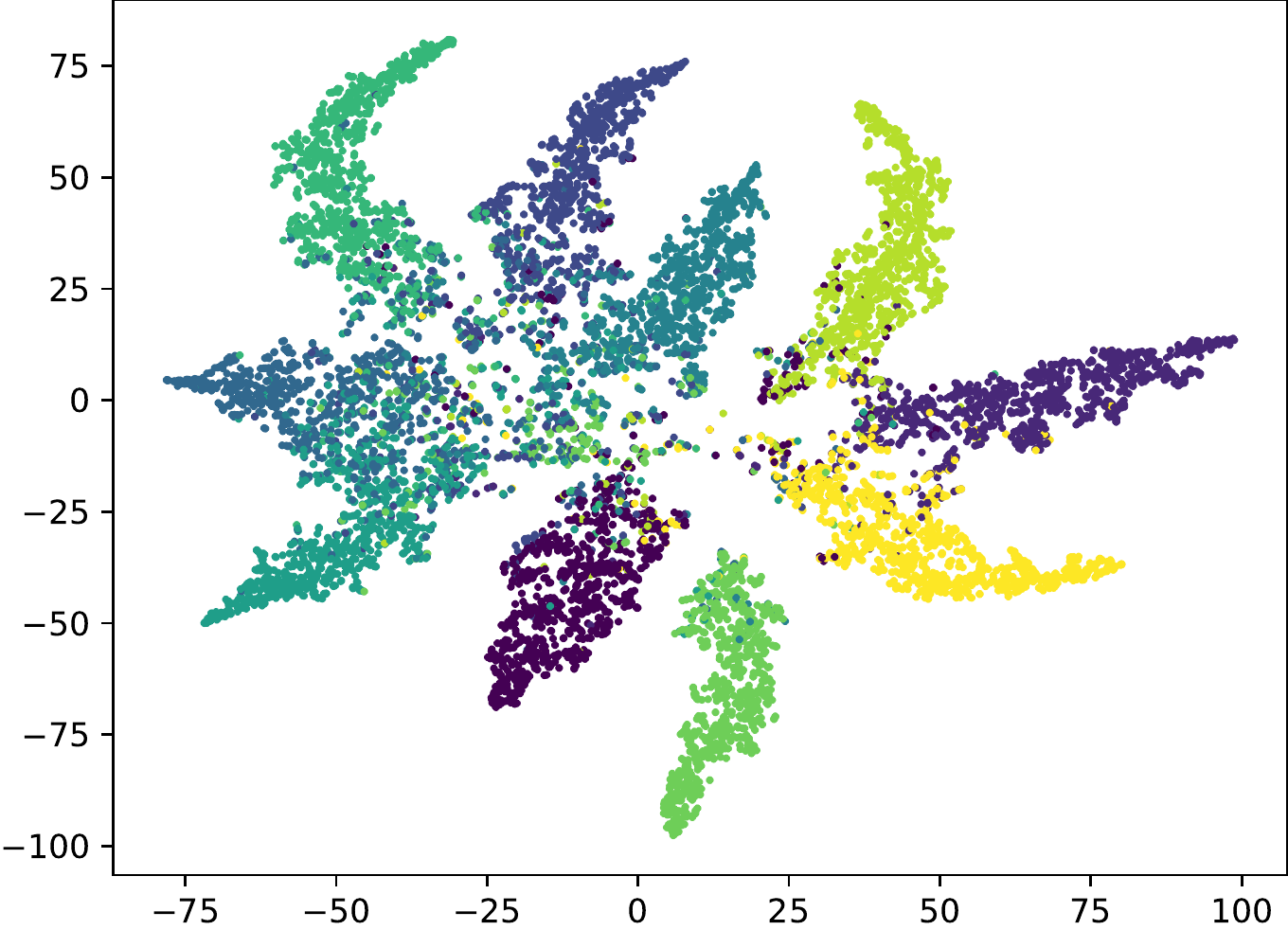}
\caption{SparseLSR ($\xi=0.05$) -- 14.74\%.}
\end{subfigure}%

\par\bigskip

\begin{subfigure}{0.35\textwidth}
\includegraphics[width=1\textwidth]{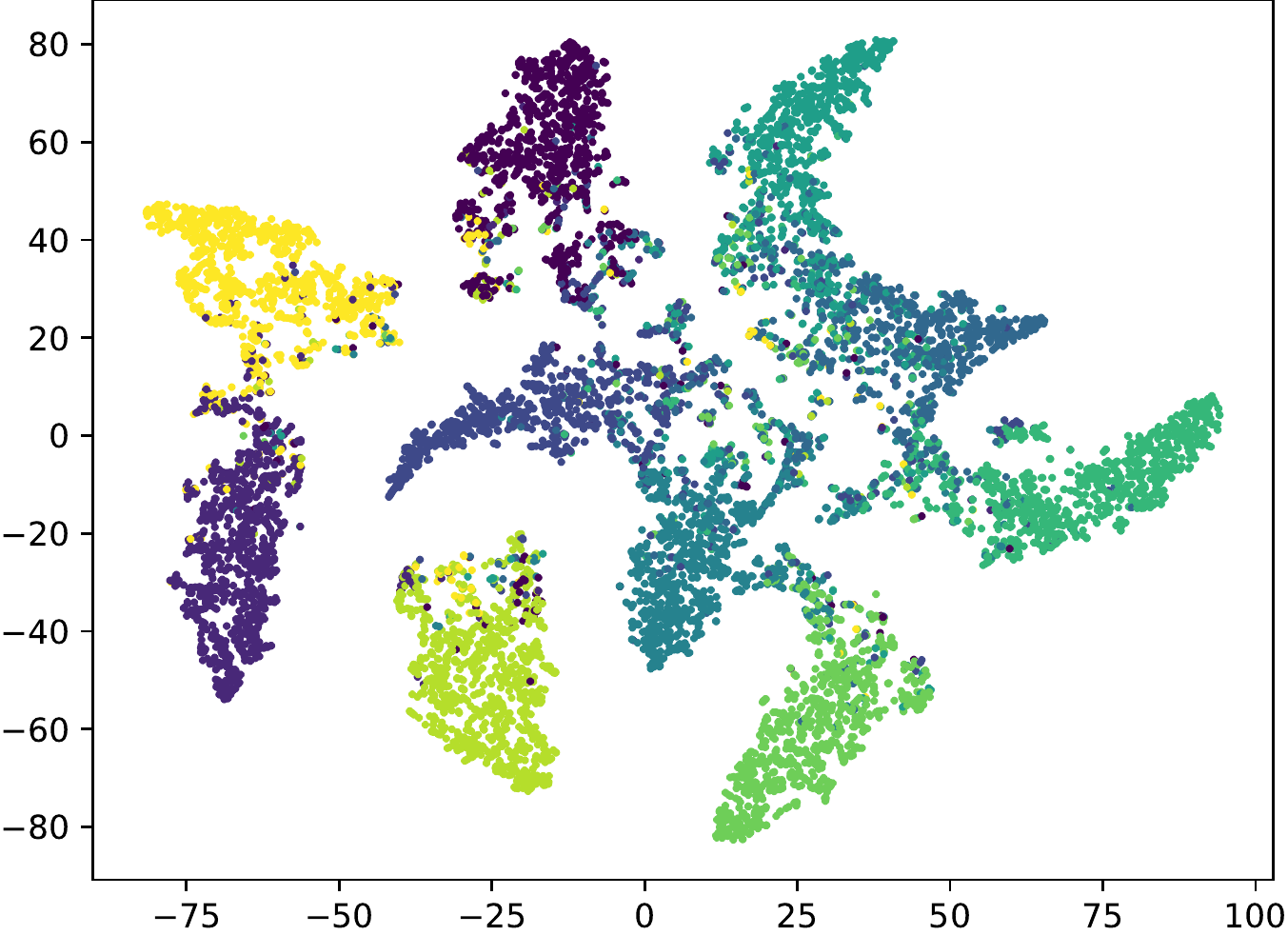}
\caption{LSR ($\xi=0.1$) -- 15.05\%.}
\end{subfigure}%
\hspace{5mm}
\begin{subfigure}{0.35\textwidth}
\includegraphics[width=1\textwidth]{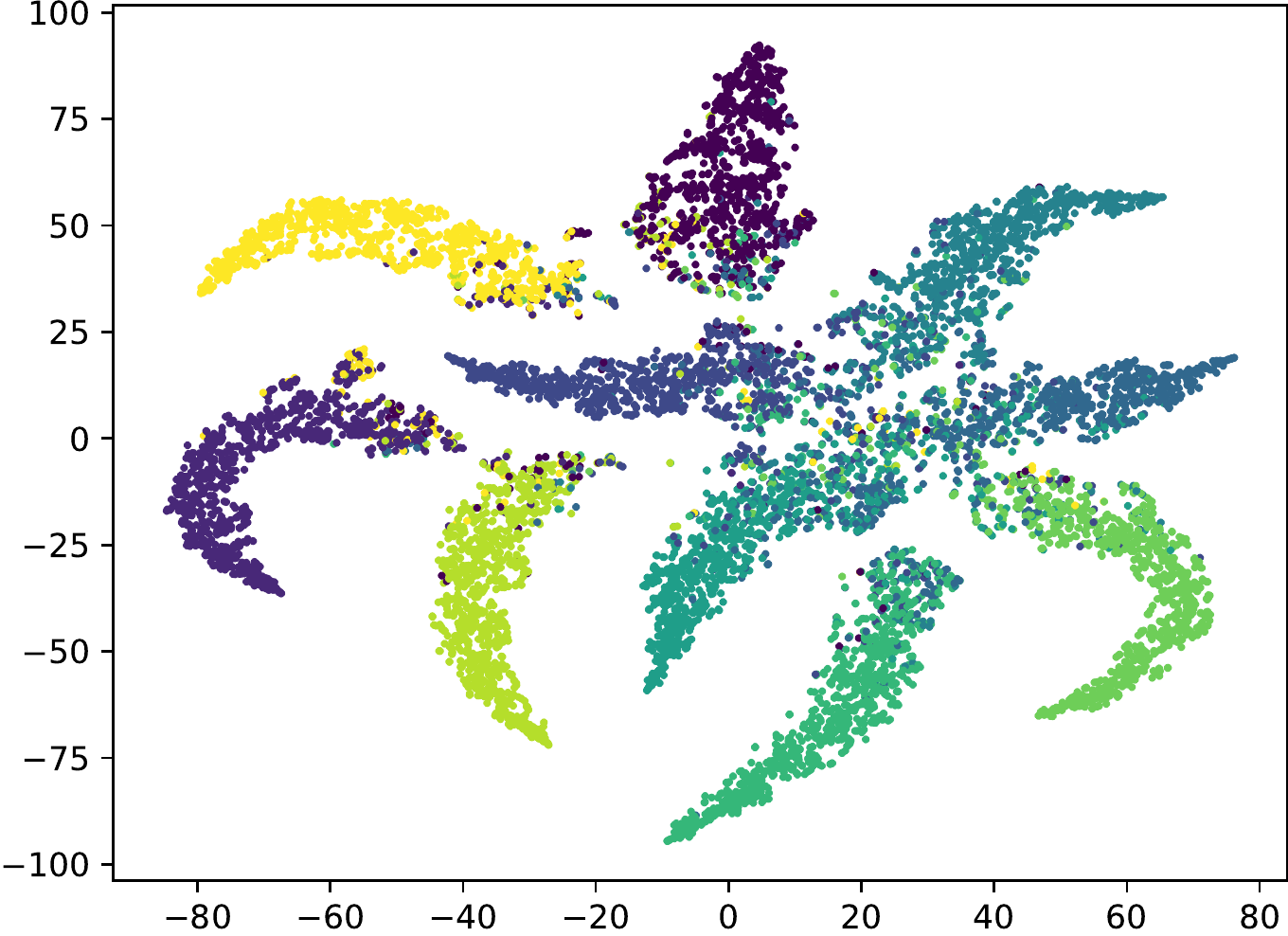}
\caption{SparseLSR ($\xi=0.1$) -- 14.68\%.}
\end{subfigure}%

\par\bigskip

\begin{subfigure}{0.35\textwidth}
\includegraphics[width=1\textwidth]{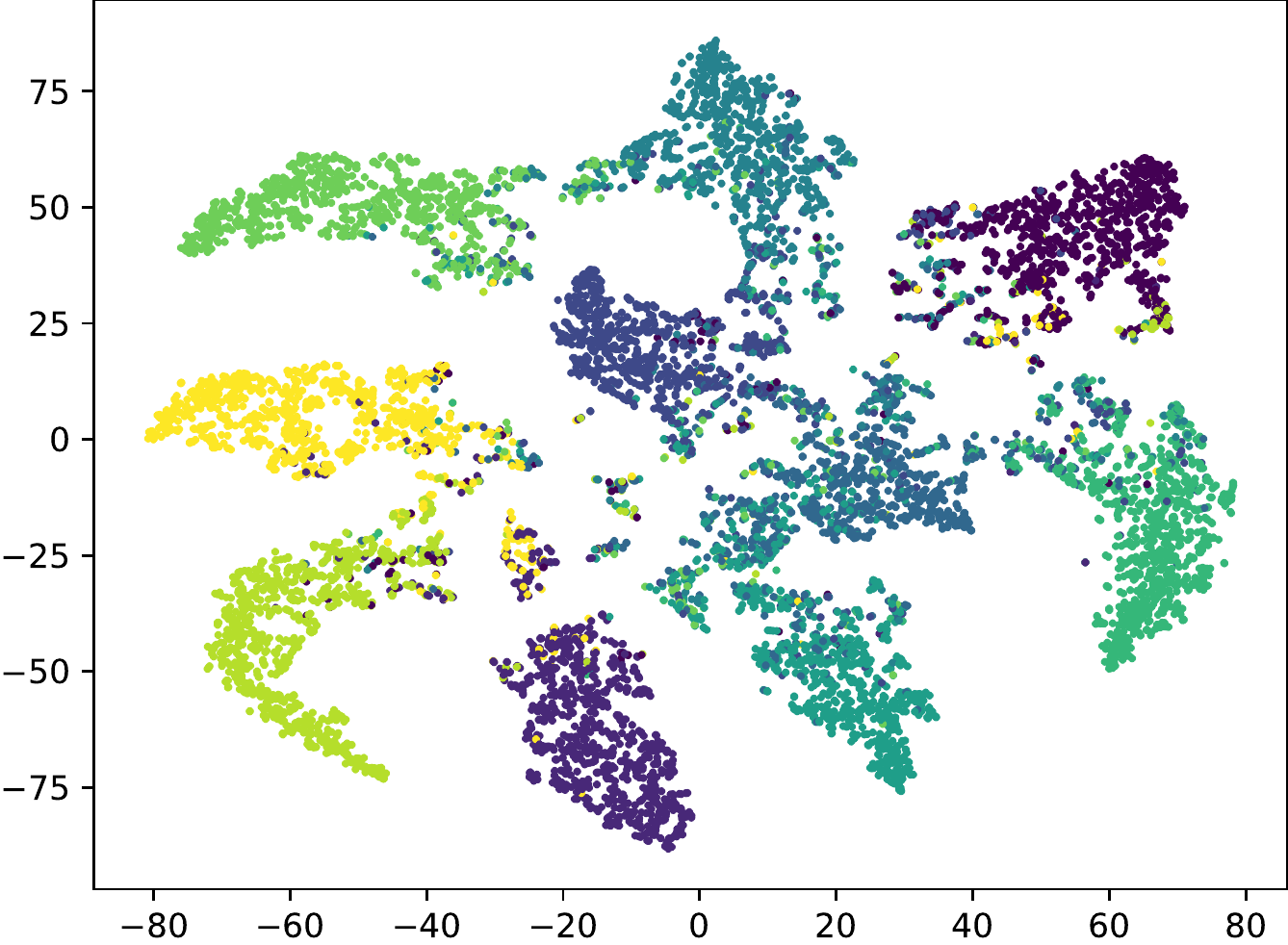}
\caption{LSR ($\xi=0.2$) -- 15.04\%.}
\end{subfigure}%
\hspace{5mm}
\begin{subfigure}{0.35\textwidth}
\includegraphics[width=1\textwidth]{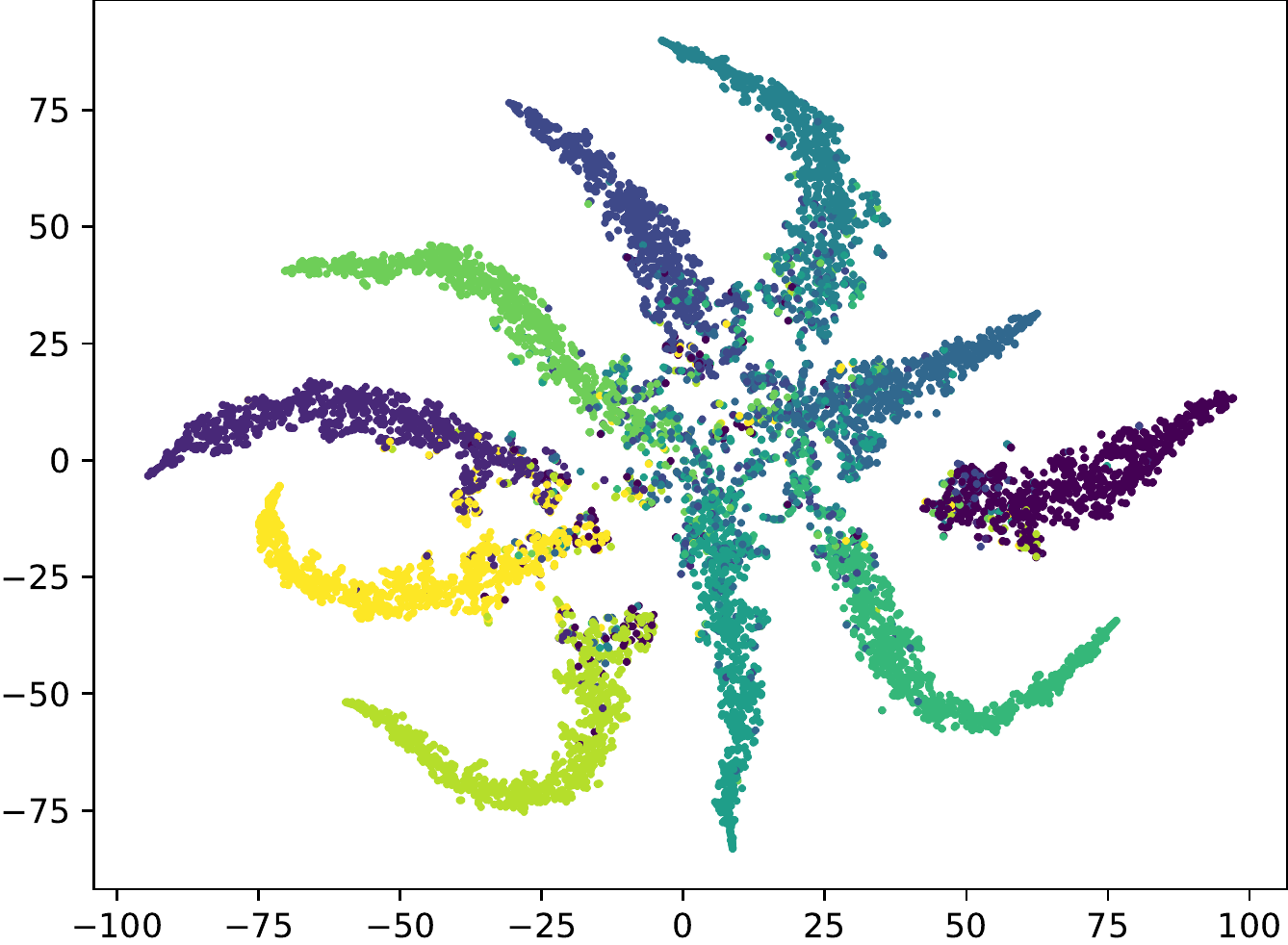}
\caption{SparseLSR ($\xi=0.2$) -- 14.41\%.}
\end{subfigure}%

\par\bigskip

\begin{subfigure}{0.35\textwidth}
\includegraphics[width=1\textwidth]{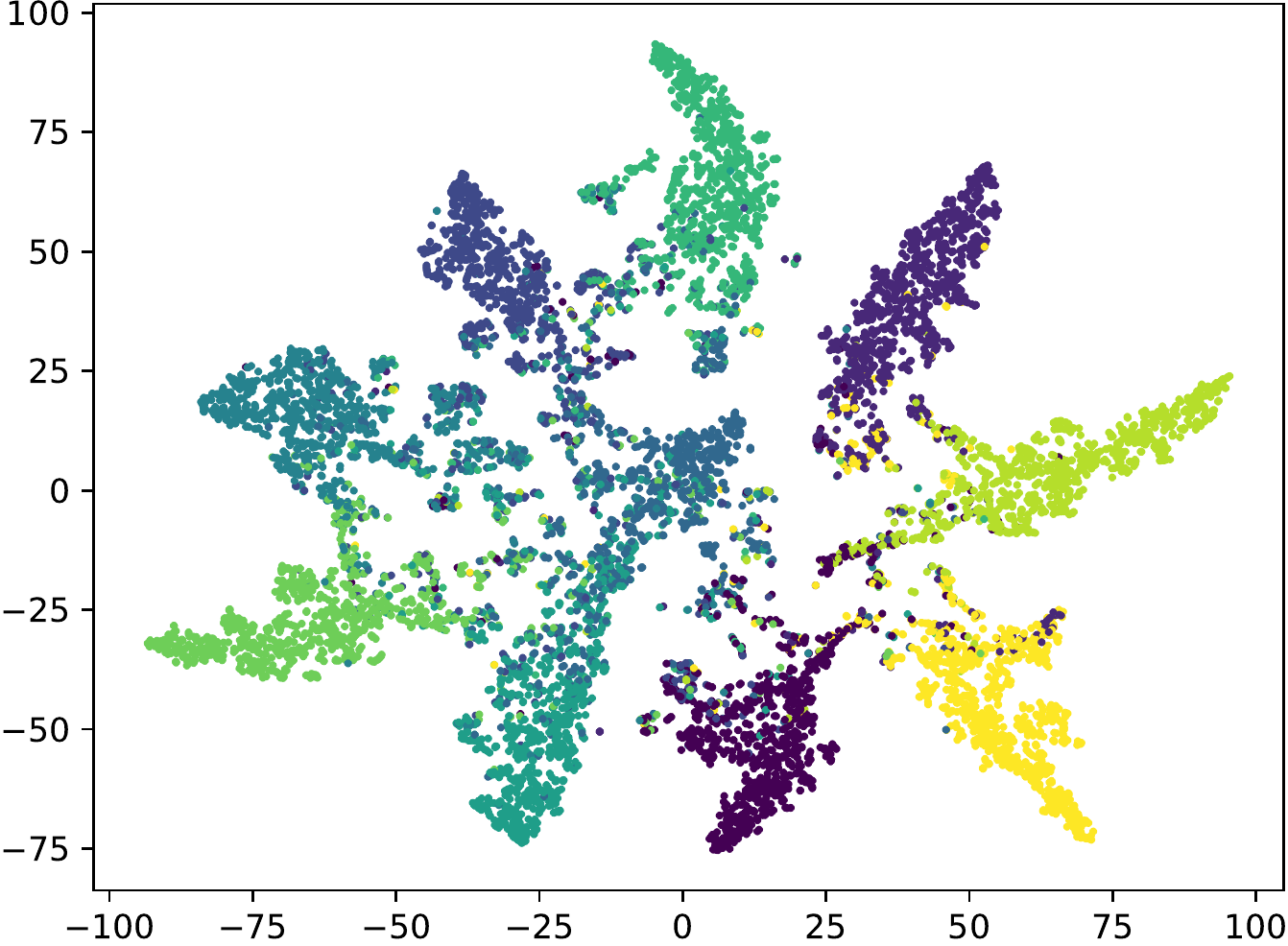}
\caption{LSR ($\xi=0.4$) -- 16.72\%.}
\end{subfigure}%
\hspace{5mm}
\begin{subfigure}{0.35\textwidth}
\includegraphics[width=1\textwidth]{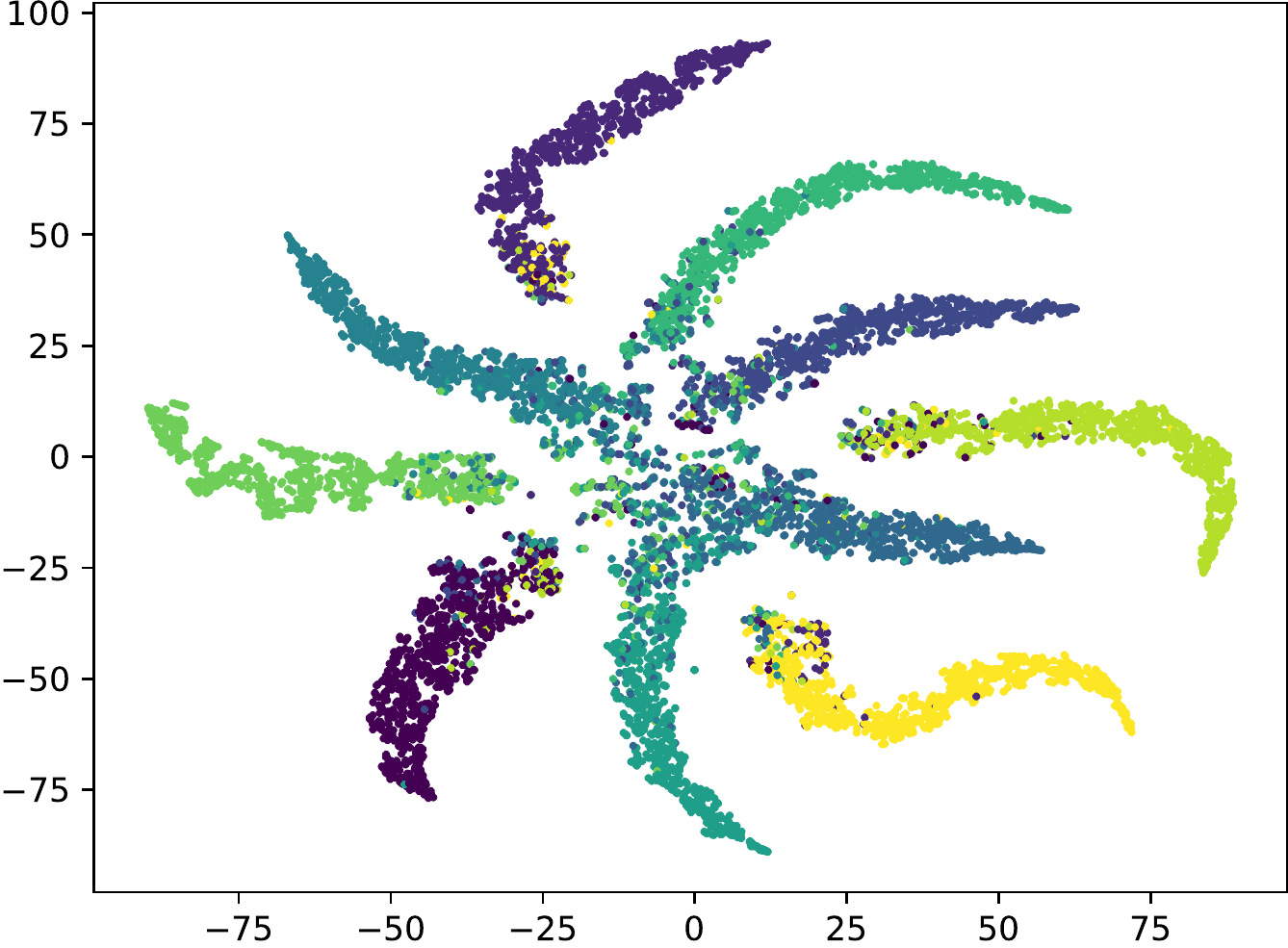}
\caption{SparseLSR ($\xi=0.4$) -- 14.31\%.}
\end{subfigure}%

\captionsetup{justification=centering}
\caption{Visualizing the penultimate layer representations on AlexNet CIFAR-10. Comparing \textit{Label Smoothing Regularization} (LSR) to our proposed \textit{Sparse Label Smoothing Regularization} (SparseLSR).}
\label{fig:fast-label-smoothing-regularization-2}
\end{figure*}

\endgroup

\end{document}